\documentclass[10pt]{article} %
\usepackage[preprint]{tmlr}

\usepackage[utf8]{inputenc} %
\usepackage[T1]{fontenc}    %
\usepackage{hyperref}       %
\usepackage{url}            %
\usepackage{booktabs}       %
\usepackage{amsfonts}       %
\usepackage{nicefrac}       %
\usepackage{microtype}      %
\usepackage[dvipsnames]{xcolor}

\usepackage{graphicx}

\usepackage{amsmath}
\usepackage{amssymb}
\usepackage{mathtools}
\usepackage{amsthm}

\usepackage{stmaryrd}

\hypersetup{
    colorlinks,
    linkcolor={blue!60!black},
    citecolor={cyan!80!black},
    urlcolor={magenta!60!black}
}

\usepackage[capitalize]{cleveref}

\usepackage{dsfont} %
\usepackage{colortbl}
\usepackage{enumitem}
\usepackage{wrapfig}
\usepackage{multirow}
\usepackage{fancybox}
\usepackage{changepage}
\usepackage{blindtext}
\usepackage{titlesec}
\usepackage{bm}
\usepackage{array}
\usepackage{diagbox}
\usepackage{subcaption}
\usepackage{threeparttable}
\usepackage{tabularx}  
\usepackage{algorithm}
\usepackage{algorithmic}
\usepackage{pgfmath}

\usepackage{pgfplots}
\usepackage{pgfplotstable}
\usepgfplotslibrary{groupplots} %
\usepgfplotslibrary{fillbetween}
\usetikzlibrary{arrows.meta, positioning}
\pgfplotsset{compat=1.18}
\pgfplotsset{grid = major, grid style={gray!30!white}}

\definecolor{darkgreen}{rgb}{0.31, 0.47, 0.26}

\usepackage{mdframed}
\definecolor{theoremcolor}{rgb}{255, 255, 255}
\mdfsetup{
    backgroundcolor=theoremcolor,
    linewidth=1pt,
    leftline=true,
    topline=false,
    rightline=false,
    bottomline=false 
}

\newmdtheoremenv{definition}{Definition}
\newmdtheoremenv{proposition}{Proposition}
\newmdtheoremenv{corollary}{Corollary}
\newmdtheoremenv{theorem}{Theorem}
\newmdtheoremenv{lemma}{Lemma}
\newmdtheoremenv{example}{Example}
\newmdtheoremenv{remark}{Remark}

\DeclareMathOperator{\interior}{\mathsf{int}}
\DeclareMathOperator{\cov}{\mathsf{cov}}
\DeclareMathOperator{\conv}{\mathsf{conv}}
\DeclareMathOperator{\dom}{\mathsf{dom}}
\DeclareMathOperator{\relint}{\mathsf{relint}}
\DeclareMathOperator{\diag}{\mathsf{diag}}

\DeclareMathOperator*{\argmax}{\mathsf{argmax}}

\DeclareMathOperator*{\Proj}{\mathsf{P}^\perp}

\def\yt{{\widehat{\bm{y}}_t}}
\def\yk{{\bm{y}^{(k)}}}
\def\x{{\bm{x}}}
\def\y{{\bm{y}}}

\def\cN{{\mathcal{N}}}
\def\cC{{\mathcal{C}}}
\def\cR{{\mathcal{R}}}

\def\cX{{\mathcal{X}}}
\def\cY{{\mathcal{Y}}}

\def\RR{{\mathbb{R}}}
\def\EE{{\mathbb{E}}}

\def\thetav{{\bm{\theta}}}
\def\Thetav{{\bm{\Theta}}}

\def\muv{{\bm{\mu}}}

\def\pfirstsup{\bm{p}^{(1)}_{\bm{\theta},\y}}
\def\pfirstunsup{\bm{p}^{(1)}_{\bm{\theta},\bar{Y}_N}}

\crefname{problem}{Problem}{Problems}

\newlist{inlinelist}{enumerate*}{1}
\setlist[inlinelist]{label=(\roman*)} %

\definecolor{LightGray}{gray}{0.92}

\title{Learning with Local Search MCMC Layers}

\author{\name Germain Vivier-Ardisson \email gvivier@google.com \\
      \addr Google DeepMind \& CERMICS, Paris, France
      \AND
      \name Mathieu Blondel \email mblondel@google.com \\
      \addr Google DeepMind, Paris, France
      \AND
      \name Axel Parmentier \email axel.parmentier@enpc.fr \\
      \addr CERMICS, CNRS, ENPC, Institut Polytechnique de Paris, Marne-la-Vallée, France
}

\def\month{08}  %
\def\year{2026} %
\def\openreview{\url{https://openreview.net/forum?id=M2a0mbACNf}} %

\begin{document}

\maketitle

\begin{abstract}
Integrating combinatorial optimization layers into neural networks has recently attracted significant research interest. However, many existing approaches lack theoretical guarantees or fail to perform adequately when relying on inexact solvers. This is a critical limitation, as many operations research problems are NP-hard, often necessitating the use of neighborhood-based local search heuristics. In this paper, we introduce a principled approach for learning with such inexact solvers. Inspired by the connection between simulated annealing and Metropolis-Hastings, we transform the problem specific neighborhood systems used in local search heuristics into proposal distributions, implementing MCMC on the set of feasible solutions. This allows us to construct differentiable, stochastic combinatorial layers and associated loss functions. Replacing an exact solver with a local search strongly reduces the computational burden of learning on many applications. We demonstrate our approach on a dynamic vehicle routing problem with time windows, binary vector and $k$-subset prediction tasks, as well as a multi-dimensional knapsack decision-focused learning problem.
\end{abstract}

\section{Introduction}

Models that combine neural networks and combinatorial optimization
have recently attracted significant attention \citep{sadana_survey_2024, %
mandi_decision-focused_2024, Donti2017, berthet_learning_2020, %
bengio_machine_2021, edpbook}.
They enrich combinatorial optimization algorithms with context-dependent features, making decisions more resilient to uncertainty. An important subset of this line of research integrates, within a neural network, an integer linear programming layer of the form:
\begin{align}
\label{pb:intro}
\thetav \mapsto \argmax_{\y \in \cY} ~ \langle \thetav, \y \rangle,
\end{align}
where $\mathcal{Y}$ is a finite set of feasible outputs. A black-box algorithm that returns the optimal solution to Problem~\eqref{pb:intro} is known as a \emph{maximum a posteriori (MAP) oracle} in the graphical models and structured prediction literature \citep{wainwright_graphical_2008}, and as a black-box exact solver in operations research.
Such combinatorial optimization layers enable the transformation of learned, continuous
latent representations into structured, discrete outputs, that satisfy complex
constraints.
This setting is known as \emph{decision-focused learning}, where a fixed solver is parameterized by $\thetav$, predicted from features $\x$, in contrast to \emph{neural combinatorial optimization}, which aims to replace the solver entirely with ML-based heuristics.

The main challenge in using such layers lies in end-to-end training: as piecewise-constant functions, they lack meaningful gradients. Many relaxations and loss functions exist for this setting; see \cref{sec:background} for a review. \cref{tab:overview} contrasts them by the type of oracle they require. Some rely on an oracle for a regularized version of Problem~\eqref{pb:intro}, while others use a MAP oracle for the original problem, performing multiple calls per instance for smoothing reasons. Crucially, theoretical guarantees for these methods typically assume exact solvers.

Unfortunately, many problems in operations research are NP-hard in nature, making exact oracles impractical.
Instead, applications often rely on \emph{local search heuristics} (e.g., simulated annealing), which iteratively generate and then accept or reject a neighbor of the current solution. We aim to provide a principled approach for learning with such inexact combinatorial solvers. This is crucial for exploiting popular heuristics from the operations research literature as layers in neural networks.

To do so, we open the solver ``black box'', by bridging local search heuristics and Markov chain Monte-Carlo (MCMC) methods. These lines of research have evolved quite separately, and their links remain unexploited for designing principled combinatorial optimization layers. Specifically, we make the following contributions:
\begin{itemize}[topsep=0pt,itemsep=1.5pt,parsep=2pt,leftmargin=15pt]
\item We integrate local search heuristics as differentiable, stochastic layers in neural networks, by converting their neighborhood systems to proposal distributions, turning the local search oracle into a discrete MCMC sampler over the combinatorial set of solutions.

\item We extend our framework to handle local search heuristics that leverage a diversity of neighborhood systems, enabling this class of powerful solvers to be used as a unified MCMC sampler.

\item We show that the proposed layer yields stochastic gradients of a Fenchel-Young loss \citep{blondel_learning_2020} (even with a single MCMC iteration), leading to principled learning algorithms for conditional and unconditional settings, for which we provide a convergence analysis.

\item The proposed layer reduces the computational bottleneck, especially with few MCMC iterations, enabling larger training instances and better generalization at scale~\citep{parmentier2021learning,parmentier2022learning}.

\item We demonstrate our approach on the EURO Meets NeurIPS 2022 challenge \citep{kool_euro_2023}, a large-scale dynamic vehicle routing problem with time windows, a multi-dimensional knapsack problem, and on binary vector prediction tasks. Abundant additional experiments are included in \cref{sec:numerical} of the appendix.
\end{itemize}

A summary of notations is available at the beginning of the appendix.

\begin{table*}[t]
\centering
\caption{The proposed approach leverages the neighborhood systems used by local search heuristics (inexact solvers) to obtain a differentiable combinatorial layer when usual oracles are not available.}
\begin{small}
\begin{tabular}{cccc}
\toprule
 & Regularization & Oracle & Approach \\
\midrule
Differentiable DP (\citeyear{li_eisner_2009,mensch2018differentiabledynamicprogrammingstructured})
& Entropy & Exact marginal & DP \\
SparseMAP (\citeyear{niculae2018sparsemapdifferentiablesparsestructured})
& Quadratic & Exact MAP & Frank-Wolfe \\
Barrier FW (\citeyear{krishnan2015barrierfrankwolfemarginalinference})
& TRW Entropy & Exact MAP & Frank-Wolfe \\
IntOpt (\citeyear{mandi_interior_2020}) & Log barrier & Interior point solver & Primal-Dual\\
Perturbed optimizers (\citeyear{berthet_learning_2020})
& Implicit via noise & Exact MAP & Monte-Carlo \\
DYS-net (\citeyear{mckenzie_differentiating_2024}) & Quadratic & Projection oracles & Davis-Yin Splitting\\
Blackbox solvers (\citeyear{vlastelica_differentiation_2020})
& None & Exact MAP &
Interpolation \\
Contrastive divergences (\citeyear{hinton_training_2000})
& Entropy & Gibbs / Langevin sampler &
MCMC \\
\midrule
\rowcolor{LightGray}
Proposed & Entropy & Local search
& %
MCMC \\
\bottomrule
\end{tabular}
\end{small}
\label{tab:overview}
\end{table*}
\section{Background and Related Work}
\label{sec:background}

\subsection{Problem setup}
In this paper, our goal is to learn models that incorporate optimization layers of the form:
\begin{equation}
\label{pb:co}
     \widehat{\y}:\thetav\mapsto \argmax_{\bm{y}\in \mathcal{Y}} \, \langle \bm{\theta},\,\bm{y}\rangle + \varphi(\bm{y}),
\end{equation}
where $\mathcal{Y}\subset\mathbb{R}^d$ is a finite but combinatorially large set, and $\varphi$ encodes problem-specific structural costs or preferences on outputs that do not depend on $\thetav$. For example, in our dynamic vehicle routing application (\cref{sec:dvrptw}), $\thetav$ represents context-dependent predicted ``prizes'' for serving clients, while $\varphi(\y)$ encodes the fixed, context-independent routing cost of solution $\y$.
This formulation therefore extends the standard linear objective in Problem~\eqref{pb:intro} by allowing additional problem-specific structure.

We focus on settings where Problem~\eqref{pb:co} is intractable and only
heuristic algorithms are available to obtain an approximate solution. Our goal
is to integrate NP-hard problems arising in operations research (e.g., routing, scheduling, network design), within a neural network. Unfortunately, many existing approaches lack formal guarantees or simply do not work when used with inexact solvers.

We distinguish between two learning settings. 
In the unconditional setting, our goal will be to learn $\thetav \in \RR^d$ from observations $\y_1, \dots, \y_N \in \cY$ which we will later model as samples from a generative process governed by some $\thetav_0$, as detailed in \cref{sec:unsup}. In the conditional setting, we will assume that $\thetav = g_W(\x)$ and our goal will be to learn the model parameters $W$ from observation pairs $(\x_1,\y_1),\dots,(\x_N,\y_N)$.

\subsection{Combinatorial optimization as a layer}

Since the layer defined in \cref{pb:intro} is piecewise constant, a frequent strategy consists in introducing regularization in the problem so as to obtain a continuous relaxation. In some cases, we may have access to an oracle for directly solving a regularized version of the problem.
For instance, if the original problem admits a dynamic programming solution, one can introduce entropic regularization via algorithmic smoothing  with a change of semiring \citep{li_eisner_2009, mensch2018differentiabledynamicprogrammingstructured}. As another example, interior point solvers can be used to compute a logarithmic barrier-regularized solution \citep{mandi_interior_2020}. More recently, \citet{mckenzie_differentiating_2024} handle quadratic regularization by leveraging projection oracles. 

We focus on settings where only a MAP oracle is available for the original, unregularized optimization problem. While prior work is often limited to the linear form in Problem~\eqref{pb:intro} for the latter, our framework also handles the more general Problem~\eqref{pb:co}. Frank-Wolfe-like methods can be used to solve the regularized problem using only MAP oracle calls \citep{niculae2018sparsemapdifferentiablesparsestructured,krishnan2015barrierfrankwolfemarginalinference}. Another strategy consists in injecting noise perturbations
\citep{berthet_learning_2020} in the oracle, which can be shown to be implicitly using regularization.
In both cases, a Fenchel-Young loss \citep{blondel_learning_2020} can be associated, enabling principled supervised learning with such combinatorial problems as output layers.
However, formal guarantees require an exact oracle, which is often called multiple times during the forward pass for smoothing reasons. Our proposal enjoys guarantees even with inexact solvers and a single oracle call.

Regarding differentiation, several strategies are possible. When the approach only needs to differentiate through a (regularized) $\max$, as is the case of Fenchel-Young losses, we can use Danskin's theorem \citep{danskin_theory_1966}. When the approach needs to differentiate a (regularized) $\argmax$, one can use autodiff on the unrolled solver iterations, or implicit differentiation of the Karush-Kuhn-Tucker conditions \citep{Amos2017optnet, agrawal2019differentiableconvexoptimizationlayers,ferber_mipaal_2019,blondel2022efficient}.
Alternatively, \citet{vlastelica_differentiation_2020} propose to compute gradients via continuous interpolation of the solver.

\subsection{Contrastive divergences}

An alternative approach to learning in combinatorial spaces is to use energy-based models (EBMs) \citep{lecun_ebm}, which define a distribution over outputs via a parameterized energy function $E_{\bm{\theta}}$:
\begin{align*}
    \bm{p}_{\bm{\theta}}(\bm{y}) \coloneqq \frac{\exp(E_\thetav(\y))}{\sum_{\y'\in\cY}\exp(E_\thetav(\y'))}\propto\exp(E_{\bm{\theta}}(\bm{y})).
\end{align*}
Such distributions lead to log-likelihood gradients of the form:
\begin{align*}
    \quad\nabla_{\bm{\theta}}\log \bm{p}_{\bm{\theta}}(\bm{y}) = \nabla_{\bm{\theta}}E_{\bm{\theta}}(\bm{y}) - \mathbb{E}_{Y\sim\bm{p}_{\bm{\theta}}}\left[ \nabla_{\bm{\theta}}E_{\bm{\theta}}(Y)\right].
\end{align*}
Therefore, one can use stochastic gradient methods for maximum likelihood estimation (MLE) by sampling from $\bm{p}_{\bm{\theta}}$. However, this is hard both in continuous and combinatorial settings, due to the intractable normalization constant. Contrastive divergences \citep{hinton_training_2000, carreira-perpinan_contrastive_2005, song2021trainenergybasedmodels} address this by using MCMC to obtain (biased) stochastic gradients. Originally developed for restricted Boltzmann machines with $\mathcal{Y}=\{0,1\}^d$ and a Gibbs sampler, they have also been applied in continuous domains via Langevin dynamics \citep{du2020implicitgenerationgeneralizationenergybased, du2021improvedcontrastivedivergencetraining}.

\paragraph{MCMC in discrete spaces.}
Contrastive divergences rely on MCMC to sample the model distribution. Unfortunately, designing MCMC samplers is often case-by-case, and discrete domains have received less attention than continuous ones. Recent efforts adapt continuous techniques, such as Langevin dynamics \citep{zhang2022langevinlikesamplerdiscretedistributions, sun_discrete_2023} or gradient-informed proposals \citep{grathwohl2021oopsitookgradient, rhodes_enhanced_2022}, to discrete settings. However, these works often assume simple state spaces (e.g., the hypercube or categorical codebooks), and do not handle complex constraints ubiquitous in operations research.
\citet{sun_revisiting_2023} allow structured spaces via relaxed constraints in the energy function, yet ignore these structures in their proposal supports. Notably, we emphasize that all these works focus on sampling, not on designing differentiable MCMC layers. 

\section{Local search-based MCMC layers}
\label{sec:mcmc_layers}

This section introduces our core contribution. We first connect local search heuristics and MCMC methods, then use this link to define a differentiable, stochastic layer based on a single neighborhood system (\cref{algo:SA}), and subsequently generalize it to leverage diverse neighborhood systems (\cref{algo:mixture_SA}).

\subsection{From local search to MCMC}
\label{sec:mh}

\paragraph{Local search and neighborhood systems.}

Local search heuristics~\citep{gendreau2010handbook} iteratively generate a neighbor $\y' \in \cN(\yk)$ of the current solution $\yk$, and either accept it or reject it based on an acceptance rule, that depends on the objective function, $\yk$ and $\y'$.
In this context, a neighborhood system $\mathcal{N}$ defines, for each feasible solution $\y \in \mathcal{Y}$, a set of neighbors $\mathcal{N}(\y) \subseteq \mathcal{Y}$.

Neighborhoods are problem-specific, and must respect the structure of the problem, i.e., must maintain solution feasibility. They are typically defined implicitly via a set of allowed \emph{moves} from $\y$. For instance, \cref{table:operators} lists example moves for a vehicle routing problem.

Formally, we denote the neighborhood graph by $G_\mathcal{N} \coloneqq (\mathcal{Y}, E_\mathcal{N})$, where edges are defined by $\mathcal{N}$. We assume the graph is undirected, i.e., $\bm{y}'\in\mathcal{N}(\bm{y})$ if and only if $\bm{y}\in\mathcal{N}(\bm{y}')$, and without self-loops – i.e., $\bm{y}\notin\mathcal{N}(\bm{y})$. A stochastic neighbor generating function is also provided, in the form of a proposal distribution $q(\bm{y}\,, \cdot\,)$ with support either equal to $\mathcal{N}(\bm{y})$ or $\mathcal{N}(\bm{y})\cup \{\bm{y}\}$.

\paragraph{Link between simulated annealing and Metropolis-Hastings.}

A well-known example of local search heuristic is simulated annealing (SA) \citep{sa_kirkpatrick_1983}. It is intimately related to Metropolis-Hastings (MH) \citep{hastings_1970}, an instance of an MCMC algorithm. We provide a unified view of both in \cref{algo:SA}.

\begin{figure*}[t]
\begin{minipage}{0.485\textwidth}
\begin{algorithm}[H]
\caption{SA / MH as a layer}
\label{algo:SA}
\begin{algorithmic}
    \STATE \textbf{Inputs:} $\bm{\theta}\!\in\!\mathbb{R}^d$, $\y^{(0)}\!\in\!\mathcal{Y}$, $(t_k)$, $K\!\in\!\mathbb{N}$, $\mathcal{N}$, $q$
    \FOR{$k=0:K$}
        \STATE {Sample a neighbor  in $\mathcal{N}(\bm{y}^{(k)})$:\\ $\bm{y}' \sim q\left(\bm{y}^{(k)}, \,\cdot\,\right)$}
        \STATE $\alpha(\bm{y}^{(k)},\bm{y}')\gets 1$ (SA) or\\$\alpha(\bm{y}^{(k)},\bm{y}')\gets\frac{q(\bm{y}', \bm{y}^{(k)})}{q(\bm{y}^{(k)}, \bm{y}')}$ (MH)
        \STATE {$U \sim \mathcal{U}([0, 1])$}
        \STATE {$\Delta^{\!(k)}\!\gets\! \langle \bm{\theta} ,\bm{y}' \rangle + \varphi(\bm{y}') -  \langle \bm{\theta} , \bm{y}^{(k)} \rangle - \varphi(\bm{y}^{(k)})$}
        \STATE $p^{(k)} \leftarrow \alpha(\bm{y}^{(k)},\bm{y}')\exp\left(\Delta^{\!(k)}/t_k\right)$
        \STATE {If $U \leq p^{(k)}$, accept move: $\bm{y}^{(k+1)} \gets \bm{y}'$}
        \STATE {If $U > p^{(k)}$, reject move: $\bm{y}^{(k+1)} \gets \bm{y}^{(k)}$}
    \ENDFOR
    \STATE \textbf{Output:}  $\widehat{\y}(\bm{\theta}) \approx \bm{y}^{(K)}$ (SA) or\\ $\yt(\bm{\theta}) = \mathbb{E}_{{\pi}_{\bm{\theta},t}}\left[Y\right] \approx \frac{1}{K}\sum_{k=1}^{K} \bm{y}^{(k)}$ (MH)

\end{algorithmic}
\end{algorithm}
\end{minipage}
\hfill
\begin{minipage}[c]{0.01\textwidth}
    \vspace*{11pt}
    \hspace*{0.01pt}
    \rule{0.4pt}{6.9cm}
\end{minipage}
\hfill
\begin{minipage}{0.485\textwidth}
\begin{algorithm}[H]
\caption{Neighborhood mixture MCMC}
\label{algo:mixture_SA}
\begin{algorithmic}
    \STATE \textbf{Inputs:} $\bm{\theta}\!\in\!\mathbb{R}^d$\!, $\y^{(0)}\!\!\in\!\mathcal{Y}$\!, $t$, $K\!\in\!\mathbb{N}$, $(\mathcal{N}_s,\!q_s)_{s=1}^S$
    \FOR{$k=0:K$}
        \STATE Sample a neighborhood system: \\$s \sim \mathcal{U}(Q(\bm{y}^{(k)}))$
        
        \STATE Sample a neighbor  in $\mathcal{N}_s(\bm{y}^{(k)})$:\\$\bm{y}' \sim q_s(\bm{y}^{(k)},\, \cdot)$
    
        \STATE $\alpha_s(\bm{y}^{(k)}, \bm{y}') \gets \frac{|Q(\bm{y}^{(k)})|}{|Q(\bm{y}')|} \frac{q_s(\bm{y}', \bm{y}^{(k)})}{q_s(\bm{y}^{(k)}, \bm{y}')}$
        \STATE $U \sim \mathcal{U}([0, 1])$
    
       \STATE {$\Delta^{\!(k)}\!\gets\! \langle \bm{\theta} ,\bm{y}' \rangle + \varphi(\bm{y}') -  \langle \bm{\theta} , \bm{y}^{(k)} \rangle - \varphi(\bm{y}^{(k)})$}
        \STATE $p^{(k)} \leftarrow \alpha_s(\bm{y}^{(k)},\bm{y}')\exp\left(\Delta^{\!(k)}/t\right)$
        \STATE {If $U \leq p^{(k)}$, accept move: $\bm{y}^{(k+1)} \gets \bm{y}'$}
        \STATE {If $U > p^{(k)}$, reject move: $\bm{y}^{(k+1)} \gets \bm{y}^{(k)}$}
    \vspace{2pt}
    \ENDFOR

    \STATE \textbf{Output:} $\yt(\bm{\theta}) = \mathbb{E}_{{\pi}_{\bm{\theta},t}}\left[Y\right] \approx \frac{1}{K}\sum_{k=1}^{K} \bm{y}^{(k)}$

\end{algorithmic}
\end{algorithm}
\end{minipage}
\end{figure*}

The difference lies in the acceptance rule, which incorporates a proposal correction ratio for MH, and in the choice of the sequence of temperatures $(t_k)_{k\in\mathbb{N}}$. In the case of SA, it is chosen to ensure $t_k \xrightarrow[]{} 0$. In the case of MH, it is such that $t_k \equiv t$. In this case, the iterates $\yk$ of \cref{algo:SA} follow a time-homogenous Markov chain on $\mathcal{Y}$, defined by the following transition kernel:
\begin{equation}
\label{eq:markov_kernel}
    P_{\bm{\theta}, t}(\bm{y},\bm{y}')= \begin{cases}
        q\left(\bm{y},\bm{y}'\right)\min\left[1, \frac{q(\bm{y}',\bm{y})}{q(\bm{y},\bm{y}')}
        \exp
        \left({\frac{\langle \bm{\theta} \,,\, \bm{y}' \rangle + \varphi(\bm{y}') - \langle \bm{\theta} ,\, \bm{y} \rangle - \varphi(\bm{y})}{t}}\right)
        \right] & \text{if $\bm{y}'\in\mathcal{N}(\bm{y})$,}\\
        1 - \sum_{\bm{y}''\in\mathcal{N}(\bm{y})}P_{\bm{\theta}, t_k}(\bm{y},\bm{y}'') & \text{if $\bm{y}'=\bm{y}$,}\\
        0 &\text{else.}
    \end{cases}
\end{equation}

In past work, the link between the two algorithms has primarily been used to show that SA converges to the exact MAP solution in the limit of infinite iterations \citep{mitra_convergence_1986, faigle_convergence_1988}. If $G_\cN$ is connected and the Markov chain is aperiodic (which is ensured, e.g., as soon as there exists at least one state $\y\in\cY$ such that $P_{\thetav,t}(\y,\y')>0$),  the iterates $(\yk)_{k\in\mathbb{N}}$ of \cref{algo:SA} (MH case) converge in distribution to the following Gibbs distribution (see \cref{proof:stationary_gibbs} for a proof):
\begin{equation}
{\pi}_{\bm{\theta},t}(\bm{y}) \propto \exp\left(\left[\langle \bm{\theta} \,,\,\bm{y}\rangle + \varphi(\bm{y})\right]/t\right).
\label{eq:stationary_gibbs}
\end{equation}
\paragraph{Proposed layer.}

\cref{algo:SA} and this result motivate us to define the combinatorial MCMC layer
\begin{equation}
    \yt(\thetav)\coloneqq \mathbb{E}_{{\pi}_{\bm{\theta},t}}\left[Y\right],
\end{equation}
where $\bm{\theta}\in\mathbb{R}^d$ is the predicted maximization direction and $t>0$ is a temperature parameter, defaulting to $t=1$. Computing $\yt(\thetav)$ is known as the \emph{marginal inference} problem in the graphical models literature \citep{wainwright_graphical_2008}.

Naturally, the estimate of $\yt(\bm{\theta})$ returned by \cref{algo:SA} (MH case) is biased, as the Markov chain cannot perfectly mix in a finite number of iterations.
In \cref{sec:conv_exact_stat}, we empirically find that discarding early iterates of the chain (``burn-in'') does not help mitigate this estimation bias, so we average all iterates by default.
In \cref{sec:learn_sa},  we theoretically show that this bias does not hinder the convergence of the proposed learning algorithms. The next proposition, proved in \cref{proof:layer_properties}, states some useful properties of the proposed layer.
\begin{proposition}
\label{prop:layer_properties}
    Let $\thetav \in \RR^d$, and $\cC\coloneqq\conv(\cY)$ be the convex hull of the feasible set. We have:
    \begin{align*}
        \yt(\bm{\theta})\in\relint(\mathcal{C}),\quad \yt(\bm{\theta})\xrightarrow[t\rightarrow0^+]{}\argmax_{\bm{y}\in \mathcal{Y}} \langle \bm{\theta},\,\bm{y}\rangle + \varphi(\bm{y}),\quad\text{and}\;\quad\yt(\bm{\theta})\xrightarrow[t\rightarrow \infty]{}\frac{1}{|\mathcal{Y}|}\sum_{\bm{y}\in\mathcal{Y}}\bm{y}\;.
    \end{align*}
    Moreover, $\yt$ is differentiable and its Jacobian matrix is given by
    $
    J_{\bm{\theta}}\yt(\bm{\theta}) = \frac{1}{t}\cov_{{\pi}_{\bm{\theta},t}}\left[Y\right].
    $
\end{proposition}

As a covariance matrix under the stationary distribution ${\pi}_{\bm{\theta},t}$, this Jacobian can be estimated via MCMC. This theoretically allows the local search MCMC sampler to function as a generic differentiable layer, enabling the optimization of arbitrary downstream loss functions that depend on $\thetav$ through $\yt(\thetav)$. However, as we show in \cref{sec:learn_sa}, the specific Fenchel-Young losses we rely on provide a strong advantage: their gradients only require evaluating the forward pass $\yt(\thetav)$, bypassing the need to explicitly compute the Jacobian.

\subsection{Mixing neighborhood systems}

Central to local search algorithms in combinatorial optimization is the use of multiple neighborhood systems, or move types, to more effectively explore the solution space \citep{mladenovic1997variable, blum_metaheuristics_2003}. In this section, we propose a tractable way to incorporate such diversity of neighborhood systems into the combinatorial MCMC layer, while preserving the correct stationary distribution.

\paragraph{Move types and neighborhood systems.} Let $\left(\mathcal{N}_s\right)_{s=1}^S$ be a set of different neighborhood systems. In practice, each system corresponds to a specific type of \emph{move} (such as those in \cref{table:operators} for our vehicle routing application), so that $\mathcal{N}_s(\bm{y})$ is the set of solutions reachable from $\bm{y}$ by applying move type $s$. Since not all move types are necessarily applicable to every solution $\bm{y}\in\mathcal{Y}$, we denote by $Q(\bm{y})\subseteq \llbracket 1, S\rrbracket$ the subset of moves valid on $\bm{y}$. Let $\left(q_s\right)_{s\in Q(\bm{y})}$ be the corresponding proposal distributions, such that the support of $q_s(\bm{y},\,\cdot\,)$ is either $\mathcal{N}_s(\bm{y})$ or $\mathcal{N}_s(\bm{y})\cup\{\bm{y}\}$. Let $\bar{\mathcal{N}}$ be the aggregate neighborhood system defined by $\bar{\mathcal{N}}(\y)\coloneqq \bigcup_{s\in Q(\bm{y})}\mathcal{N}_s(\bm{y})$.

\paragraph{Computational challenge of neighborhood mixing.} A standard way to combine these neighborhood systems would be to use \cref{algo:SA} by defining an aggregate proposal $q(\y,\,\cdot\,)$ on $\bar{\cN}(\y)$ as, e.g.:
\begin{align*}
    q(\y,\y') \coloneqq \frac{1}{|Q(\y)|}\sum_{s\in Q(\y)}q_s(\y,\y'),\quad \text{giving:}\quad \alpha(\y,\y') = \frac{|Q(\y)|}{|Q(\y')|}\cdot\frac{\sum_{s\in Q(\y')}q_s(\y',\y)}{\sum_{s\in Q(\y)}q_s(\y,\y')}.
\end{align*}
However, this leads to intractable updates. Indeed, computing the MH correction ratio $\alpha(\y,\y')$ is prohibitively expensive as it involves summing the forward proposal probabilities for all move types in $Q(\y)$ and the reverse probabilities for all move types in $Q(\y')$. The difficulty is that multiple, distinct proposal types can generate the same solution $\y'$ from $\y$. For example, in our vehicle routing application in \cref{sec:dvrptw}, relocating a pair of clients before the first one in a route of three gives the same solution $\y'$ as relocating the first client at the last position (see the \texttt{relocate} and \texttt{relocate pair} moves from \cref{table:operators}). Identifying and calculating all these potential forward and reverse pathways for every step is a significant computational hurdle.

\paragraph{Proposed efficient sampler.} In contrast, the update we propose in \cref{algo:mixture_SA} circumvents this summation entirely by using a move-specific correction term. This extension of MH only requires computing the single individual ratio $\alpha_s(\y,\y')\coloneqq\frac{|Q(\y)|}{|Q(\y')|}\cdot\frac{q_s(\bm{y}', \bm{y})}{q_s(\bm{y}, \bm{y}')}$ for the move type $s$ that was actually sampled.

\begin{proposition}
\label{prop:mixture_sa}
If each neighborhood graph $G_{\cN_s}$ is undirected and without self-loops, and the aggregate neighborhood graph $G_{\bar{\mathcal{N}}}$ is connected, the iterations $(\yk)_{k\in\mathbb{N}}$ produced by \cref{algo:mixture_SA} follow a Markov chain with unique stationary distribution ${\pi}_{\bm{\theta},t}$. 
\end{proposition}

See \cref{proof:mixture_sa} for the proof. Importantly,
only the \emph{aggregate} neighborhood graph $G_{\bar{\mathcal{N}}}$ must be connected. This enables combining neighborhood systems $\cN_s$ that could not connect $\mathcal{Y}$ independently, and an irreducible Markov chain can be obtained by mixing the proposal distributions of reducible ones. As a concrete example, the proposal moves in our dynamic vehicle routing experiment (\cref{sec:dvrptw}) are defined in \cref{table:operators}.

\section{Loss functions and theoretical analysis}
\label{sec:learn_sa}
Building on the differentiable MCMC layer developed in \cref{sec:mcmc_layers}, we construct the corresponding learning framework. We derive principled Fenchel-Young loss functions for our layer, present practical stochastic gradient algorithms for conditional and unconditional learning, and provide theoretical convergence guarantees.

\subsection{Negative log-likelihood and associated Fenchel-Young loss}
\label{sec:associated_fy_loss}

We now show that, similarly to related work on structured prediction and combinatorial optimization layers (see \cref{tab:overview}), the proposed layer $\yt(\thetav)$ can be viewed as the solution of a regularized optimization problem on $\mathcal{C}=\conv(\mathcal{Y})$.
Let $A_t(\thetav) \coloneqq t\cdot \log\sum_{\bm{y}\in\mathcal{Y}}\exp\left(\left[\langle \bm{\theta} \,,\,\bm{y}\rangle + \varphi(\bm{y})\right]/t\right)$ be the cumulant function \citep{wainwright_graphical_2008} associated to ${\pi}_{\bm{\theta},t}$, scaled by $t$. 
We define the regularization function $\Omega_t$ and the corresponding Fenchel-Young loss \citep{blondel_learning_2020} as:
\begin{align*}
    \Omega_t(\muv) \coloneqq A^*_t(\bm{\mu}) = \sup_{\bm{\theta}\in\mathbb{R}^d}\langle \bm{\mu} \,,\,\bm{\theta}\rangle - A_t(\bm{\theta}),\quad\text{and}\quad\ell_t(\bm{\theta}\,;\bm{y}) \coloneqq A_t(\thetav)+ \Omega_t(\bm{y}) - \langle\bm{\theta},\,\bm{y}\rangle.    
\end{align*}
Since $\Omega_t$ is strictly convex on $\relint(\mathcal{C})$ (see \cref{proof:fyl} for a proof) and $\yt(\thetav)=\nabla_{\thetav}A_t(\thetav)$,
the proposed layer is the unique solution of the regularized optimization problem:
\begin{equation*}
    \yt(\thetav)=\argmax_{\bm{\mu}\in\mathcal{C}}\left\{\langle\bm{\theta},\bm{\mu}\rangle-\Omega_t(\bm{\mu})\right\},
\end{equation*}
the Fenchel-Young loss $\ell_t$ is differentiable, 
satisfies $\ell_t(\thetav, \y) = 0 \Leftrightarrow \yt(\thetav) = \y$,
and has gradient given by \text{$\nabla_{\bm{\theta}}\ell_t(\bm{\theta}\,;\bm{y}) = \yt(\thetav) - \bm{y}$} \citep[Proposition 2]{blondel_learning_2020}. It is therefore equivalent, up to a constant, to the negative log-likelihood loss, as we have $-\nabla_{\thetav}\log\pi_{\thetav,t}(\y)=(\yt(\thetav)-\y)/t$. Algorithms \ref{algo:SA} and \ref{algo:mixture_SA} can thus be used to perform MLE, by returning a (biased) stochastic estimate of the gradient of $\ell_t$.

Note that the regularizer $\Omega_t$ extends the influence of the objective term $\varphi$ from the vertices $\cY$ to the interior of the polytope $\mathcal{C}$. This  can be seen from the low temperature behavior of the regularized maximizer $\yt$, which recovers the maximizer of the unregularized objective containing $\varphi$ (see \cref{prop:layer_properties}).
\subsection{Empirical risk minimization}
\label{sec:erm}
In the conditional learning setting, we are given observations $\left(\x_{i},\, \y_i\right)_{i=1}^N \in \left(\mathbb{R}^p\times\mathcal{Y}\right)^N$, and want to fit a model $g_W \colon \RR^p \to \RR^d$ such that $\yt(g_W(\bm{x}_i)) \approx \y_i$. This is motivated by a generative model where, for some weights $W_0$, the data is generated with $\y_i\sim {\pi}_{g_{W_0}(\bm{x}_i), t}$. We aim at minimizing the empirical risk $L_N$:%
\begin{align*}
L_N(W) \!\coloneqq \!\frac{1}{N}\!\sum_{i=1}^N \ell_t\left(g_W(\bm{x}_i)\,; \y_i\right),
\;\; \text{with} \quad
\nabla_W L_N(W) \!=\! \frac{1}{N}\!\sum_{i=1}^N J_W g_W (\bm{x}_i) \left(\yt(g_W(\bm{x}_i)) \!- \!\y_i\right).
\end{align*}
\paragraph{Doubly stochastic gradient estimator.}

In practice, we cannot exactly compute the gradient above: evaluating $\hat{\y}_t$ requires performing exact marginal inference over the combinatorially large set $\cY$, which is generally intractable even if an exact MAP oracle were available. Instead, using \cref{algo:SA} or \ref{algo:mixture_SA} to get an MCMC estimate of $\hat{\y}_t(g_W(\x_i))$, we propose the following estimator:
\begin{align*}
    \nabla_W L_N(W) \approx J_W g_W (\bm{x}_i) \biggl( \frac{1}{K}\sum_{k=1}^{K} \bm{y}_i^{(k)}\!\! - \y_i\biggr),
\end{align*}
\noindent where $\bm{y}_i^{(k)}$ is the $k$-th iterate of the algorithm with maximization direction $\bm{\theta}_i=g_W(\bm{x}_i)$ and temperature $t$. 
This estimator is doubly stochastic, since we sample both data points and Markov iterations, and can be seamlessly used with mini-batches. The term $J_Wg_W(\bm{x}_i)$ is computed via autodiff.

\paragraph{Markov chain initialization.}

Following the contrastive divergence literature \citep{hinton_training_2000}, we initialize the Markov chain at the ground-truth solution in the conditional setting, as $\y_i^{(0)}=\y_i$. In the unconditional setting, we use a persistent initialization \citep{tieleman_training_2008} instead: see \cref{sec:cd_initialization} for a detailed discussion. 
\subsection{Associated Fenchel-Young loss with a single MCMC iteration}
\label{sec:1step}
To obtain an unbiased gradient estimator for the Fenchel-Young loss $\ell_t$ associated with $\yt$, the MCMC sampler must be run until it reaches its stationary distribution $\pi_{\thetav,t}$. This requirement makes any practical estimator with a finite number of steps $K$ inherently biased. Although our theoretical analysis in \cref{sec:unsup} shows that this bias does not hinder the convergence of the proposed learning algorithms, we now demonstrate that when a single MCMC iteration is used ($K=1$), there exists \emph{another} target-dependent Fenchel-Young loss such that the gradient estimator is \emph{unbiased} with respect to that loss.
\begin{proposition}[Existence of a Fenchel-Young loss when $K=1$]
\label{prop:1step_fyl_sup}
Let $\pfirstsup$ denote the distribution of the first iterate $\y^{(1)}$ of the Markov chain defined by \cref{eq:markov_kernel}, with proposal distribution $q$ and initialized at ground-truth $\y\in\mathcal{Y}$. There exists a target-dependent regularization function $\Omega_{\y}$ with the following properties: $\Omega_{\y}$ is $t/\mathbb{E}_{q(\y,\,\cdot\,)}||Y-\y||_2^2$-strongly convex, it is such that
\begin{align*}
    \mathbb{E}_{\pfirstsup}\left[Y\right] = \argmax_{\bm{\mu}\in\conv\left(\mathcal{N}(\y)\cup\{\y\}\right)}\left\{\langle\bm{\theta},\bm{\mu}\rangle-\Omega_{\y}(\bm{\mu})\right\},
\end{align*}
 \noindent and the Fenchel-Young loss $ \ell_{\Omega_{\y}}$ generated by $\Omega_{\y}$ is $\mathbb{E}_{q(\y,\,\cdot\,)}||Y-\y||_2^2/t$-smooth in its first argument, and such that $\nabla_{\bm{\theta}} \ell_{\Omega_{\y}}(\bm{\theta}\,;\y) = \mathbb{E}_{\pfirstsup}\left[Y\right] - \y$.
\end{proposition}
See \cref{proof:1step_fyl_sup} for the construction of $\Omega_{\y}$ and the proof. Thus, the expected first iterate of the MCMC layer, when initialized at the ground-truth solution $\y$, is in fact the solution to an optimization problem regularized by $\Omega_{\y}$ on the local polytope $\conv(\cN(\y)\cup\{\y\})$. Moreover, $\y^{(1)}-\y$ is an unbiased estimator of $\nabla_{\bm{\theta}} \ell_{\Omega_{\y}}(\bm{\theta}\,;\y)$. We give properties of the local regularized maximizer $\EE_{\pfirstsup}[Y]$ in \cref{prop:one_step_properties}.

A similar result in  the unconditional setting with data-based initialization is given in \cref{prop:1step_fyl_unsup}. Interestingly, these results contrast with prior work on the expected CD-$1$ update. Indeed, when applied with Gibbs sampling to train restricted Boltzmann machines, the latter was shown not to be the gradient of any function, let alone a convex one \citep{sutskever_convergence_2010}.

\subsection{Convergence analysis in the unconditional setting}
\label{sec:unsup}

In the unconditional setting, we are given observations $\left(\y_i\right)_{i=1}^N \!\!\in\!\!\mathcal{Y}^N\!$ and want to fit a model ${\pi}_{\bm{\theta},t}$, motivated by an underlying generative model such that $\y_i \sim {\pi}_{\bm{\theta}_0,t}$ for some true parameter $\bm{\theta}_0$. We assume here that $\mathcal{C}=\conv(\mathcal{Y})$ is of full dimension in $\mathbb{R}^d$ (if not, the model is specified only up to vectors $\muv$ orthogonal to the affine subspace spanned by $\mathcal{C}$, as $\pi_{\thetav+\muv,t}\!=\!\pi_{\thetav,t}$).

We have the corresponding empirical and expected risks $L_{N}$ and $L_{\thetav_0}$:
\begin{align*}
L_{N}(\bm{\theta}; \y_1, \dots, \y_N) \coloneqq \frac{1}{N}\sum_{i=1}^N \ell_t\left(\bm{\theta};\, \y_i\right)\;,
\quad
L_{\bm{\theta}_0}(\bm{\theta}) \coloneqq \mathbb{E}_{\left(\y_i\right)_{i=1}^N\sim({\pi}_{\bm{\theta}_0,t})^{\otimes N}}\left[L_{N}(\thetav; \y_1, \dots, \y_N)\right],
\end{align*}
which are minimized for $\bm{\theta}$ such that $\yt(\bm{\theta}) = \bar{Y}_N\coloneqq \frac{1}{N}\sum_{i=1}^N \bm{y}_i $, and for $\bm{\theta}$ such that $\yt(\bm{\theta}) = \yt(\bm{\theta}_0)$, respectively. While $L_N$ is the objective actually minimized in practice, our controlled experimental setting in \cref{sec:binary} allows us to minimize $L_{\thetav_0}$ directly, decoupling the stochasticity of our MCMC estimators from the statistical noise inherent in finite datasets.

\paragraph{Empirical risk minimizer.}
Let $\bm{\theta}^\star_N$ as the minimizer of the empirical risk $L_N$. For it to be defined, we assume that $\bar{Y}_N \in\interior(\mathcal{C})$ (which is always the case for $N$ large enough, as ${\pi}_{\bm{\theta}_0,t}$ has dense support on $\mathcal{Y}$).
A slight variation on Proposition 4.1 in \citet{berthet_learning_2020} gives the following asymptotic normality:
\begin{proposition}[Convergence of the empirical risk minimizer to the true parameter]
\label{prop:asymp_normality}
$$\sqrt{N}(\bm{\theta}^\star_N - \bm{\theta}_0) \xrightarrow[N\rightarrow \infty]{\mathcal{D}} \mathcal{N}\left(\bm{0},\,t^2\cov_{{\pi}_{\bm{\theta}_0,t}}\left[Y\right] ^{-1}\right).$$
\end{proposition}
The proof is given in \cref{proof:asymp}. Naturally, this convergence result is theoretical, as it only holds if one is able to compute the exact minimizer of the empirical risk $L_{N}$, which is impossible in our setting, as we only have access to noisy estimates of its gradient.

\paragraph{Stochastic gradient estimate.} 
We now consider the sample size as fixed to $N$ samples,
and define $\hat{\bm{\theta}}_n$ as the $n$-th iterate of the following stochastic gradient algorithm:
\begin{align}
\label{eq:sgd_mcmc}
    \hat{\bm{\theta}}_{n+1} = \hat{\bm{\theta}}_n + \gamma_{n+1} \left[\bar{Y}_N - \frac{1}{K_{n+1}}\sum_{k=1}^{K_{n+1}}\bm{y}^{(n+1,\,k)}\right] \,,
\end{align}
\noindent where $\bm{y}^{(n+1, k)}$ is the $k$-th iterate of \cref{algo:SA} with temperature $t$, maximization direction $\hat{\bm{\theta}}_n$, and initialized at $\bm{y}^{(n+1, 1)} = \bm{y}^{(n, K_n)}$. This initialization method, reminiscent of persistent contrastive divergence (PCD) learning \citep{tieleman_training_2008}, %
is further discussed in \cref{sec:cd_initialization}.
\begin{proposition}[Convergence of the stochastic gradient estimate]
\label{prop:stochastic_gradient_estimate}
Suppose the following hold for the step sizes $(\gamma_n)_{n\geq 1}$,
sample sizes $(K_n)_{n\geq1}$, and proposal distribution $q$:
\begin{description}[leftmargin=1.2em,style=sameline,itemsep=0.2pt]
  \item[(i)] $\gamma_n = an^{-b}$, with $b \in \left(\tfrac{1}{2},1\right]$ and $a>0$.
  \item[(ii)] $K_{n+1} > \lfloor 1 + a'\exp(\tfrac{8R_\mathcal{C}}{t}\|\hat{\bm{\theta}}_n\|)\rfloor$,
  with $a'>0$ and $R_\mathcal{C} = \max_{\bm{y}\in\mathcal{Y}}\|\bm{y}\|$.
  \item[(iii)] $\tfrac{1}{\sqrt{K_n}} - \tfrac{1}{\sqrt{K_{n-1}}} \leq a''n^{-c}$,
  with $a''>0$ and $c>1-\tfrac{b}{2}$.
  \item[(iv)] $q(\bm{y}, \bm{y}') =
  \begin{cases}
      \tfrac{1}{2d^*}, & \bm{y}' \in \mathcal{N}(\bm{y}),\\
      1 - \tfrac{d(\bm{y})}{2d^*}, & \bm{y}' = \bm{y},\\
      0, & \text{else},
  \end{cases}$
  where $d(\bm{y}) \coloneqq |\mathcal{N}(\bm{y})|$ and
  $d^* \coloneqq \max_{\bm{y}\in\mathcal{Y}} d(\bm{y})$.
\end{description}
Then the iterates $\hat{\bm{\theta}}_n$ defined by \cref{eq:sgd_mcmc}
converge almost surely: $\hat{\bm{\theta}}_n \xrightarrow{\textit{a.s.}} \bm{\theta}^\star_N$.
\end{proposition}
See \cref{proof:stochastic_gradient_estimate} for the proof. The assumption on the proposal distribution $q$ is used for obtaining a closed-form convergence rate bound for the Markov chain  using graph-based geometric bounds \citep{ingrassia_rate_1994}.

\section{Numerical experiments}
\subsection{Dynamic vehicle routing}
\label{sec:dvrptw}

\begin{figure*}[t]
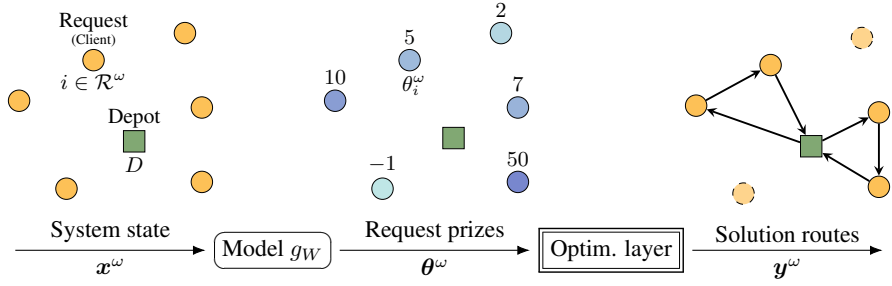

    \centering
    \begin{tikzpicture}[scale=0.95, transform shape]
    \node (start) at (0, 0) {};
    \node[right = 80pt of start] (encoder) {\ovalbox{\begin{Bcenter} Model $g_W$ \end{Bcenter}}};
    \node[right = 80pt of encoder] (optimizer) {\doublebox{\begin{Bcenter} Combinatorial layer \end{Bcenter}}};
    \node[right = 80pt of optimizer] (end) {};

    \draw[->, line width=0.6pt, >=Latex] (start) -- (encoder) node[midway,above] (map_text) {System state} node[midway,below] (map_text2) {$\bm{x}^\omega$};
    \draw[->, line width=0.6pt, >=Latex] (encoder) -- (optimizer) node[midway,above] (weights_text) {Request prizes} node[midway,below] (weights_text2) {$\bm{\theta}^\omega$};
    \draw[->, line width=0.6pt, >=Latex] (optimizer) -- (end) node[midway,above] (path_text) {Solution routes} node[midway,below] {$\bm{y}^\omega$};

    \node[above = 5pt of map_text, inner sep=0pt] (map) {\input{plots/DVRPTW_plot/pipeline/instance}};
    \node[above right = 5pt and -17.7ex of weights_text, inner sep=0pt] (weights) {\input{plots/DVRPTW_plot/pipeline/weights2}};
    \node[above = 5pt of path_text, inner sep=0pt] (path) {\input{plots/DVRPTW_plot/pipeline/solution}};
\end{tikzpicture}
    \vspace{-0.15cm}
    \caption{Overview of the vehicle routing pipeline, represented at request wave $\omega$.}
    \label{fig:dvrptw_pipeline}
\end{figure*}

\begin{table}[t]
\caption{Local search moves used for creating neighborhoods in our vehicle routing experiments.}
\vspace{-0.15cm}
\small
\begin{threeparttable}
\begin{tabularx}{\textwidth}{p{2.5cm} X}
\hline
\texttt{relocate} & Removes a single request from its route and re-inserts it at a different position in the solution. \\
\cellcolor{LightGray} \texttt{relocate pair}  & \cellcolor{LightGray} Removes two consecutive requests from their route and re-inserts them at a different position. \\
\texttt{swap} & Exchanges the position of two requests in the solution. \\
\cellcolor{LightGray} \texttt{swap pair} & \cellcolor{LightGray} Exchanges the positions of two pairs of consecutive requests in the solution. \\
\texttt{2-opt} & Reverses a route segment in the solution. \\
\cellcolor{LightGray} \texttt{serve request} & \cellcolor{LightGray} Inserts a currently unserved request into the solution. \\
\texttt{remove request} & Removes a request from the solution. \\ \hline
\end{tabularx}
\end{threeparttable}
\label{table:operators}
\end{table}

We first empirically validate our approach on the dynamic vehicle routing problem with time windows (DVRPTW) from the \emph{EURO Meets NeurIPS 2022 Vehicle Routing Competition} \citep{kool_euro_2023}. 
\paragraph{Problem setting.}
In this DVRPTW, requests arrive continuously throughout a planning horizon, which is partitioned into a series of discrete delivery waves $\mathcal{W}$. At the start of each wave $\omega \in \mathcal{W}$, a dispatching and routing problem must be solved for the current set of requests $\mathcal{R}^\omega$. A feasible solution $\bm{y}^\omega \in \mathcal{Y}(\mathcal{R}^\omega)$ must dispatch all mandatory requests before their deadline, respect vehicle capacity constraints, and ensure each route visits its assigned requests within their designated time windows. Unserved, postponable requests are carried over to the next wave $\omega+1$. Solutions are represented as binary matrices, where $y^\omega_{i,j} = 1$ if a vehicle travels directly from request $i$ to $j$, and $y^\omega_{i,j} = 0$ otherwise. Further details are provided in \cref{sec:challenge}.

\paragraph{Reduction to supervised learning.}
Following \citet{baty_combinatorial_2024}, we frame each wave's dispatching problem as a prize-collecting vehicle routing problem (PC-VRPTW), where a machine learning model $g_W$ predicts a ``prize'' vector $\thetav^\omega \in \mathbb{R}^{|\mathcal{R}^\omega|}$, representing the benefit of serving each request during wave $\omega$. 

To frame this within our linear objective formulation, we define a mapping $\mathsf{expand} \colon \mathbb{R}^{|\mathcal{R}^\omega|} \to \mathbb{R}^{|\mathcal{R}^\omega| \times |\mathcal{R}^\omega|}$ that broadcasts the predicted vertex prizes $\bm{\theta}^\omega$ into an edge-level prize matrix $\bm{\Theta}^\omega \coloneqq \mathsf{expand}(\bm{\theta}^\omega)$. Since a request $i$ is served if and only if exactly one outgoing edge from $i$ is traversed in the routing graph, we set $\Theta^\omega_{i,j} = \theta^\omega_i$ for all $j \in \mathcal{R}^\omega$. This allows the total collected prize to be written as a dot product over edge-level variables:
\begin{align*}
\sum_{i,j \in \mathcal{R}^\omega} \theta^\omega_i y^\omega_{i,j} = \sum_{i,j \in \mathcal{R}^\omega} \Theta^\omega_{i,j} y^\omega_{i,j} = \langle \bm{\Theta}^\omega, \bm{y}^\omega \rangle.
\end{align*}
The resulting PC-VRPTW fits the general form of Problem~\eqref{pb:co}:
\begin{align}
\label{pb:dvrptw}
\widehat{\bm{y}}(\Thetav^\omega) = \argmax_{\bm{y}\in\mathcal{Y}(\mathcal{R}^\omega)} \langle \Thetav^\omega, \bm{y}\rangle + \varphi(\bm{y}),
\end{align}
where $\varphi(\bm{y}) \coloneqq -\langle \bm{c}, \bm{y}\rangle$ is the negative routing cost, with $c_{i,j} \geq 0$ being the travel cost from request $i$ to $j$. \cref{fig:dvrptw_pipeline} shows the overall pipeline. The model is trained to imitate an anticipative oracle $f^\star$, providing the ground-truth for supervised learning. We compute $f^\star$ by solving a static VRPTW where all future information in the instance is revealed upfront, turning dynamic dispatching waves into static time windows.

\begin{figure}[t]
    \centering
    \begin{subfigure}[b]{0.329\textwidth}
        \centering
        \includegraphics[width=\textwidth]{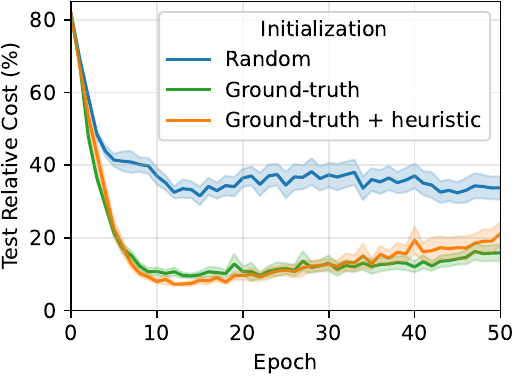}
        \label{fig:dvrptw_init}
    \end{subfigure}
    \begin{subfigure}[b]{0.329\textwidth}
        \centering
        \includegraphics[width=\textwidth]{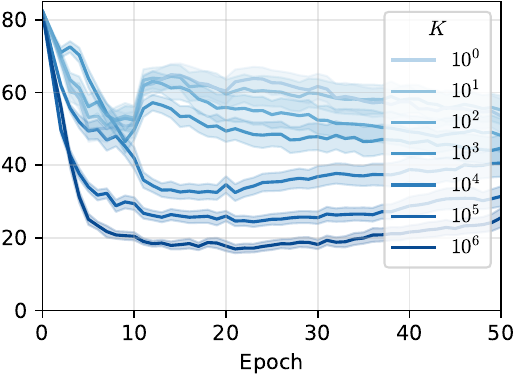}
        \label{fig:dvrptw_iter_random}
    \end{subfigure}
    \begin{subfigure}[b]{0.329\textwidth}
        \centering
        \includegraphics[width=\textwidth]{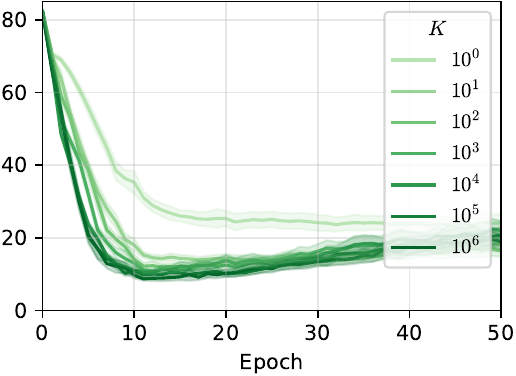}
        \label{fig:dvrptw_iter_gt}
    \end{subfigure}
    \caption{Test relative cost (\%) w.r.t. $f^\star$. \textbf{Left}: varying initialization method. \textbf{Center}:  varying number of Markov iterations $K$, random initialization. \textbf{Right}: varying $K$, ground-truth initialization.}
    \label{fig:dvrptw_init_iter}
\end{figure}

\paragraph{Approach and baseline.}

Problem~\eqref{pb:dvrptw} is intractable as such, so that one cannot rely on an exact MAP oracle to solve it.
The baseline \citet{baty_combinatorial_2024}, winner of the competition, relies on a perturbation-based method (PFY, \citet{berthet_learning_2020}) with the state-of-the-art prize-collecting hybrid genetic search (PC-HGS) heuristic \citep{Vidal_2022} $\Thetav\mapsto\tilde{\y}(\Thetav)$ as a combinatorial optimization layer. Since $\tilde{\y}$ is an inexact solver, the theoretical guarantees granted by the framework of \citet{berthet_learning_2020} no longer hold.

Our approach instead uses a local search MCMC layer to train $g_W$. We use a mixture of proposals (\cref{algo:mixture_SA}) defined precisely in \cref{sec:dvrptw_proposals}, derived from the local search moves used by the PC-HGS solver itself (which are summarized in \cref{table:operators}). At inference time, we follow the baseline, and use $f_W\coloneqq \tilde{\y}\circ\mathsf{expand}\circ g_W$.

\paragraph{Results.}

We use the competition's metric: the routing cost over full instances with multiple dispatching waves, relative to that of the anticipative oracle $f^\star\!$. In \cref{fig:dvrptw_init_iter} (left), we observe that initializing the Markov chain with the ground-truth solution clearly outperforms a random start, even more so when the initialization point is further refined by a fast initialization heuristic (which is the same as the one used by $\tilde{\y}$). While performance increases with the number of MCMC iterations $K$ (\cref{fig:dvrptw_init_iter}, center), ground-truth initialization allows for efficient learning even with $K=1$ (\cref{fig:dvrptw_init_iter}, right), highlighting the theoretical result of \cref{prop:1step_fyl_sup}, in which we showed that it yields unbiased gradient estimates of another Fenchel-Young loss.

In \cref{tab:fixed_compute}, we compare training methods under a fixed time budget for the layer's forward pass, which is the main computational bottleneck. While the heavily engineered, population-based (perturbed) PC-HGS baseline eventually overtakes our method given ample time due to its broader exploration capabilities, it does so at the cost of theoretical guarantees. In the lower time-limit regimes critical for efficient neural network training, our principled MCMC approach provides a superior gradient signal. Full experimental details and additional results are in \cref{sec:additional_dvrptw}.

\begin{table}[t!]
\centering
\caption{Best test relative cost (\%) w.r.t. $f^\star$ for different training methods and time limits.}
\label{tab:fixed_compute}
\begin{tabular}{lcccccc}
\toprule
\textbf{Time limit (ms)} & \textbf{1} & \textbf{5} & \textbf{10} & \textbf{50} & \textbf{100} & \textbf{1000} \\
\midrule
PFY with $\tilde{\y}$ & $65.2\pm 5.8$  & $13.1\pm 3.4$ & $8.7\pm 1.9$ &$ 6.5\pm 1.1$ & $6.3\pm 0.8$ & $\mathbf{5.5}\pm 0.4$ \\
Proposed ($\y^{(0)}=\y$) & $10.0\pm 1.7$ & $12.0\pm 2.6$ & $11.8\pm 2.8$ & $9.1\pm 1.7$ & $8.4\pm 1.7$ & $7.7 \pm 1.1$ \\
Proposed ($\y^{(0)} = \text{heuristic}(\y)$)
 & $\mathbf{7.8}\pm 0.8$ & $\mathbf{7.2}\pm 0.6$ & $\mathbf{6.3}\pm 0.7$ & $\mathbf{6.2}\pm 0.8$ & $\mathbf{5.9}\pm 0.7$ & $5.9\pm 0.6$ \\
\bottomrule
\end{tabular}
\end{table}

\subsection{Learning to predict binary vectors}
\label{sec:binary}

\paragraph{Controlled setting.} To further validate the proposed gradient estimators, we use a synthetic unconditional learning task with hypercube output space, $\cY=\{0,1\}^d$.
This setting is ideal for controlled experiments as the Gibbs distribution $\pi_{\thetav,t}$ is fully factorized, leading to trivial sampling with independent Bernoulli variables $y_i\sim\sigma(\theta_i/t)$, and a closed-form expectation $\EE_{\pi_{\thetav,t}}[Y]=\sigma(\thetav/t)$, where $\sigma$ is the logistic sigmoid function.

\begin{figure}[t]
\centering
\begin{subfigure}[b]{0.329\textwidth}
\centering
\includegraphics[width=\textwidth]{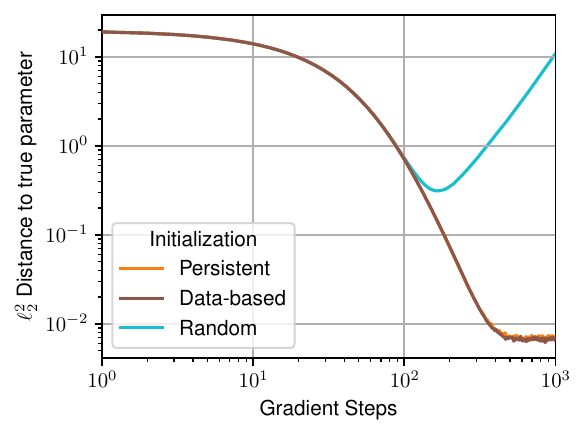}
\caption{Varying initialization}
\label{fig:sa_cube_init_unsupervised_distance_main}
\end{subfigure}
\hfill
\begin{subfigure}[b]{0.329\textwidth}
\centering
\includegraphics[width=\textwidth]{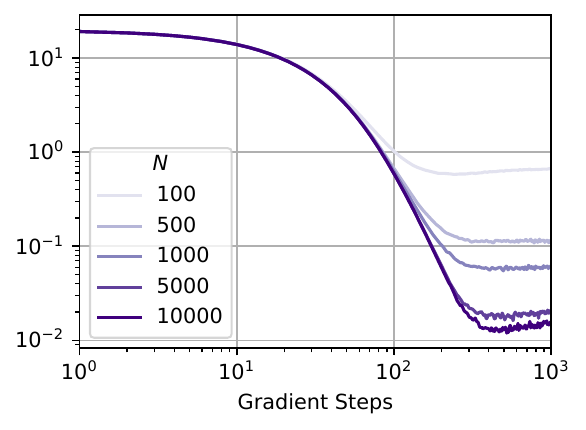}
\caption{Varying dataset size $N$}
\label{fig:sa_cube_samples_unsupervised_distance_main}
\end{subfigure}
\hfill
\begin{subfigure}[b]{0.329\textwidth}
\centering
\includegraphics[width=\textwidth]{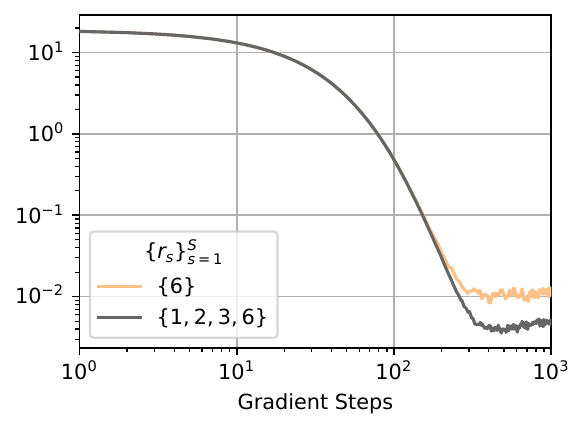}
\caption{Varying neighborhoods}
\label{fig:sa_mixture_1236_unsupervised_distance_main}
\end{subfigure}
\caption{Squared distance $||\hat{\thetav}_n-\thetav_0||_2^2$ to the true parameter over optimization steps.}
\label{fig:hypercube_ablation}
\end{figure}

This allows us to both faithfully generate datasets from a known distribution  $\pi_{\thetav,t}$\,, and to minimize the expected risk $L_{\thetav}$ directly (see \cref{sec:unsup} for its definition). This enables decoupling the noise from our MCMC estimator from the statistical noise inherent in finite datasets.

The goal is to recover a known ``true'' parameter vector $\thetav_0$ from independent samples $(\y_i)_{i=1}^N\sim (\pi_{\thetav_0,t})^{\otimes N}$. We summarize our key findings in \cref{fig:hypercube_ablation}, where we plot the distance to $\thetav_0$ along a stochastic gradient trajectory, either minimizing $L_N$ (left) or $L_{\thetav_0}$ (center, right). Full experimental and theoretical details are available in \cref{sec:numerical}, together with additional results on both the hypercube and the top-$\kappa$ polytope.

\paragraph{Results.} The results highlight three important aspects for effective learning. First, persistent and data-based initializations for the MCMC chains are critical (see \cref{sec:cd_initialization} for a detailed discussion), vastly outperforming random restarts, which introduce systematic bias in the gradient estimation (\cref{fig:hypercube_ablation}, center). 
Second, a larger dataset size $N$ provides a better approximation of the expected risk, leading to a more accurate parameter recovery (\cref{fig:hypercube_ablation}, left), in line with \cref{prop:asymp_normality}. 
Finally, using a mixture of proposals with \cref{algo:mixture_SA} (defining Hamming distance-based neighborhood systems $(\mathcal{N}_{r_s})_{s=1}^S$ by $\y'\in\mathcal{N}_{r_s}(\bm{y}) \Leftrightarrow d_H\left(\bm{y},\,\bm{y}'\right) = r_s$) enables more effective exploration, improving convergence compared to a single proposal type (\cref{fig:hypercube_ablation}, right).

\subsection{Multi-dimensional knapsack problem}
\label{sec:mkp}

Finally, we evaluate our framework on a decision-focused learning problem where the combinatorial optimization layer is a multi-dimensional knapsack problem (MKP) \citep{martello_knapsack_1990, kellerer_knapsack_2004}. We benchmark our method against a broader landscape of predict-and-optimize baselines provided by the \texttt{PyEPO} library \citep{tang_pyepo_2023}. Experimental and methodological details are given in \cref{sec:mkp_details}.

\paragraph{Problem formulation.}
We consider the decision-focused learning setup where the goal is to select a subset of items to maximize a total value while respecting $M$ capacity constraints. The item values $\thetav$ are predicted from features $\x$. Formally, the combinatorial optimization problem is defined as:
\begin{align*}
    \widehat{\y}(\thetav)\coloneqq \argmax_{\y\in\{0,1\}^d}\;\sum_{i=1}^d\theta_i y_i \quad \text{s.t.}\quad\sum_{i=1}^d w_{i,j}y_i\leq C_j\quad \forall j\in[M],
\end{align*}
where $\thetav=g_W(\x) \in \RR^d$ are the item values, $w_{i,j}\geq 0$ is the weight of item $i$ in dimension $j$, and $C_j$ is the capacity of dimension $j$. The feasible set is $\cY\coloneqq \{\y\in\{0,1\}^d\mid \forall j\in[M],\;\sum_{i=1}^d w_{i,j}y_i\leq C_j\}$. We are given a training set $(\x_i,\y_i)_{i=1}^N$ (the \texttt{SPO+} baseline also requires access to the true values $\thetav_i$). At test time, given only $\x$, the goal is to predict $\y$ with minimal regret compared to the ground-truth.

\paragraph{Proposed layer.} For our Local Search-MCMC layer $\widehat{\y}_t$, we use \cref{algo:mixture_SA} with ground-truth initialization, temperature $t=1.0$, and a mixture of three proposal distributions, detailed in \cref{sec:mkp_proposals}.

\paragraph{Baselines.}
We compare against four established decision-focused learning methods from the \texttt{PyEPO} library: smart predict-then-optimize (\texttt{SPO+}, \citet{elmachtoub_smart_2020}), perturbed optimizers using $K=5$ Monte-Carlo samples (\texttt{PFY}, \citet{berthet_learning_2020}), negative identity backpropagation (\texttt{NID}, \citet{sahoo_backpropagation_2023}), and noise-contrastive estimation (\texttt{NCE}, \citet{mulamba_contrastive_2021}). 

To ensure a strictly fair comparison in terms of test-time compute, all methods (including the proposed approach) are evaluated identically on the validation and test sets: the model predicts $\thetav$, and a Gurobi ILP solver is called with a 50ms time limit to generate the final predicted item selection $\y$.

\begin{figure*}[t]
    \centering
    \includegraphics[width=0.99\linewidth]{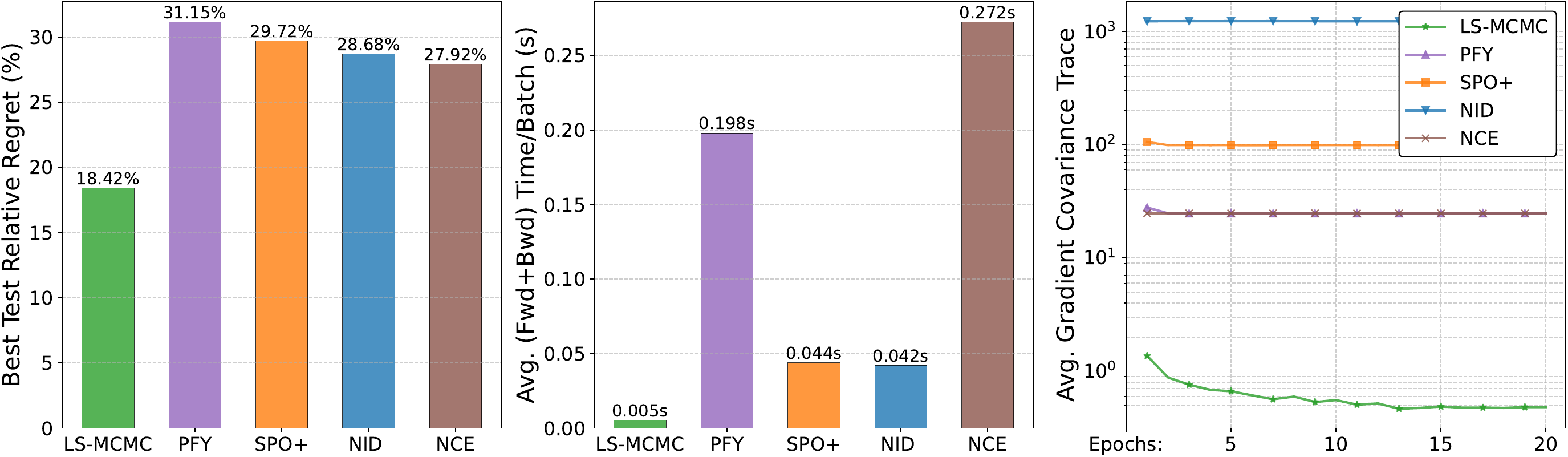}
    \caption{Benchmark results on the MKP.}
    \label{fig:mkp_benchmark}
    \vspace{-0.1cm}
\end{figure*}

\paragraph{Results.}
We generate a dataset $(\x_i,\y_i)_{i=1}^N$ using \texttt{PyEPO} with $N=2000$, $d=100$ and $J=50$ (we also give access to the true item values $\thetav_i$ to the \texttt{SPO+} loss). Our approach achieves competitive test relative regret (\cref{fig:mkp_benchmark}, left) while drastically reducing the computational burden (\cref{fig:mkp_benchmark}, center). The variance of the LS-MCMC gradients remains consistently lower than that of other methods throughout training (\cref{fig:mkp_benchmark}, right), showing that the proposed method provides a stable signal for learning.

In \cref{sec:mkp_additional_results}, we provide additional results on the effect of varying the combinatorial layer's time limit.

\section{Conclusion, limitations and future work}

In this paper, we introduced a principled framework for integrating NP-hard combinatorial optimization layers into neural networks without relying on exact solvers. Our approach adapts neighborhood systems from the metaheuristics community to design structure-aware proposal distributions for combinatorial MCMC. 
This leads to significant training speed-ups, making it possible to tackle larger problem instances, which is crucial in operations research, where scaling up leads to substantial value creation by reducing marginal costs.
\paragraph{Limitations.} A central trade-off of our method is that it is less ``black-box'' than approaches based on solver perturbations. Indeed, our framework requires opening the heuristic solver to access its internal local search moves and computing the associated MCMC correction ratios, rather than simply treating the solver as an opaque optimization oracle. However, this loss of abstraction is precisely what allows us to retrieve theoretical guarantees when relying on inexact solvers.

A second limitation is that our theoretical framework, via its connection to Metropolis-Hastings, currently restricts the optimization layer to single-trajectory local search algorithms. While highly efficient in low-compute regimes, the latter naturally possess lower global exploration capabilities than more involved metaheuristics, such as the hybrid genetic search used as a baseline in our vehicle routing experiment.

\paragraph{Future work.} We frame gradient estimation through combinatorial layers as a sampling problem, and this work is a first step towards revisiting traditional operations research metaheuristics from an MCMC perspective in this context. While we focus on local search heuristics, recent work extends this paradigm to large neighborhood search algorithms, which leverage local MAP oracles \citep{vivier-ardisson_regularized_2026}. Exploring connections with population-based metaheuristics remains a promising future direction.

\bibliographystyle{plainnat}
\bibliography{bibliography}
\newpage

\appendix

\section*{Notation}

\begin{tabular}{l p{13cm}}
\hline
\textbf{Notation}&\textbf{Description}\\
\hline
$\langle\thetav,\y\rangle$ & Euclidean inner product between two vectors $\thetav, \y \in \RR^d$. \\
\cellcolor{LightGray}$\conv(\cY)$ & \cellcolor{LightGray}Convex hull of a set $\cY$. \\
$\relint(\cC)$ & Relative interior of a set $\cC$. \\
\cellcolor{LightGray}$\dom(\Omega)$ & \cellcolor{LightGray}Domain of a function $\Omega:\RR^d\rightarrow\RR\cup\{\infty\}$, defined as $\{\muv\in\RR^d:\Omega(\muv)<\infty\}$. \\
$\Omega^*$ & Fenchel conjugate of $\Omega$, defined as $\Omega^*(\thetav) \coloneqq \sup_{\muv\in\RR^d}\langle\thetav,\muv\rangle-\Omega(\muv)$. \\
\cellcolor{LightGray}$\nabla\Omega$ & \cellcolor{LightGray}Gradient of $\Omega$. \\
$\partial\Omega$ & Subgradient of $\Omega$.\\
\cellcolor{LightGray}$J_\x f(\x,\y)$ & \cellcolor{LightGray}Jacobian of a function $f:\mathcal{X}\times\cY\rightarrow$ $\,\mathbb{R}^d$ with $\mathcal{X}\subseteq$$\,\mathbb{R}^n$ at point $(\x,\y)$ w.r.t. $\x$, viewed as a matrix $J_\x f(\x,\y)\in \RR^{n\times d}$.\\
$\mathcal{U}(\mathcal{X})$ & Uniform distribution on a set $\mathcal{X}$. \\
\cellcolor{LightGray}$\cN(\x, \Sigma)$ & \cellcolor{LightGray}Normal distribution with mean $\x\in\RR^d$ and covariance $\Sigma\in\RR^{d\times d}$. \\
\hline
\end{tabular}

\section{Experiments on empirical convergence of gradients and parameters}
\label{sec:numerical}

In this section, we evaluate the proposed approach on two discrete output spaces: sets and $\kappa$-subsets. These output spaces are for instance useful for multilabel classification. We focus on these output spaces because the exact MAP and marginal inference oracles are available, allowing us to compare our gradient estimators to exact gradients. We set $\varphi\equiv0$ in these experiments.

\subsection{Polytopes and corresponding oracles}
\label{sec:polytopes}

The vertex set of the first polytope is the set of binary vectors in $\mathbb{R}^d$, which we denote $\mathcal{Y}^d \coloneqq \{0, 1\}^d$, and $\conv(\mathcal{Y}^d) = [0,1]^d$ is the ``hypercube''. The vertex set of the second is the set of binary vectors with exactly $\kappa$ ones and $d-\kappa$ zeros (with $0< \kappa < d$),
\begin{align*}
\mathcal{Y}_\kappa^d \coloneqq \bigl\{ \bm{y} \in \{0, 1\}^d:\, \langle \bm{y},\bm{1}\rangle = \kappa \bigr\},
\end{align*}

and $\conv(\mathcal{Y}_\kappa^d)$ is referred to as ``top-$\kappa$ polytope'' or ``hypersimplex''. 
Although these polytopes would not provide relevant use cases of the proposed approach in practice, since exact marginal inference oracles are available (see below), they allow us to compare the Fenchel-Young loss value and gradient estimated by our algorithm to their true value.

\paragraph{Marginal inference.} 

For the hypercube, we have:
\begin{align*}
    \mathbb{E}_{{\pi}_{\bm{\theta},t}}\left[Y_i\right] &= \sum_{\bm{y}\in\mathcal{Y}^d}\frac{\exp\left(\langle \bm{\theta},\bm{y}\rangle/t\right)}{\sum_{\bm{y}'\in\mathcal{Y}^d}\exp\left(\langle \bm{\theta},\bm{y}'\rangle/t\right)}y_i = \sum_{\bm{y} \in \{0, 1\}^d} \frac{\exp\left(\sum_{j=1}^d \theta_j y_j / t \right)}{\sum_{\bm{y}' \in \{0, 1\}^d} \exp\left(\sum_{j=1}^d \theta_j y'_j / t \right)} y_i \\
    &= \sum_{y_i \in \{0, 1\}} \sum_{\bm{y}_{-i} \in \{0, 1\}^{d-1}} \frac{\exp\left( \theta_i y_i / t + \sum_{j \neq i} \theta_j y_j / t \right)}{\sum_{y'_i \in \{0, 1\}} \sum_{\bm{y}'_{-i} \in \{0, 1\}^{d-1}} \exp\left( \theta_i y'_i / t + \sum_{j \neq i} \theta_j y'_j / t \right)} y_i \\
    &= \sum_{y_i \in \{0, 1\}} \frac{\exp\left( \theta_i y_i / t \right)}{\sum_{y'_i \in \{0, 1\}} \exp\left( \theta_i y'_i / t \right)} y_i \sum_{\bm{y}_{-i} \in \{0, 1\}^{d-1}} \frac{\exp\left(\sum_{j \neq i} \theta_j y_j / t \right)}{\sum_{\bm{y}'_{-i} \in \{0, 1\}^{d-1}} \exp\left(\sum_{j \neq i} \theta_j y'_j / t \right)} \\
    &= \sum_{y_i \in \{0, 1\}} \frac{\exp\left( \theta_i y_i / t \right)}{\sum_{y'_i \in \{0, 1\}} \exp\left( \theta_i y'_i / t \right)} y_i \\
    &= \frac{\exp\left( \theta_i / t \right)}{1 + \exp\left( \theta_i / t \right)} \\
    &= \sigma\left( \frac{\theta_i}{t} \right),
\end{align*}
which gives $\mathbb{E}_{{\pi}_{\bm{\theta},t}}\left[Y\right]= \sigma\left( \frac{\bm{\theta}}{t} \right)$, where the logistic sigmoid function $\sigma$ is applied component-wise. The cumulant function is also tractable, as we have
\begin{align*}
    \log \sum_{\bm{y} \in \mathcal{Y}^d} \exp\left( \langle \bm{\theta}, \bm{y} \rangle/t \right)
    &= \log \sum_{\bm{y} \in \{0, 1\}^d} \exp\left( \sum_{i=1}^d \theta_i y_i/t \right) \\
    &= \log \sum_{\bm{y}_1=0}^1 \sum_{\bm{y}_2=0}^1 \cdots \sum_{\bm{y}_d=0}^1 \exp\left( \sum_{i=1}^d \theta_i y_i/t \right) \\
    &= \log \prod_{i=1}^d \sum_{y_i=0}^1 \exp\left( \theta_i y_i/t \right) \\
    &= \log \prod_{i=1}^d \left( 1 + \exp\left( \theta_i/t \right) \right) \\
    &= \sum_{i=1}^d \log\left( 1 + \exp\left(\theta_i/t\right) \right).
\end{align*}
Another way to derive this is via the Fenchel conjugate.\newline

For the top-$\kappa$ polytope, such closed-form formulas do not exist for the cumulant and its gradient. However, they can be computed efficiently using Shannon entropy-regularized dynamic programming. We rely on the framework of \citet{vivier-ardisson_differentiable_2026}, using their Algorithm 1 to compute the cumulant function and their Algorithm 2 to compute its gradient (marginal inference). This allows us to obtain true Fenchel-Young loss values and their gradients in $\mathcal{O}(d\kappa)$ time and space complexity.

\paragraph{Sampling.} For the hypercube, sampling from the Gibbs distribution on $\mathcal{Y}^d$ has closed form. Indeed, the latter is fully factorized, and we can sample $\bm{y} \sim {\pi}_{\bm{\theta},t}$ by sampling independently each component with $y_i \sim \textrm{Bern} \left(\sigma(\theta_i/t)\right)$. Sampling from ${\pi}_{\bm{\theta},t}$ is also possible on $\mathcal{Y}_\kappa^d$ by sampling coordinates iteratively using the computed dynamic programming table, as per Algorithm 3 in \citet{vivier-ardisson_differentiable_2026} or Algorithm 2 in \citet{ahmed_simple_2024}.

\subsection{Neighborhood graphs}

\paragraph{Hypercube.} On $\mathcal{Y}^d$, we use a family of neighborhood systems $\mathcal{N}_\leq^r$ parameterized by a Hamming distance radius $r\in [d-1]$. The graph is defined  by:
\begin{align*}
\forall \bm{y},\bm{y}'\in \mathcal{Y}^d:\;\bm{y}'\in \mathcal{N}_\leq^r(\bm{y}) \Leftrightarrow 1\leq d_H\left(\bm{y},\,\bm{y}'\right) \leq r.
\end{align*}
That is, two vertices are neighbors if their Hamming distance is at most $r$. This graph is regular, with degree $|\mathcal{N}_\leq^r(\bm{y})| = \sum_{i=1}^r \binom{d}{i}
$. This graph is naturally connected, as any binary vector $\bm{y}'$ can be reached from any other binary vector $\bm{y}$ in $||\bm{y}'-\bm{y}||_1$ moves, by flipping each bit where $y_i' \neq y_i$, iteratively. Indeed, this trajectory consists in moves between vertices with Hamming distance equal to $1$, and are therefore along edges of the neighborhood graph, regardless of the value of $r$.\newline

We also use a slight variation on this family of neighborhood systems, the graphs $\mathcal{N}_=^r$, defined by:
\begin{align*}
\forall \bm{y},\bm{y}'\in \mathcal{Y}^d:\;\bm{y}'\in \mathcal{N}_=^r(\bm{y}) \Leftrightarrow d_H\left(\bm{y},\,\bm{y}'\right) = r.
\end{align*}
These graphs, on the contrary, are not always connected: indeed, if $r$ is even, they contain two connected components (binary vectors with an even sum, and binary vectors with an odd sum). We only use such graphs when experimenting with neighborhood mixtures (see \cref{algo:mixture_SA}), by aggregating them into a connected graph.

\paragraph{Top-$\kappa$ polytope.} 

On $\mathcal{Y}_\kappa^d$, we use a family of neighborhoods systems $\mathcal{N}^{s}$ parameterized by a number of ``swaps'' $s\in \llbracket 1, \min(\kappa, d-\kappa) \rrbracket$. The graph is defined by
\begin{align*}
\forall \bm{y}, \bm{y}' \in \mathcal{Y}_\kappa^d:\; \bm{y}'\in\mathcal{N}^s(\bm{y}) \Leftrightarrow d_H\left(\bm{y},\,\bm{y}'\right) = 2s.
\end{align*}
That is, two vertices are neighbors if one can be reached from the other by performing $s$ ``swaps'', each swap corresponding to flipping a $1$ to a $0$ and vice-versa. This ensures that the resulting vector is still in $\mathcal{Y}_\kappa^d$. All $s$ swaps must be performed on distinct components. The resulting graph is known as the \textit{Generalized Johnson Graph} $J(d, \kappa, \kappa-s)$, or \textit{Uniform Subset Graph} \citep{chen_hamiltonian_1987}. It is a regular graph, with degree $|\mathcal{N}^s(\bm{y})| = \binom{\kappa}{s}\binom{d-\kappa}{s}
$. It is proved to be connected in \citet{jones_automorphisms_2005}, except if $d = 2\kappa$ and $s = \kappa$ (in this case, it consists in $\frac{1}{2}\binom{d}{\kappa}$ disjoint edges).

When $s = 1$, the neighborhood graph is the Johnson Graph $J(d, \kappa)$, which coincides with the graph associated to the polytope $\conv(\mathcal{Y}_\kappa^d) = \Delta_{d, \kappa}$ \citep{rispoli_graph_2008}. 

\subsection{Convergence to exact gradients}
\label{sec:conv_exact_stat}

In this section, we conduct experiments on the convergence of the MCMC estimators to the exact corresponding expectation (that is, convergence of the stochastic gradient estimator to the true gradient). The estimators are defined as
\begin{align*}
\yt(\thetav) =\mathbb{E}_{{\pi}_{\bm{\theta}, t}}\left[Y\right]\approx \frac{1}{K- K_0}\sum_{k=K_0+1}^{K} \bm{y}^{(k)},
\end{align*}
where $\bm{y}^{(k)}$ is the $k$-th iterate of \cref{algo:SA} with maximization direction $\bm{\theta}$, final temperature $t$, and $K_0$ is a number of burn-in (or warm-up) iterations. The obtained estimator is compared to the exact expectation by performing marginal inference as described in \cref{sec:polytopes} (with a closed-form formula in the case of $\mathcal{Y}^d$, and by dynamic programming in the case of $\mathcal{Y}_\kappa^d$).\newline

\paragraph{Setup.} 

For $T>K_0$, let $\Tilde{\mathbb{E}}(\bm{\theta}, T) \coloneqq \frac{1}{T-K_0}\sum_{k=K_0+1}^{T}\bm{y}^{(k)}$ be the stochastic estimate of the expectation at step $T$. We proceed by first randomly generating $\bm{\Theta} \in \mathbb{R}^{M\times d}$, with $M$ being the number of instances, by sampling $\bm{\Theta}_{i, j}\sim \mathcal{N}(0, 1)$ independently. Then, we evaluate the impact of the following hyperparameters on the estimation of $\mathbb{E}_{{\pi}_{\bm{\Theta}_i}, t}\left[Y\right]$, for $i\in[M]$:
\begin{enumerate}
    \item $K_0$, the number of burn-in iterations,
    \item $t$, the temperature parameter,
    \item $C$, the number of parallel Markov chains.
\end{enumerate}

\paragraph{Metric.} 

The metric used is the squared Euclidean distance to the exact expectation, averaged on the $M$ instances
\begin{align*}
\frac{1}{M}\sum_{i=1}^M ||\mathbb{E}_{{\pi}_{\bm{\Theta}_i}, t}\left[Y\right] - \Tilde{\mathbb{E}}(\bm{\Theta}_i, T)||_2^2,
\end{align*}
which we measure for $T\in\llbracket K_0+1, K\rrbracket$.

\paragraph{Polytopes.} 

For the hypercube $\mathcal{Y}^d$ and its neighborhood system $\mathcal{N}_\leq^r$, we use $d=10$ and $r=1$, which gives $|\mathcal{Y}^d| = 2^{10}$ and $|\mathcal{N}_\leq^r(\bm{y})| = 10$. For the top-$\kappa$ polytope $\mathcal{Y}_\kappa^d$ and its neighborhood system $\mathcal{N}^s$, we use $d=10$, $\kappa = 3$ and $s=1$, which gives $|\mathcal{Y}_\kappa^d| = 120$ and $|\mathcal{N}^s(\bm{y})| = 30$. We also use a larger scale for both polytopes in order to highlight the varying impact of the temperature $t$ depending on the combinatorial size of the problem, in the second experiment. For the large scale, we use $d=1000$ and $r=10$ for the hypercube, which give $|\mathcal{Y}^d| = 2^{1000}\approx 10^{301}$ and $|\mathcal{N}_\leq^r(\bm{y})| \approx 2.7 \times 10^{23}$, and we use $d=1000$, $\kappa = 50$ and $s=10$ for the top-$\kappa$ polytope, which give $|\mathcal{Y}_\kappa^d| \approx 9.5\times10^{84}$ and $|\mathcal{N}^s(\bm{y})| \approx 1.6\times 10^{33}$.

\paragraph{Hyperparameters.} For each experiment, we use $K=3000$. We average over $M=1000$ problem instances for statistical significance. We use $K_0 = 0$, except for the first experiment, where it varies. We use a final temperature $t=1$, except for the second experiment, where it varies. We use an initial temperature $t_0 = t = 1$ (leading to a constant temperature schedule), except for the first experiment, where it depends on $K_0$. We use only one Markov chain and thus have $C=1$, except for the third experiment, where it varies.

\begin{figure}[htbp]
    \centering
    \begin{subfigure}[b]{0.49\textwidth}
        \centering
        \includegraphics[width=\textwidth]{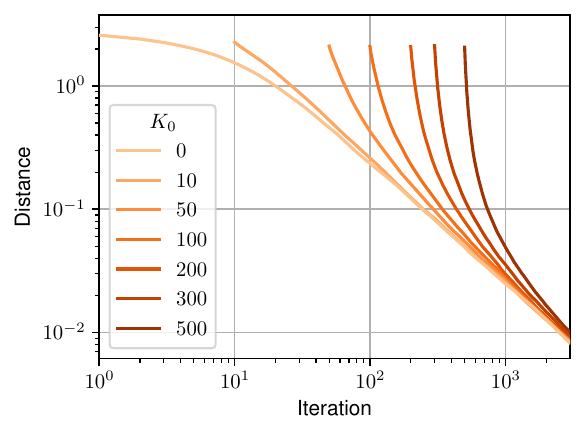}
        \caption{Hypercube}
        \label{fig:sa_cube_warmup_grad_convergence}
    \end{subfigure}
    \hfill
    \begin{subfigure}[b]{0.49\textwidth}
        \centering
        \includegraphics[width=\textwidth]{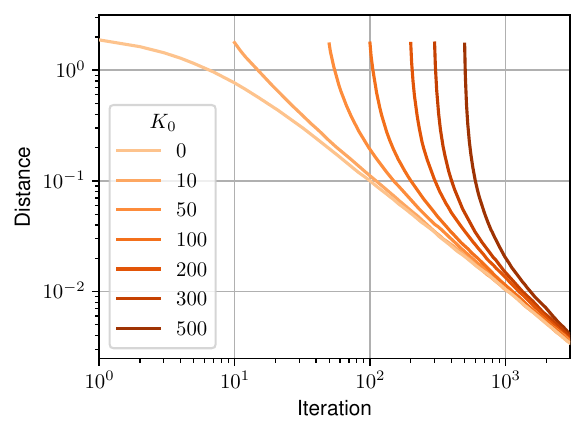}
        \caption{Top-$\kappa$ polytope}
        \label{fig:sa_topk_warmup_grad_convergence}
    \end{subfigure}
    \caption{Convergence to exact expectation on the hypercube and the top-$\kappa$ polytope, for varying number of burn-in iterations $K_0$. We conclude that burn-in is not beneficial to the estimation, and taking $K_0=0$ is the best option.}
    \label{fig:warmup_grad_convergence}
\end{figure}

\paragraph{(1) Impact of burn-in.} First, we evaluate the impact of $K_0$, the number of burn-in iterations. We use a truncated geometric cooling schedule $t_k=\max(\gamma^k\cdot t_0, \,t)$ with $\gamma = 0.995$. The initial temperature $t_0$ is set to $1/(\gamma^{K_0})$, so that $\forall k\geq K_0+1,\,t_{k} = t = 1$. The results are gathered in \cref{fig:warmup_grad_convergence}.

\begin{figure}[htbp]
    \centering
    \begin{subfigure}[b]{0.49\textwidth}
        \centering
        \includegraphics[width=\textwidth]{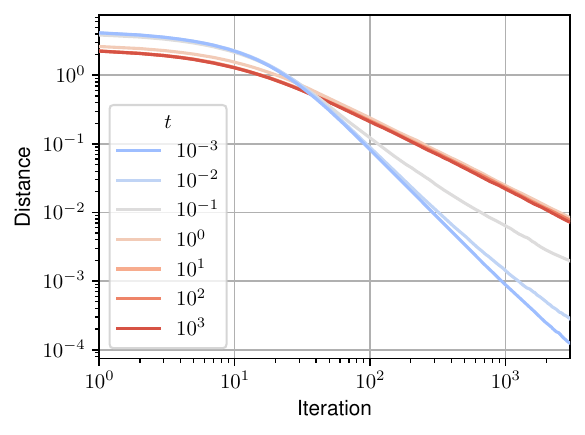}
        \caption{Hypercube}
        \label{fig:sa_cube_tmin_grad_convergence}
    \end{subfigure}
    \hfill
    \begin{subfigure}[b]{0.49\textwidth}
        \centering
        \includegraphics[width=\textwidth]{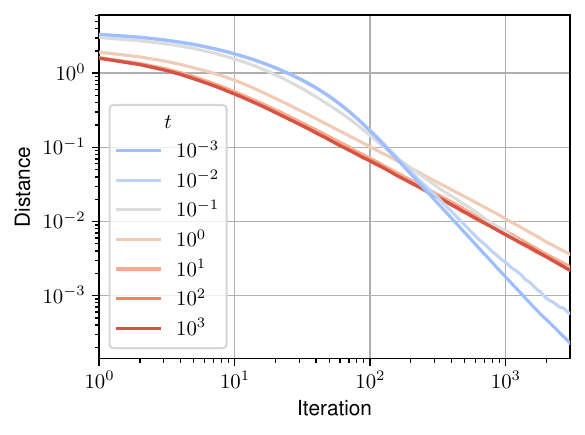}
        \caption{Top-$\kappa$ polytope}
        \label{fig:sa_topk_tmin_grad_convergence}
    \end{subfigure}
    \caption{Convergence to exact expectation on the hypercube and the top-$\kappa$ polytope, for varying final temperature $t$ (small scale experiment). We conclude that lower temperatures facilitate the estimation.}
    \label{fig:tmin_grad_convergence}
\end{figure}

\begin{figure}[htbp]
    \centering
    \begin{subfigure}[b]{0.49\textwidth}
        \centering
        \includegraphics[width=\textwidth]{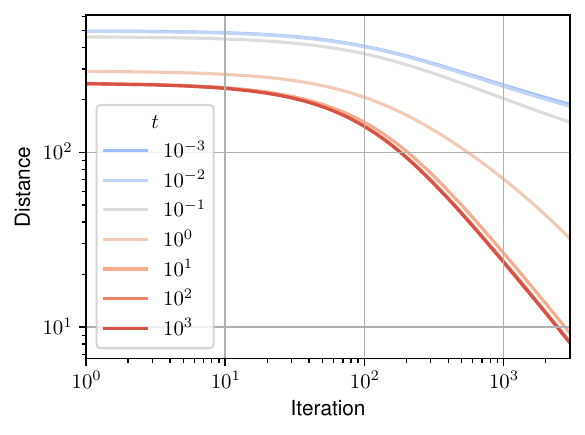}
        \caption{Hypercube}
        \label{fig:sa_cube_large_tmin_grad_convergence}
    \end{subfigure}
    \hfill
    \begin{subfigure}[b]{0.49\textwidth}
        \centering
        \includegraphics[width=\textwidth]{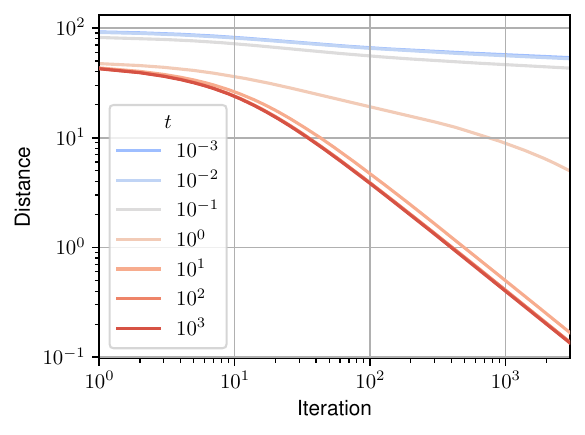}
        \caption{Top-$\kappa$ polytope}
        \label{fig:sa_topk_large_tmin_grad_convergence}
    \end{subfigure}
    \caption{Convergence to exact expectation on the hypercube and the top-$\kappa$ polytope, for varying final temperature $t$ (large scale experiment). Contrary to the small scale case, larger temperatures are beneficial to the estimation when the solution set is combinatorially large.}
    \label{fig:large_tmin_grad_convergence}
\end{figure}

\paragraph{(2) Impact of temperature.} We then evaluate the impact of the final temperature $t$ on the difficulty of the estimation task (different temperatures lead to different target expectations). The results for the small scale are gathered in \cref{fig:tmin_grad_convergence}, and the results for the large scale are gathered in \cref{fig:large_tmin_grad_convergence}.

\begin{figure}[htbp]
    \centering
    \begin{subfigure}[b]{0.49\textwidth}
        \centering
        \includegraphics[width=\textwidth]{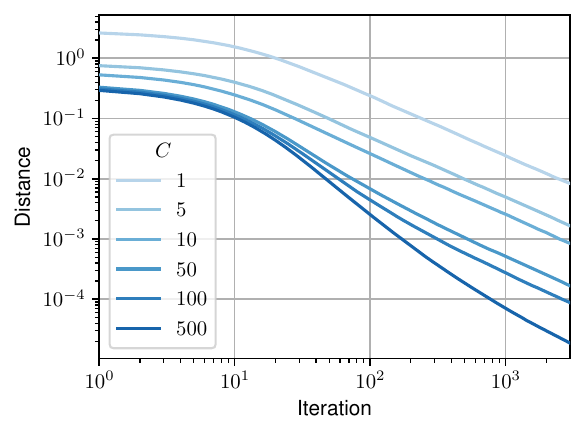}
        \caption{Hypercube}
        \label{fig:sa_cube_chains_grad_convergence}
    \end{subfigure}
    \hfill
    \begin{subfigure}[b]{0.49\textwidth}
        \centering
        \includegraphics[width=\textwidth]{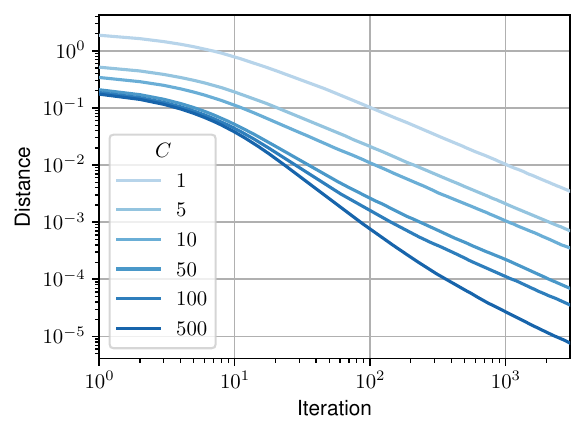}
        \caption{Top-$\kappa$ polytope}
        \label{fig:sa_topk_chains_grad_convergence}
    \end{subfigure}
    
    \caption{Convergence to exact expectation on the hypercube and the top-$\kappa$ polytope, for varying number of parallel Markov chains $C$. Running $10$ times more chains in parallel provides roughly the same benefit as extending each chain by $10$ times more iterations, highlighting the advantage of massively parallelized estimation.}
    \label{fig:chains_grad_convergence}
\end{figure}

\paragraph{(3) Impact of the number of parallel Markov chains.} Finally, we evaluate the impact of the number of parallel Markov chains $C$ on the estimation. The results are gathered in \cref{fig:chains_grad_convergence}.

\newpage

\subsection{Convergence to exact parameters}

In this section, we conduct experiments in the unconditional setting described in \cref{sec:unsup}. As a reminder, the empirical and expected risks $L_{N}$ and $L_{\thetav_0}$ are given by:
\begin{align}
L_{N}(\bm{\theta}; \y_1, \dots, \y_N) &\coloneqq \frac{1}{N}\sum_{i=1}^N \ell_t\left(\bm{\theta};\, \y_i\right) \notag\\
&= A_t(\bm{\theta}) + \frac{1}{N}\sum_{i=1}^N\Omega_t(\y_i) - \langle \bm{\theta},\,\bar{Y}_N\rangle \notag\\
&= \ell_t(\bm{\theta};\, \bar{Y}_N) + C_1(Y)\label{eq:emp_risk_unsup},
\end{align}

and
\begin{align}
L_{\bm{\theta}_0}(\bm{\theta}) &\coloneqq \mathbb{E}_{\left(\y_i\right)_{i=1}^N\sim({\pi}_{\bm{\theta}_0,t})^{\otimes N}}\left[L_{N}(\thetav; \y_1, \dots, \y_N)\right] \notag\\
&= A_t(\bm{\theta}) + \mathbb{E}_{{\pi}_{\bm{\theta}_0,t}}\left[\Omega_t(Y)\right] - \langle \bm{\theta},\,\yt(\bm{\theta}_0)\rangle \notag\\
&= \ell_t(\bm{\theta};\, \yt(\bm{\theta}_0)) + C_2(\bm{\theta}_0)\label{eq:true_risk_unsup},
\end{align}

\noindent where the constants $C_1(Y) \!=\! \frac{1}{N}\!\!\sum_{i=1}^N\!\Omega_t(\y_i) \!- \!\Omega_t(\bar{Y}_N)$ 
and $C_2(\bm{\theta}_0)\! =\! \mathbb{E}_{{\pi}_{\bm{\theta}_0,t}}\left[\Omega_t(Y)\right] \!-\! \Omega_t\left(\yt(\bm{\theta}_0)\right)$ do not depend on $\bm{\theta}$. As Jensen gaps, they are non-negative by convexity of $\Omega_t$.

\paragraph{2D visualization.} 

\begin{figure}[h]
    \centering
    \includegraphics[width=0.7\linewidth]{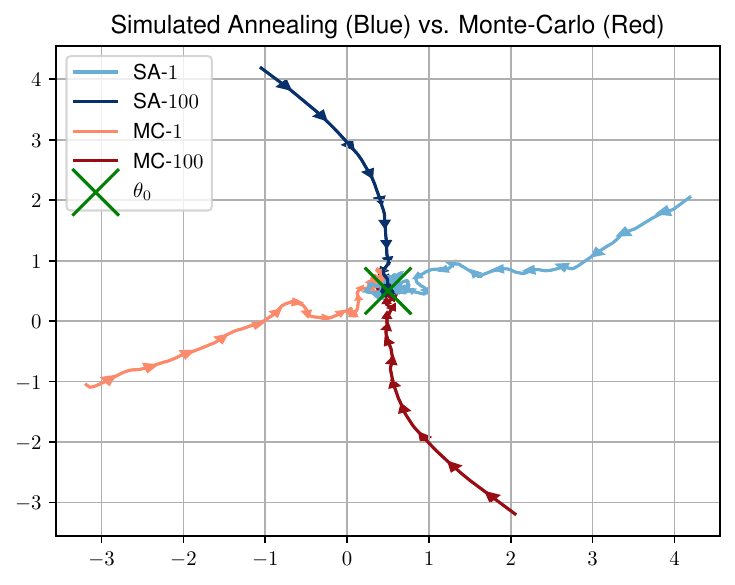}
    \caption{Comparison of stochastic gradient trajectories for a SA / M-H oracle on $\mathcal{Y}^2$ and unbiased stochastic gradients obtained via Monte Carlo sampling. Increasing the number of Markov chain iterations yields smoother trajectories, similar to the effect of using more Monte Carlo samples in the case of perturbation-based methods \citep{berthet_learning_2020}.}
    \label{fig:2d_traj}
\end{figure}

As an introductory example, we display stochastic gradient trajectories in \cref{fig:2d_traj}. The parameter $\bm{\theta}\in\mathbb{R}^d$ is updated following \cref{eq:sgd_mcmc} to minimize the expected risk $L_{\bm{\theta}_0}$ defined in \cref{eq:true_risk_unsup}, with $\bm{\theta}_0=(1/2,\,1/2)$. The polytope used is the $2$-dimensional hypercube $\mathcal{Y}^2$, with neighborhood graph $\mathcal{N}_1$ (neighbors are adjacent vertices of the square). We present trajectories obtained using MCMC-sampled gradients, comparing results from both $1$ and $100$ Markov chain iterations with \cref{algo:SA}. For comparison, we include trajectories obtained using Monte Carlo-sampled (i.e., unbiased) gradients, using 1 and 100 samples.

\paragraph{General setup.} 

We proceed by first randomly generating true parameters $\bm{\Theta}_0 \in \mathbb{R}^{M\times d}$, with $M$ being a number of problem instances we average on (in order to reduce noise in our observations), by sampling $\bm{\Theta}_{i, j}\sim \mathcal{N}(0, 1)$ independently. The goal is to learn each parameter vector $(\bm{\Theta}_0)_i\in\mathbb{R}^d,i\in[M]$, as $M$ independent problems. The model is randomly initialized at $\hat{\bm{\Theta}}_0$, and updated with Adam \citep{kingma_adam_2017} to minimize the loss. In order to better separate noise due to the optimization process and noise due to the sampling process, we use the expected risk $L_{(\bm{\Theta}_0)_i}$ for general experiments, and use the empirical risk $L_N$ only when focusing on the impact of the dataset size $N$. In this case, we create a dataset $Y \in \mathbb{R}^{M\times N\times d}$, with $N$ being the number of samples, by sampling independently $Y_{i, j} \sim  {\pi}_{(\bm{\Theta}_0)_i},\;\forall i\in[M],\,\forall j\in[N]$.

We study the impact of the following hyperparameters on learning:

\begin{enumerate}
    \item $K$, the number of Markov chain iterations,
    \item $C$, the number of parallel Markov chains,
    \item the initialization method used for the chains (either random, persistent, or data-based),
    \item $N$, the number of samples in the dataset.
\end{enumerate}

\paragraph{Metrics.} The first metric used is the objective function actually minimized, i.e., the expected risk, averaged on the $M$ instances:

\begin{align*}
\frac{1}{M}\sum_{i=1}^M L_{(\bm{\Theta}_0)_i}((\hat{\bm{\Theta}}_n)_i),
\end{align*}

\noindent where $(\hat{\bm{\Theta}}_n)_i$ is the $n$-th iterate of the optimization process for the problem instance $i\in[M]$. We measure this loss for $n\in[n_\text{max}]$, with $n_\text{max}$ the total number of gradient iterations. For the fourth experiment, where we evaluate the impact of the number of samples $N$, we measure instead the empirical risk:

\begin{align*}
\frac{1}{M}\sum_{i=1}^M L_N((\hat{\bm{\Theta}}_n)_i\,;\,Y_{i, 1},\dots Y_{i,N})
\end{align*}

In both cases, the best loss value that can be reached is positive but cannot be computed: it corresponds to the constants $C_1$ and $C_2$ in \cref{eq:emp_risk_unsup} and \cref{eq:true_risk_unsup}. Thus, we also provide ``stretched'' figures, where we plot the loss minus the best loss found during the optimization process. \newline

The second metric used is the squared euclidean distance of the estimate to the true parameter, also averaged on the $M$ instances:

\begin{align*}
\frac{1}{M}\sum_{i=1}^M ||(\bm{\Theta}_0)_i - (\hat{\bm{\Theta}}_n)_i||_2^2.
\end{align*}

As the top-$\kappa$ polytope is of dimension $d-1$, the model is only specified up to vectors orthogonal to the direction of the smallest affine subspace it spans. Thus, in this case, we measure instead:

\begin{align*}
\frac{1}{M}\sum_{i=1}^M ||\Proj_D\left((\bm{\Theta}_0)_i\right) - \Proj_D\left((\hat{\bm{\Theta}}_n)_i\right)||_2^2,
\end{align*}

\noindent where $\Proj_D$ is the orthogonal projector on the hyperplane $D = \{\bm{x}\in\mathbb{R}^d:\,\langle \bm{1},\,\bm{x}\rangle = 0\}$, which is the corresponding direction.

\paragraph{Polytopes.} For the hypercube $\mathcal{Y}^d$ and its neighborhood system $\mathcal{N}_\leq^r$, we use $d=10$ and $r=1$, except in the fifth experiment, where we use a mixture of $\mathcal{N}_=^r$ neighborhoods (detailed below). For the top-$\kappa$ polytope $\mathcal{Y}_\kappa^d$ and its neighborhood system $\mathcal{N}^s$, we use $d=10$, $\kappa = 3$ and $s=1$.

\paragraph{Hyperparameters.} For each experiment, we perform $1000$ gradient steps. We use $K_0 = 0$, final temperature $t=1$ and initial temperature $t_0 = t = 1$ (leading to a constant temperature schedule). We use $K = 1000$ Markov chain iterations, except in the first experiment, where it varies. We use only one Markov chain and thus have $C=1$, except for the second experiment, where it varies. We use a persistent initialization method for the Markov chains, except in the third experiment, where we compare the three different methods. For statistical significance, we average over $M=100$ problem instances for each experiment, except in the third experiment, where we use $M=1000$. We work in the limit case $N\rightarrow\infty$, except in the fourth experiment, where $N$ varies.

\begin{figure}[htbp]
    \centering
    \begin{subfigure}[b]{0.49\textwidth}
        \centering
        \includegraphics[width=\textwidth]{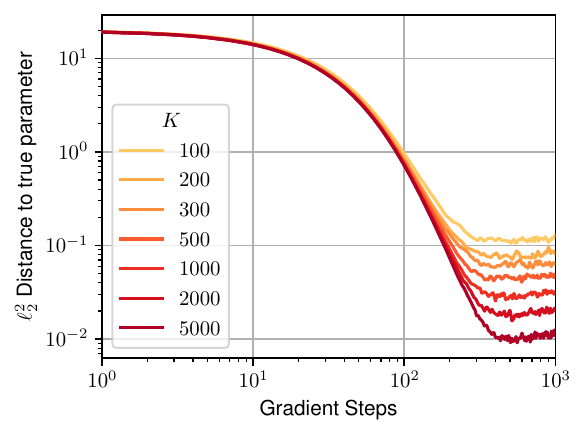}
        \caption{Distance to true parameter, hypercube}
        \label{fig:sa_cube_inner_unsupervised_distance}
    \end{subfigure}
    \hfill
    \begin{subfigure}[b]{0.49\textwidth}
        \centering
        \includegraphics[width=\textwidth]{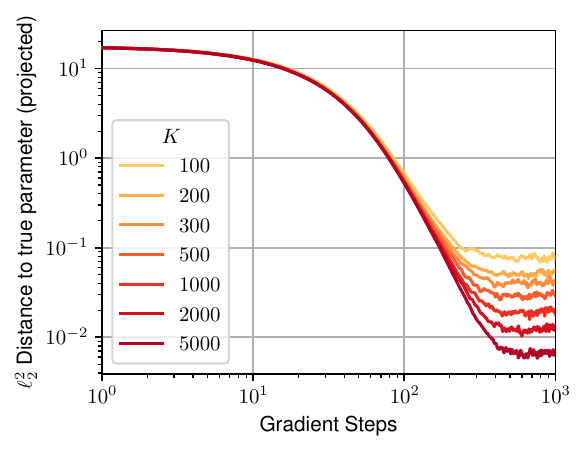}
        \caption{Distance to true parameter, top-$\kappa$ polytope}
        \label{fig:sa_topk_inner_unsupervised_distance}
    \end{subfigure}
    \vfill
    \begin{subfigure}[b]{0.49\textwidth}
        \centering
        \includegraphics[width=\textwidth]{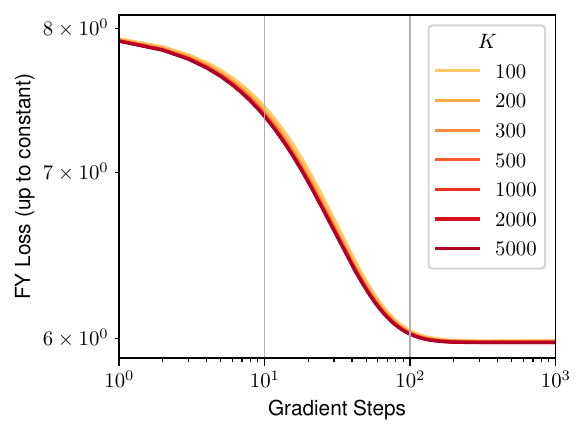}
        \caption{FY loss (up to constant), hypercube}
        \label{fig:sa_cube_inner_unsupervised_loss}
    \end{subfigure}
    \hfill
    \begin{subfigure}[b]{0.49\textwidth}
        \centering
        \includegraphics[width=\textwidth]{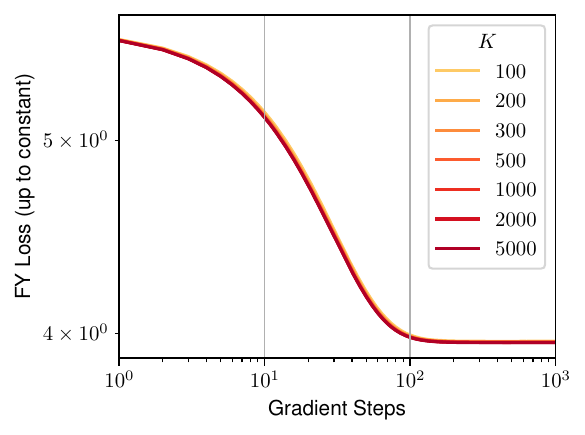}
        \caption{FY loss (up to constant), top-$\kappa$ polytope}
        \label{fig:sa_topk_inner_unsupervised_loss}
    \end{subfigure}
    \vfill
    \begin{subfigure}[b]{0.49\textwidth}
        \centering
        \includegraphics[width=\textwidth]{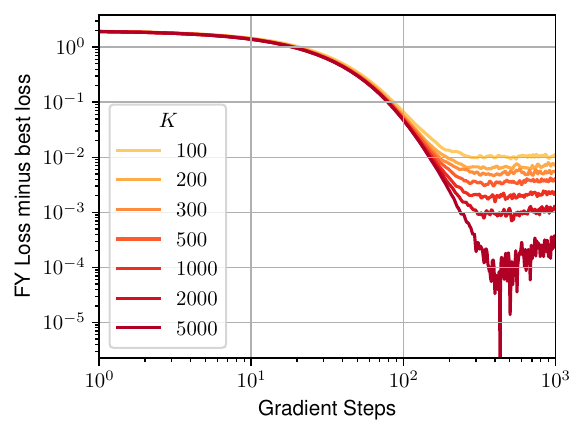}
        \caption{FY loss minus best loss, hypercube}
        \label{fig:sa_cube_inner_unsupervised_loss_stretched}
    \end{subfigure}
    \hfill
    \begin{subfigure}[b]{0.49\textwidth}
        \centering
        \includegraphics[width=\textwidth]{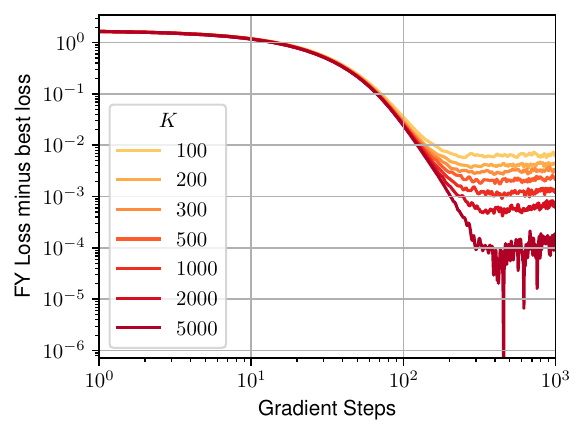}
        \caption{FY loss minus best loss, top-$\kappa$ polytope}
        \label{fig:sa_topk_inner_unsupervised_loss_stretched}
    \end{subfigure}
    \caption{Convergence to the true parameter on the hypercube (left) and the top-$\kappa$ polytope (right), for varying number of Markov chain iterations $K$. Longer chains improve learning.}
    \label{fig:kmax_unsupervised}
\end{figure}

\paragraph{(1) Impact of the length of Markov chains.} First, we evaluate the impact of $K$, the number of inner iterations, i.e., the length of each Markov chain. The results are gathered in \cref{fig:kmax_unsupervised}.

\begin{figure}[htbp]
    \centering
    \begin{subfigure}[b]{0.49\textwidth}
        \centering
        \includegraphics[width=\textwidth]{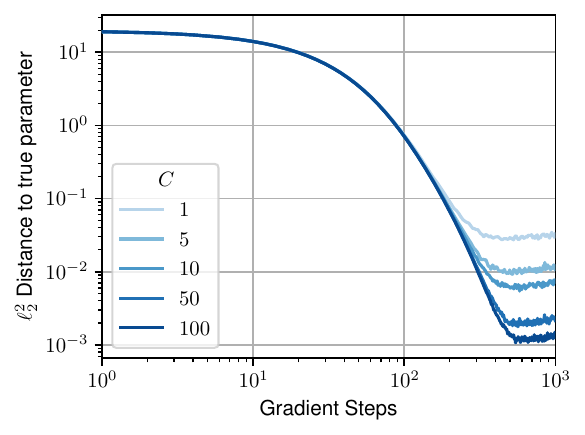}
        \caption{Distance to true parameter, hypercube}
        \label{fig:sa_cube_chains_unsupervised_distance}
    \end{subfigure}
    \hfill
    \begin{subfigure}[b]{0.49\textwidth}
        \centering
        \includegraphics[width=\textwidth]{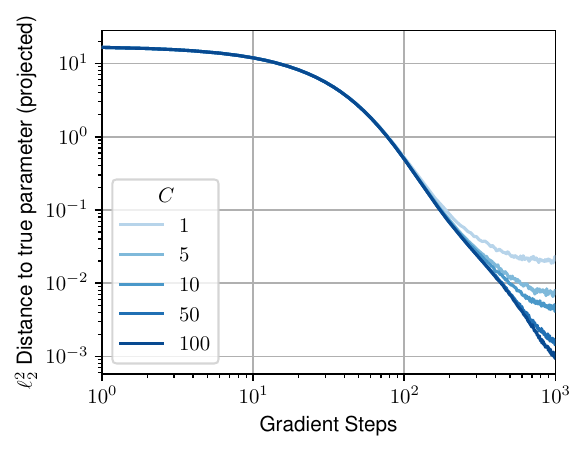}
        \caption{Distance to true parameter, top-$\kappa$ polytope}
        \label{fig:sa_topk_chains_unsupervised_distance}
    \end{subfigure}
    \vfill
    \begin{subfigure}[b]{0.49\textwidth}
        \centering
        \includegraphics[width=\textwidth]{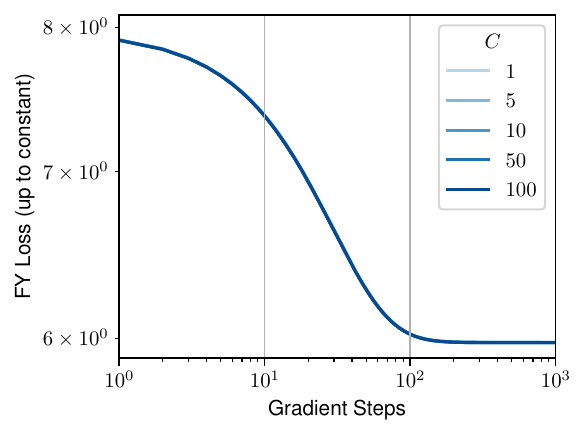}
        \caption{FY loss (up to constant), hypercube}
        \label{fig:sa_cube_chains_unsupervised_loss}
    \end{subfigure}
    \hfill
    \begin{subfigure}[b]{0.49\textwidth}
        \centering
        \includegraphics[width=\textwidth]{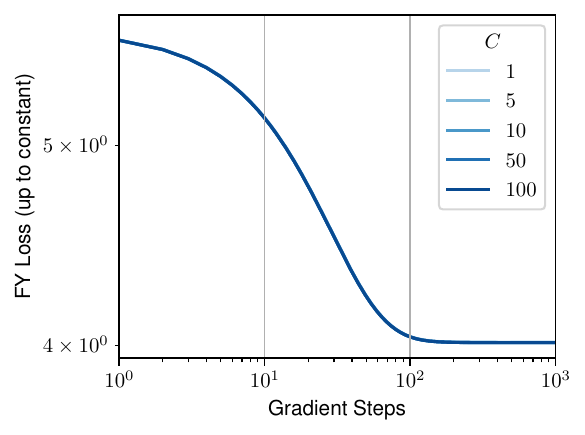}
        \caption{FY loss (up to constant), top-$\kappa$ polytope}
        \label{fig:sa_topk_chains_unsupervised_loss}
    \end{subfigure}
    \vfill
    \begin{subfigure}[b]{0.49\textwidth}
        \centering
        \includegraphics[width=\textwidth]{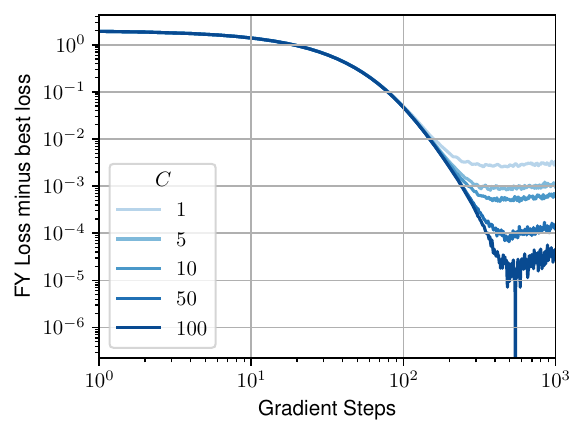}
        \caption{FY loss minus best loss, hypercube}
        \label{fig:sa_cube_chains_unsupervised_loss_stretched}
    \end{subfigure}
    \hfill
    \begin{subfigure}[b]{0.49\textwidth}
        \centering
        \includegraphics[width=\textwidth]{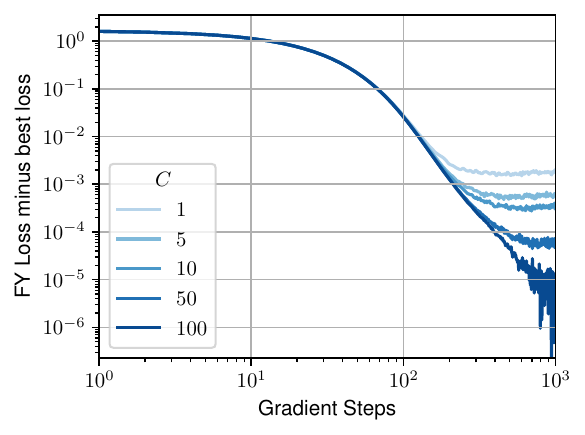}
        \caption{FY loss minus best loss, top-$\kappa$ polytope}
        \label{fig:sa_topk_chains_unsupervised_loss_stretched}
    \end{subfigure}
    \caption{Convergence to the true parameter on the hypercube (left) and the top-$\kappa$ polytope (right), for varying number of parallel Markov chains $C$. Adding Markov chains improves estimation.}
    \label{fig:c_unsupervised}
\end{figure}

\paragraph{(2) Impact of the number of parallel Markov chains.} We now evaluate the impact of the number of Markov chains $C$ run in parallel to perform each gradient estimation on the learning process. The results are gathered in \cref{fig:c_unsupervised}.

\begin{figure}[htbp]
    \centering
    \begin{subfigure}[b]{0.49\textwidth}
        \centering
        \includegraphics[width=\textwidth]{unsupervised_toy/sa_init_cube_distances.pdf}
        \caption{Distance to true parameter, hypercube}
        \label{fig:sa_cube_init_unsupervised_distance}
    \end{subfigure}
    \hfill
    \begin{subfigure}[b]{0.49\textwidth}
        \centering
        \includegraphics[width=\textwidth]{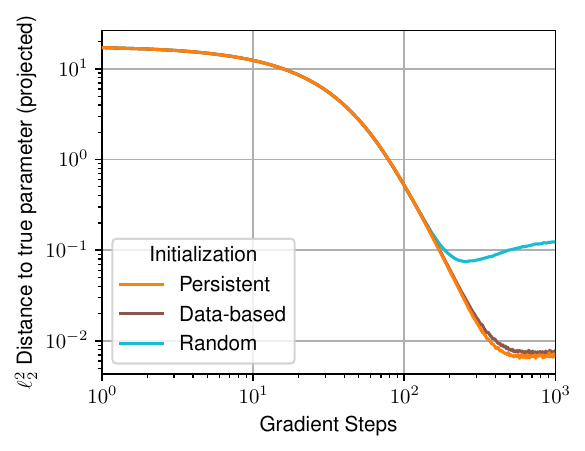}
        \caption{Distance to true parameter, top-$\kappa$ polytope}
        \label{fig:sa_topk_init_unsupervised_distance}
    \end{subfigure}
    \vfill
    \begin{subfigure}[b]{0.49\textwidth}
        \centering
        \includegraphics[width=\textwidth]{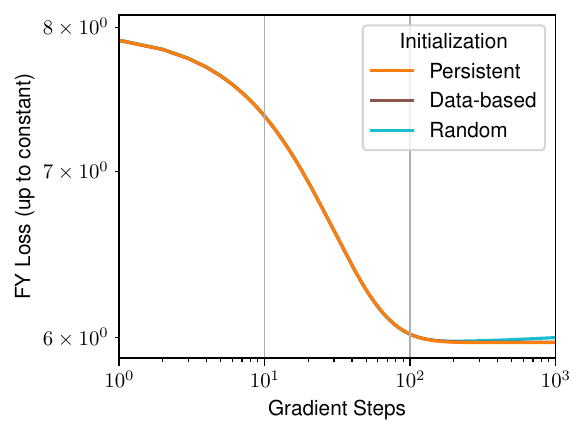}
        \caption{FY loss (up to constant), hypercube}
        \label{fig:sa_cube_init_unsupervised_loss}
    \end{subfigure}
    \hfill
    \begin{subfigure}[b]{0.49\textwidth}
        \centering
        \includegraphics[width=\textwidth]{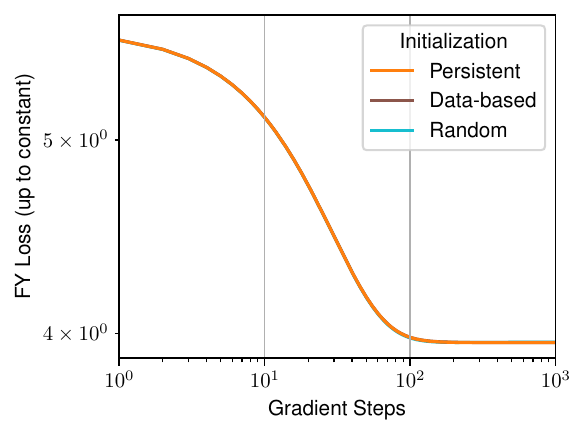}
        \caption{FY loss (up to constant), top-$\kappa$ polytope}
        \label{fig:sa_topk_init_unsupervised_loss}
    \end{subfigure}
    \vfill
    \begin{subfigure}[b]{0.49\textwidth}
        \centering
        \includegraphics[width=\textwidth]{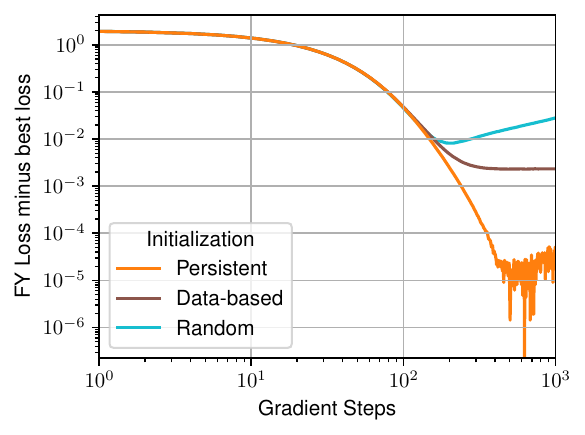}
        \caption{FY loss minus best loss, hypercube}
        \label{fig:sa_cube_init_unsupervised_loss_stretched}
    \end{subfigure}
    \hfill
    \begin{subfigure}[b]{0.49\textwidth}
        \centering
        \includegraphics[width=\textwidth]{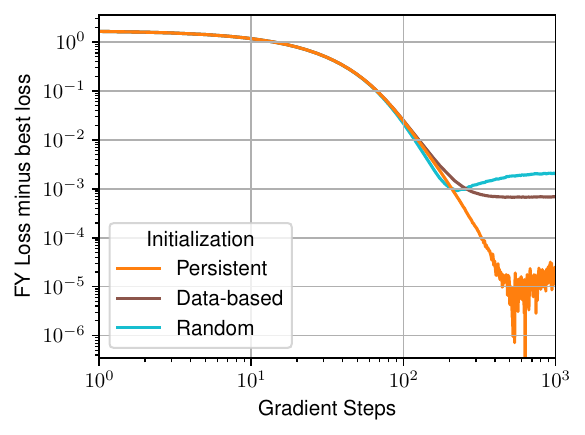}
        \caption{FY loss minus best loss, top-$\kappa$ polytope}
        \label{fig:sa_topk_init_unsupervised_loss_stretched}
    \end{subfigure}

    \caption{Convergence to the true parameter on the hypercube (left) and the top-$\kappa$ polytope (right), for varying Markov chain initialization method. The persistent and data-based initialization methods significantly outperform the random initialization method.}
    \label{fig:init_unsupervised}
\end{figure}

\paragraph{(3) Impact of the initialization method.} Then, we evaluate the impact of the method to initialize each Markov chain used for gradient estimation. The persistent method consists in setting $\bm{y}^{(n+1, 0)} = \bm{y}^{(n, K)}$, the data-based method consists in setting $\bm{y}^{(n+1, 0)} = \y_i$ with $i\sim\mathcal{U}([N])$, and the random method consists in setting $\bm{y}^{(n+1, 0)} \sim \mathcal{U}(\mathcal{Y})$ (see \cref{sec:cd_initialization} and \cref{tab:init} for a detailed explanation). The results are gathered in \cref{fig:init_unsupervised}.

\begin{figure}[htbp]
    \centering
    \begin{subfigure}[b]{0.49\textwidth}
        \centering
        \includegraphics[width=\textwidth]{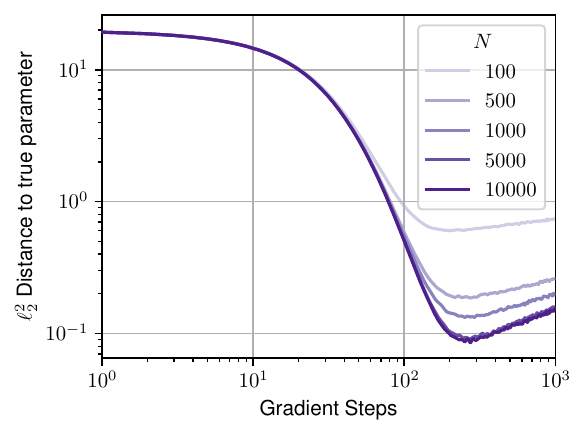}
        \caption{Distance to true parameter}
        \label{fig:sa_cube_samples_unsupervised_distance}
    \end{subfigure}
    \hfill
    \begin{subfigure}[b]{0.49\textwidth}
        \centering
        \includegraphics[width=\textwidth]{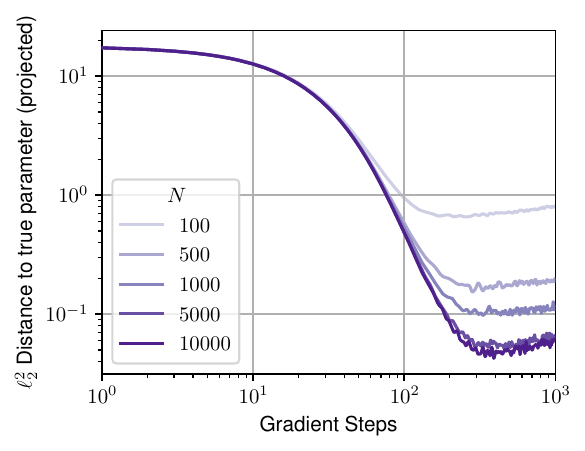}
        \caption{Distance to true parameter}
        \label{fig:sa_topk_samples_unsupervised_distance}
    \end{subfigure}
    \vfill
    \begin{subfigure}[b]{0.49\textwidth}
        \centering
        \includegraphics[width=\textwidth]{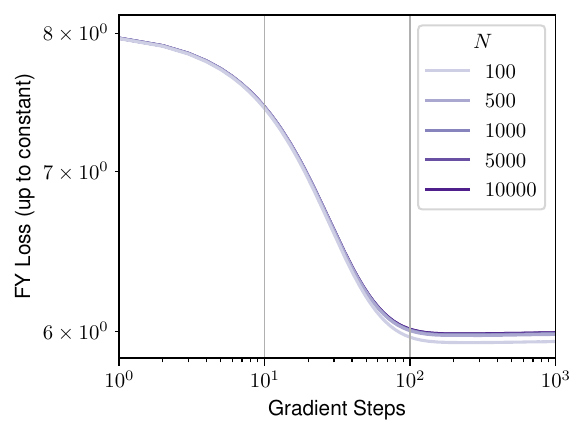}
        \caption{FY loss (up to constant)}
        \label{fig:sa_cube_samples_unsupervised_loss}
    \end{subfigure}
    \hfill
    \begin{subfigure}[b]{0.49\textwidth}
        \centering
        \includegraphics[width=\textwidth]{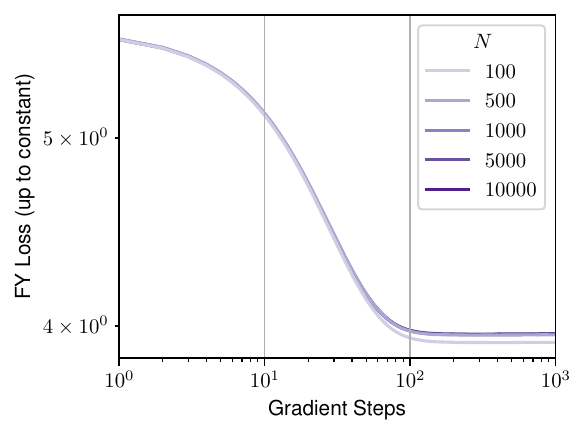}
        \caption{Fenchel-Young loss (up to constant)}
        \label{fig:sa_topk_samples_unsupervised_loss}
    \end{subfigure}
    \vfill
    \begin{subfigure}[b]{0.49\textwidth}
        \centering
        \includegraphics[width=\textwidth]{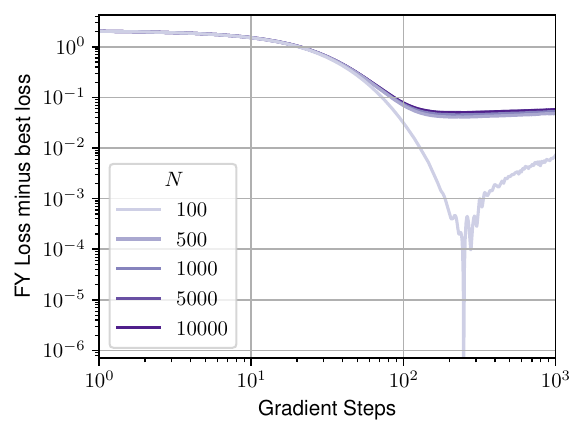}
        \caption{FY loss minus best loss}
        \label{fig:sa_cube_samples_unsupervised_loss_stretched}
    \end{subfigure}
    \hfill
    \begin{subfigure}[b]{0.49\textwidth}
        \centering
        \includegraphics[width=\textwidth]{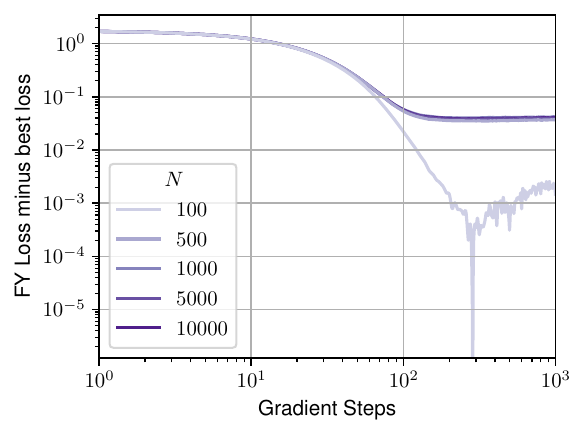}
        \caption{Fenchel-Young loss minus best loss}
        \label{fig:sa_topk_samples_unsupervised_loss_stretched}
    \end{subfigure}
    \caption{Convergence to the true parameter on the hypercube (left) and the top-$\kappa$ polytope (right), for varying number of samples $N$ in the dataset. As the dataset is different for each task, the empirical risk $L_N$, which is the minimized objective function (contrary to other experiments, where we minimize $L_{\bm{\theta}_0}$), also varies. Although empirical risks associated to smaller datasets appear easier to minimize, increasing the dataset size reduces the bias and thus the distance to $\bm{\theta}_0$, as expected.}
    \label{fig:samples_unsupervised}
\end{figure}

\paragraph{(4) Impact of the dataset size.} We now evaluate the impact of the number of samples $N$ from $\pi_{\bm{\theta}_0}$ (i.e., the size of the dataset $\left(\y_i\right)_{i=1}^N$) on the estimation of the true parameter $\bm{\theta}_0$. The results are gathered in \cref{fig:samples_unsupervised}.

\begin{figure}[htbp]
    \centering
    \begin{subfigure}[b]{0.49\textwidth}
        \centering
        \includegraphics[width=\textwidth]{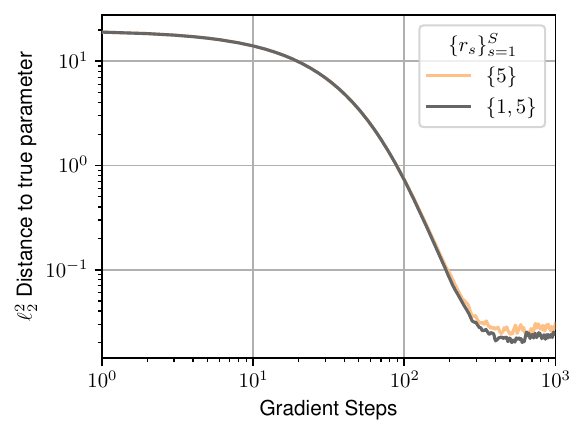}
        \caption{Distance to true parameter, $r_s\in\{5\}$ or $\{1,5\}$}
        \label{fig:sa_mixture_15_unsupervised_distance}
    \end{subfigure}
    \hfill
    \begin{subfigure}[b]{0.49\textwidth}
        \centering
        \includegraphics[width=\textwidth]{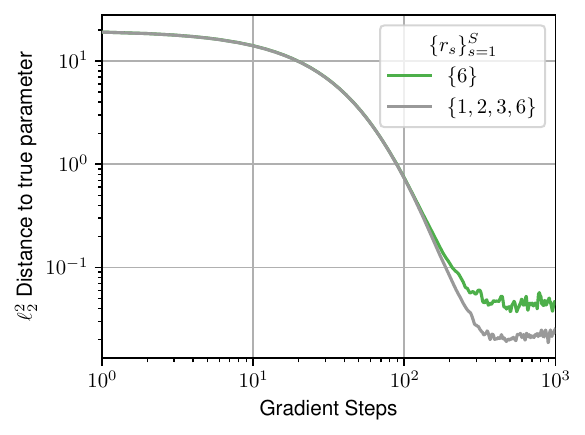}
        \caption{Distance to true parameter, $r_s\in\{6\}$ or $\{1,2,3,6\}$}
        \label{fig:sa_mixture_1236_unsupervised_distance}
    \end{subfigure}
    \vfill
    \begin{subfigure}[b]{0.49\textwidth}
        \centering
        \includegraphics[width=\textwidth]{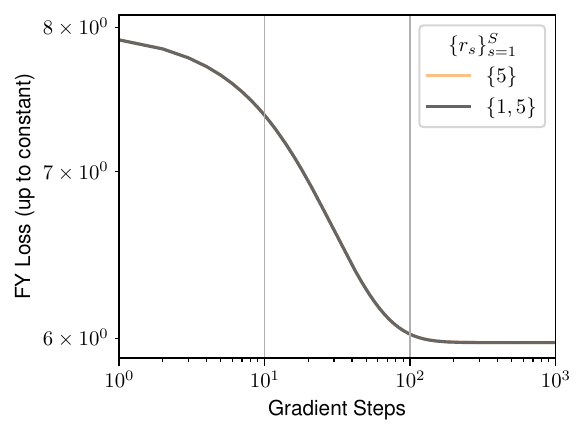}
        \caption{FY loss (up to constant), $r_s\in\{5\}$ or $\{1,5\}$}
        \label{fig:sa_mixture_15_unsupervised_loss}
    \end{subfigure}
    \hfill
    \begin{subfigure}[b]{0.49\textwidth}
        \centering
        \includegraphics[width=\textwidth]{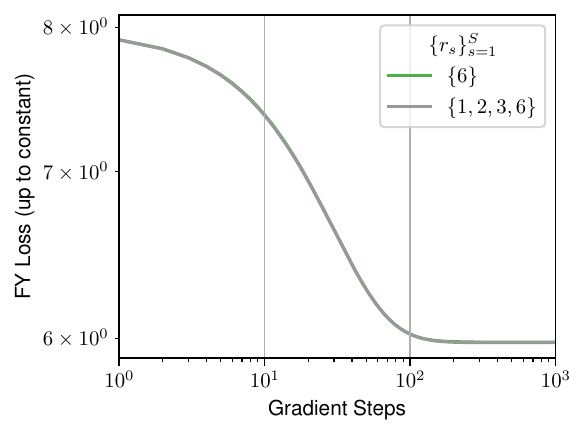}
        \caption{FY loss (up to constant), $r_s\in\{6\}$ or $\{1,2,3,6\}$}
        \label{fig:sa_mixture_1236_unsupervised_loss}
    \end{subfigure}
    \vfill
    \begin{subfigure}[b]{0.49\textwidth}
        \centering
        \includegraphics[width=\textwidth]{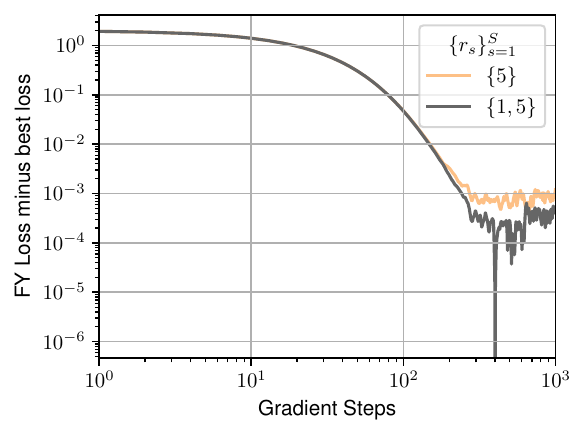}
        \caption{FY loss minus best loss, $r_s\in\{5\}$ or $\{1,5\}$}
        \label{fig:sa_mixture_15_unsupervised_loss_stretched}
    \end{subfigure}
    \hfill
    \begin{subfigure}[b]{0.49\textwidth}
        \centering
        \includegraphics[width=\textwidth]{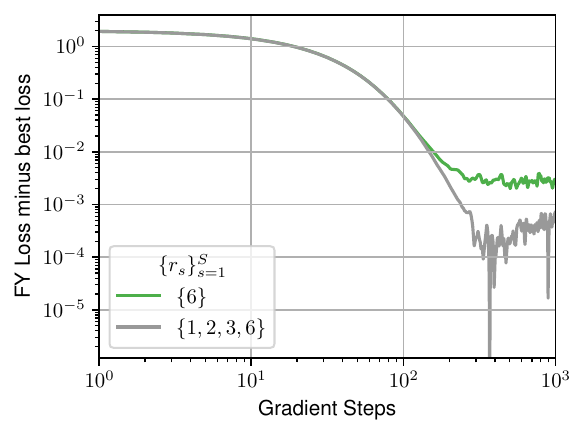}
        \caption{FY loss minus best loss, $r_s\in\{6\}$ or $\{1,2,3,6\}$}
        \label{fig:sa_mixture_1236_unsupervised_loss_stretched}
    \end{subfigure}
    \caption{Convergence to the true parameter on the hypercube, with different mixtures of neighborhood systems $\{\mathcal{N}_=^{r_s}\}_{s=1}^S$: comparing $r_s\in\{5\}$ to $r_s\in\{1,5\}$ (left), and comparing $r_s\in\{6\}$ to $r_s\in\{1,2,3,6\}$ (right). Using more neighborhoods in the mixture improves learning.}
    \label{fig:mixture_unsupervised}
\end{figure}

\paragraph{(5) Impact of neigborhood mixtures.} Finally, we evaluate the impact of the use of neighborhood mixtures. To do so, we use mixtures $\{ \mathcal{N}_=^{r_s} \}_{s=1}^S$, once with $\{r_s \}_{s=1}^S = \{5\}$ opposed to $\{r_s \}_{s=1}^S = \{1,5\}$, and once with $\{r_s \}_{s=1}^S = \{6\}$ (which gives a reducible Markov chain as $6$ is even, so that the individual neighborhood graph $\mathcal{N}_=^6$ is not connected, and has to connected components) opposed to $\{r_s \}_{s=1}^S = \{1,2,3,6\}$. The results are gathered in \cref{fig:mixture_unsupervised}.

\newpage

\section{Additional material}

\subsection{Fenchel-Young loss for $K=1$ in the unconditional setting}
This proposition is analogous to \cref{prop:1step_fyl_sup}, but in the unconditional setting, when using a data-based initialization method – i.e., the original CD initialization scheme, without persistent Markov chains. See \cref{sec:cd_initialization} for a detailed discussion about this.
\begin{proposition}
\label{prop:1step_fyl_unsup}
    Let $\pfirstunsup$ denote the distribution of the first iterate of the Markov chain defined by the Markov transition kernel given in \cref{eq:markov_kernel}, with proposal distribution $q$ and initialized by $\y^{(0)} = \y_i$, with $i\sim\mathcal{U}(\llbracket 1, N\rrbracket )$. There exists a dataset-dependent regularization $\Omega_{\bar{Y}_N}$ with the following properties: $\Omega_{\bar{Y}_N}$ is $\left(t\cdot N/\sum_{i=1}^N\mathbb{E}_{q(\y_i,\,\cdot\,)}||Y-\y_i||_2^2\right)$-strongly convex; it is such that:
    \begin{align*}
        \mathbb{E}_{\pfirstunsup}\left[Y\right] = \argmax_{\bm{\mu}\in\conv\left(\bigcup_{i=1}^N\left\{\mathcal{N}(\y_i)\cup\{\y_i\}\right\}\right)}\left\{\langle\bm{\theta},\bm{\mu}\rangle-\Omega_{\bar{Y}_N}(\bm{\mu})\right\};
    \end{align*}
    and the Fenchel-Young loss $ \ell_{\Omega_{\bar{Y}_N}}$ generated by $\Omega_{\bar{Y}_N}$ is $\frac{1}{N}\sum_{i=1}^N\mathbb{E}_{q(\y_i,\,\cdot\,)}||Y-\y_i||_2^2/t$-smooth in its first argument, and such that $\nabla_{\bm{\theta}} \ell_{\Omega_{\bar{Y}_N}}(\bm{\theta}\,;\y) = \mathbb{E}_{\pfirstunsup}\left[Y\right] - \y$.
\end{proposition}

The proof is given in \cref{proof:1step_fyl_unsup}. 

\subsection{Properties of the expected first iterate}

\begin{proposition}
\label{prop:one_step_properties}
    Let $\thetav\in\RR^d$, $\y\in\cY$. Let \begin{align*}
        \cN_{\text{better}}(\y)\coloneqq\left\{ \y'\in\cN(\y)\mid \langle\thetav,\y'\rangle+\varphi(\y')>\langle\thetav,\y\rangle+\varphi(\y) \right\}
    \end{align*} denote the set of improving neighbors of $\y$ for the unregularized objective function. We have the following properties:
    \begin{align*}
    \EE_{\pfirstsup}[Y]&\in \conv\left(\mathcal{N}(\y)\cup\{\y\}\right),\\
    \EE_{\pfirstsup}[Y]&\xrightarrow[t\rightarrow 0^+]{} \y+ \!\!\!\!\!\sum_{\y'\in\cN_{\text{better}}(\y)}q(\y,\y')\cdot(\y'-\y), \\
    \text{and}\quad  \EE_{\pfirstsup}[Y]&\xrightarrow[t\rightarrow \infty]{}\y + \!\!\!\sum_{\y'\in\cN(\y)}\min\left[q(\y,\y'),q(\y',\y)\right]\cdot(\y'-\y).
    \end{align*}
\end{proposition}

The proof is given in \cref{proof:one_step_properties}. Thus, as the set $\cN_{\text{better}}$ is defined according the value of the original, unregularized objective function $\y\mapsto\langle\thetav, \y\rangle+\varphi(\y)$, the low temperature behavior of the regularized maximizer $\EE_{\pfirstsup}[Y]$ effectively reflects the fact that the regularization function $\Omega_\y$ extends the influence of $\varphi$ from the vertices $\cN(\y)\cup\{\y\}$ to their convex hull. 

\subsection{Markov chain initialization}

\label{sec:cd_initialization}

In contrastive divergence (CD) learning, the intractable expectation in the log-likelihood gradient is approximated by short-run MCMC, initialized at the data distribution \citep{hinton_training_2000} (using a Gibbs sampler in the setting of Restricted Boltzmann Machines).

Here, we write the gradient at the $n$-th iteration of gradient descent in the conditional setting as:
\begin{align*}
    \nabla_W L_{N}(\hat{W}_n) \approx \frac{1}{|B_n|}\sum_{i\in B_n} J_W g_{\hat{W}_n} (\bm{x}_i) \left( \frac{1}{K}\sum_{k=1}^{K} \bm{y}_i^{(n+1,\, k)} \!\! - \y_i\right),
\end{align*}
\noindent with $B_n$ being the mini-batch (or full batch) used at iteration $n$, $\y_i$ the ground-truth structure associated to $\bm{x}_i$ in the dataset, and $\bm{y}_i^{(n+1,\, k)}$ the $k$-th iterate of \cref{algo:SA}, with maximization direction $g_{\hat{W}_n}(\bm{x}_i)$, and initialization point $\bm{y}_i^{(n+1,\, 0)}$. We also write:
\begin{align*}
    \nabla_{\bm{\theta}} L_N(\hat{\bm{\theta}}_n) \approx  \frac{1}{K}\sum_{k=1}^{K}\bm{y}^{(n+1,\,k)} -\bar{Y}_N
\end{align*}
\noindent for the unconditional setting, with $\bm{y}^{(n+1,\,k)}$ being the $k$-th iterate of \cref{algo:SA}, with maximization direction $\hat{\bm{\theta}}_n$, and initialization point $\bm{y}^{(n+1,\, 0)}$.\newline

In CD learning of unconditional EBMs (i.e., in our unconditional setting), the Markov Chain is initialized at the empirical data distribution \citep{hinton_training_2000, carreira-perpinan_contrastive_2005}, as explained earlier. Persistent Contrastive Divergence (PCD) learning \citep{tieleman_training_2008} modifies CD by maintaining a persistent Markov chain. Thus, instead of initializing the chain from the data distribution in each iteration, the chain continues from its last state in the previous iteration, by setting $\bm{y}^{(n+1,\, 0)} = \bm{y}^{(n,\, K)}$. This approach aims to provide a better approximation of the model distribution and to reduce the bias introduced by the initialization of the Markov chain in CD. These are two types of informative initialization methods, which aim at reducing the mixing times of the Markov Chains. \newline

However, neither of these can be applied to the conditional setting, as observed in \citep{mnih_conditional_2012} in the context of conditional Restricted Boltzmann Machines (which are a type of EBMs). Indeed, on the one hand, PCD takes advantage of the fact that the parameter $\hat{\bm{\theta}}$ does not change too much from one iteration to the next, so that a Markov Chain that has reached equilibrium on $\hat{\bm{\theta}}_n$ is not far from equilibrium on $\hat{\bm{\theta}}_{n+1}$. This does not hold in the conditional setting, as each $\bm{x}_i$ leads to a different $\hat{\bm{\theta}}_i = g_{\hat{W}}(\bm{x}_i)$. On the other hand, the data-based initialization method in CD would amount to initialize the chains at the empirical marginal data distribution on $\mathcal{Y}$, and would be irrelevant in a conditional setting, since the distribution we want each Markov Chain to approximate is conditioned on the input $\bm{x}_i$.\newline

An option is to use persistent chains if training for multiple epochs, and to initialize the Markov Chain associated to $(\bm{x}_i, \y_i)$ for epoch $j$ at the final state of the one associated to the same data point $(\bm{x}_i, \y_i)$ at epoch $j-1$. However, this method is less relevant than PCD in the unconditional setting, as $\hat{w}$ changes a lot more in a full epoch than $\hat{\theta}$ in just one gradient step in the unconditional setting. It might be relevant, however, if each epoch consists in a single, full-batch gradient step. Nevertheless, it would require to store a significant number of states $\bm{y}_i^{(n,\, K)}$ (one for each point in the dataset). The solution we propose, for both full-batch and mini-batch settings, is to initialize the chains at the empirical data distribution conditioned on the input $\bm{x}_i$, which amounts to initialize them at the ground-truth $\y_i$. \newline

This discussion is summed up in \cref{tab:init}. Note that the use of uniform distributions $\mathcal{U}(\mathcal{Y})$ for the random initialization method can naturally be replaced with any other different prior distribution.

\begin{table*}[ht]
    \centering
    \caption{Possible Markov chain initialization methods under each learning setting}
    \renewcommand{\arraystretch}{1.5}  %
    \begin{tabular}{|>{\centering\arraybackslash}p{2.5cm}|>{\centering\arraybackslash}p{3.2cm}|>{\centering\arraybackslash}p{3.2cm}|>{\centering\arraybackslash}p{3.2cm}|}
        \hline
        \diagbox[width=2.5cm,height=1.5cm]{\shortstack{\scriptsize\textbf{Init.} \\[-0.2ex] \scriptsize\textbf{Method}}}{\scriptsize \textbf{Setting}} 
        & \textbf{Unconditional} & \textbf{Conditional, Batch} & \textbf{Conditional, Mini-Batch} \\
        \hline
        \textbf{Persistent} & $\bm{y}^{(n+1, 0)} = \bm{y}^{(n, K)}$ & $\bm{y}_i^{(n+1,\, 0)} = \bm{y}_i^{(n,\, K)}$ & / \\
        \hline
        \textbf{Data-Based} & $\bm{y}^{(n+1, 0)} = \y_j$, with $j\sim\mathcal{U}(\llbracket 1, N\rrbracket)$ & $\bm{y}_i^{(n+1,\, 0)} = \y_i$ & $\bm{y}_i^{(n+1,\, 0)} = \y_i$ \\
        \hline
        \textbf{Random} & $\bm{y}^{(n+1, 0)} \sim \mathcal{U}(\mathcal{Y})$ & $\bm{y}_i^{(n+1,\, 0)} \sim \mathcal{U}(\mathcal{Y})$ & $\bm{y}_i^{(n+1,\, 0)} \sim \mathcal{U}(\mathcal{Y})$ \\
        \hline
    \end{tabular}
    \label{tab:init}
\end{table*}

\newpage
\section{Experimental and methodological details for the DVRPTW (\cref{sec:dvrptw})}
\label{sec:challenge}

\subsection{Overview of the challenge}

We evaluate the proposed approach on a large-scale, ML-enriched combinatorial optimization problem: the \emph{EURO Meets NeurIPS 2022 Vehicle Routing Competition} \citep{kool_euro_2023}. In this dynamic vehicle routing problem with time windows (DVRPTW), requests arrive continuously throughout a planning horizon, which is partitioned into a series of delivery waves $\mathcal{W} = \{\left[\tau_0, \tau_1\right], \left[\tau_1, \tau_2\right],\dots, \left[\tau_{|\mathcal{W}|-1}, \tau_{|\mathcal{W}|}\right]\}$.

At the start of each wave $\omega$, a dispatching and vehicle routing problem must be solved for the set of requests $\mathcal{R}^\omega$ specific to that wave (in which we include the depot $D$), encoded into the system state $\bm{x}^\omega$. We denote by $\mathcal{Y}(\bm{x}^\omega)$ the set of feasible decisions associated to state $\bm{x}^\omega$.

A feasible solution $\y^\omega \in \cY(\x^\omega)$ must contain all requests that must be dispatched before $\tau_\omega$ (the rest are postponable), allow each of its routes to visit the requests they dispatch within their respective time windows, and be such that the cumulative customer demand on each of its routes does not exceed a given vehicle capacity. It is encoded thanks as a matrix $\left(y^\omega_{i, j}\right)_{i, j\,\in\mathcal{R}^\omega}$, where $y^\omega_{i, j} = 1$ if the solution contains the directed route segment from $i$ to $j$, and $y^\omega_{i, j} = 0$ otherwise. The set of requests $\mathcal{R}^{\omega+1}$ is obtained by removing all requests dispatched by the chosen solution $\bm{y}^\omega$ from $\mathcal{R}^\omega$ and adding all new requests which arrived between $\tau_\omega$ and $\tau_{\omega+1}$. 

The aim of the challenge is to find an optimal policy $f \colon \cX \to \cY$ assigning decisions $\bm{y}^\omega\in\mathcal{Y}(\bm{x}^\omega)$ to system states $\bm{x}^\omega\in\mathcal{X}$. This can be cast as a reinforcement learning problem:
$$\min_f \mathbb{E}\left[ c_\mathcal{W}(f) \right], \quad\text{with}\quad c_\mathcal{W}(f)\coloneqq \sum_{\omega\in\mathcal{W}}c(f(\bm{x}^\omega)),$$
where $c:\bm{y}^\omega\mapsto\sum_{i, j\,\in\mathcal{R}^\omega}c_{i, j}\,y^\omega_{i, j}$ gives the routing cost of $\bm{y}^\omega\in\mathcal{Y}^\omega$ and where $c_{i,j}\geq0$ is the routing cost from $i$ to $j$. The expectation is taken over full problem instances.

\subsection{Reduction to supervised learning} 

We follow the method of \citep{baty_combinatorial_2024}, which was the winning approach for the challenge. Central to this approach is the concept of prize-collecting dynamic vehicle routing problem with time windows (PC-VRPTW). In this setting, each request $i\in\mathcal{R}^\omega$ is assigned an artificial \emph{prize} $\theta_i^\omega\in\mathbb{R}$, that reflects the benefit of serving it. The prize of the depot $D$ is set to $\theta^\omega_D=0$. The objective is then to identify a set of routes that maximizes the total prize collected while minimizing the associated travel costs. 
The model $g_W$ predicts the prize vector $\bm{\theta}^\omega = g_W(\bm{x}^\omega)$.
Denoting $\varphi(\bm{y}) \coloneqq -\langle \bm{c}, \bm{y}\rangle$, the corresponding optimization problem can be written as:
\begin{align}
\label{pb:dvrptw_2}
\hat{\y}(\Thetav^\omega)=\argmax_{\bm{y}\in\mathcal{Y}(\bm{x}^\omega)} \sum_{i, j\,\in\mathcal{R}^\omega} \theta^\omega_i \, y_{i, j} - \!\!\sum_{i, j\,\in\mathcal{R}^\omega}c_{i, j}\,y_{i, j} =\langle \bm{\Theta}^\omega ,\bm{y}\rangle +\varphi(\bm{y}),
\end{align}
where $\Thetav^\omega=\mathsf{expand}(\thetav^\omega)\in\RR^{|\cR^\omega|\times|\cR^\omega|}$ is an expanded, edge-level version of $\thetav^\omega$ with $\Theta^\omega_{i,j}=\theta^\omega_i$ for all $j\in\cR^\omega$.

The overall pipeline is summarized in \cref{fig:dvrptw_pipeline}.
Following \citep{baty_combinatorial_2024}, we approximately solve Problem~\eqref{pb:dvrptw_2} using
the prize-collecting HGS heuristic (PC-HGS), a variant of hybrid genetic search (HGS) \citep{Vidal_2022}. We denote this approximate solver $\widetilde \y \approx \widehat \y$, so that their proposed policy decomposes 
as $f_W \coloneqq \widetilde{\y} \circ\mathsf{expand}\circ g_W$. 
The ground-truth routes are created by using an anticipative strategy, i.e., by solving multiple instances where all future information is revealed from the start, and the requests' arrival times information is translated into time windows (thus removing the dynamic aspect of the problem). This anticipative policy, which we denote by $f^\star$ (which cannot be attained as it needs unavailable information) is thus the target policy imitated by the model – see \cref{sec:additional_dvrptw} for more details.

\subsection{Perturbation-based baseline}

In \citep{baty_combinatorial_2024}, a perturbation-based method \citep{berthet_learning_2020} was used. This method is based on injecting noise in the PC-HGS solver $\widetilde \y$. Similarly to our approach, the parameters $W$ can then be learned using a Fenchel-Young loss, since the loss is minimized when the perturbed solver correctly predicts the ground truth. However, since $\widetilde \y$ is not an exact solver, all theoretical learning guarantees associated with this method (e.g., correctness of the gradients) no longer hold.

\subsection{Proposed approach} 

Our proposed approach instead uses the Fenchel-Young loss associated with the proposed layer,
which is minimized when the proposed layer correctly predicts the ground-truth.
At inference time, however, we use $f_W \coloneqq \tilde{\y} \circ\mathsf{expand}\circ g_W$. We use a mixture of proposals, as defined in \cref{algo:mixture_SA}. To design each proposal $q_s$, we build randomized versions of moves specifically designed for the prize-collecting dynamic vehicle routing problem with time windows. More precisely, we base our proposals on moves used in the local search part of the PC-HGS algorithm, which are summarized in \cref{table:operators}. The details of turning these moves into proposal distributions with tractable individual correction ratios are given in \cref{sec:dvrptw_proposals}.

We evaluate three different initialization methods: (i) initialize $\bm{y}^{(0)}$ by constructing routes dispatching random requests, (ii) initialize $\bm{y}^{(0)}$ to the ground-truth solution, (iii) initialize $\bm{y}^{(0)}$ by starting from the dataset ground-truth and applying a heuristic initialization algorithm to improve it. This heuristic initialization, akin to a short local search, is also used by the PC-HGS algorithm $\widetilde \y$, and has its runtime limited to half the time allocated to the layer (a limit it does not reach in practice).

\subsection{Proposal distribution design}
\label{sec:dvrptw_proposals}

\paragraph{Original moves.} The selected moves, designed for Local Search algorithms on vehicle routing problems (specifically for the PC-VRPTW for \texttt{serve request} and \texttt{remove request}), are given in \cref{table:operators_2}. All of these moves (except for \texttt{remove request}) involve selecting two clients $i$ and $j$ from the request set $\mathcal{R}^\omega$ (for example, the \texttt{relocate} move relocates client $i$ after client $j$ in the solution).

In the Local Search part of the PC-HGS algorithm from \citet{Vidal_2022}, they are implemented as deterministic functions used within a quadratic loop over clients, and are performed only if they improve the solution's objective value. The search is narrowed down to client pairs $(i, j)$ such that $d(i, j)$ is among the $N_\text{prox}$ lowest values in $\bigl\{ d(i, k) \mid k\in\mathcal{R}^\omega \setminus \{D, i\} \bigr\}$, where $d$ is a problem-specific heuristic distance measure between clients, based on spatial features and time windows, and $N_\text{prox}$ is a hyperparameter. These distances are independent from the chosen solution routes (they are computed once at the start of the algorithm, from the problem features), non-negative, and symmetric: $d(i, j) = d(j, i)$.

\begin{table*}
\begin{threeparttable}
\small
\begin{tabularx}{\textwidth}{p{3cm} X}
\hline
Name & Description \\ \hline
\texttt{relocate} & removes request $i$ from its route and re-inserts it before or after request $j$ \\
\cellcolor{LightGray} \texttt{relocate pair}  & \cellcolor{LightGray} removes pair of requests $(i,\mathsf{next}(i))$ from their route and re-inserts them before or after request $j$ \\
\texttt{swap} & exchanges the position of requests $i$ and $j$ in the solution \\
\cellcolor{LightGray} \texttt{swap pair} & \cellcolor{LightGray} exchanges the positions of the pairs $(i,\mathsf{next}(i))$ and $(j,\mathsf{next}(j))$ in the solution\\
\texttt{2-opt} & reverses the route segment between $i$ and $j$ \\
\cellcolor{LightGray} \texttt{serve request} & \cellcolor{LightGray} inserts currently undispatched request $i$ before or after request $j$ \\
\texttt{remove request} & removes currently dispatched request $i$ from the solution \\ \hline
\end{tabularx}
\end{threeparttable}
\caption{PC-VRPTW Local search moves}
\label{table:operators_2}
\end{table*}

\paragraph{Randomization.} In order to transform these deterministic moves into proposals, we first adapt the choice of clients $i$ and $j$, by sampling $i$ uniformly from $V^1_s(\bm{y})$, which contains the set of valid choices of client $i$ for move $s$ from solution $\bm{y}$. Then, we sample $j$ from $V^2_s(\bm{y})[i]\setminus\{i\}$ using the following softmax distribution: $P_s(j\mid i) = \frac{\exp\left[-d(i, j)/\beta\right]}{\sum_{k\in V^2_s(\bm{y})[i]\setminus\{i\}}\exp\left[- d(i, k)/\beta\right]}$, where $\beta>0$ is a neighborhood sampling temperature. The set $V^2_s(\bm{y})[i]$ contains all valid choices of client $j$ for move $s$ from solution $\bm{y}$, and is specified along with $V^1_s(\bm{y})$ in \cref{table:valid_clients}. We normalize the distance measures inside the softmax, by dividing them by the maximum distance: $d(i, \cdot )\gets d(i,\cdot)/\max_{k\in V^2_s(\bm{y})[i]\setminus\{i\}}d(i, k)$.

\begin{table*}
\begin{threeparttable}
\begin{tabularx}{\textwidth}{p{3cm} p{4cm} X}
\hline
Move & $V^1_s(\bm{y})$ & $V^2_s(\bm{y})[i]$ \\ \hline
\texttt{relocate} & $\mathcal{D}(\bm{y}) \setminus \mathcal{D}_1(\bm{y})$ & $\mathcal{D}(\bm{y})$ \\
\cellcolor{LightGray} \texttt{relocate pair}  & \cellcolor{LightGray} $\mathcal{D}(\bm{y}) \setminus \bigl\{\mathcal{D}_2(\bm{y})\cup \mathcal{D}^\text{last}(\bm{y})\bigr\}$ & \cellcolor{LightGray} $\mathcal{D}(\bm{y}) \setminus \{\mathsf{next}(i)\}$  \\
\texttt{swap} & $\mathcal{D}(\bm{y})$ & $\mathcal{D}(\bm{y})$ \\
\cellcolor{LightGray} \texttt{swap pair} & \cellcolor{LightGray} $\mathcal{D}(\bm{y}) \setminus \mathcal{D}^\text{last}(\bm{y})$ & \cellcolor{LightGray} $\mathcal{D}(\bm{y}) \setminus \bigl\{\mathcal{D}^\text{last}(\bm{y}) \cup \{
\mathsf{prev}(i),\mathsf{next}(i)\}\bigr\}$ \\
\texttt{2-opt} & $\mathcal{D}(\bm{y}) \setminus \mathcal{D}_2(\bm{y})$ & $\mathcal{D}(\bm{y}) \setminus \mathcal{D}_2(\bm{y})$ \\
\cellcolor{LightGray} \texttt{serve request} & \cellcolor{LightGray} $\bar{\mathcal{D}}(\bm{y})$ & \cellcolor{LightGray} $\mathcal{D}(\bm{y}) \cup \mathcal{I}_D(\bm{y}) $ \\
\texttt{remove request} & $\bigl\{\mathcal{D}(\bm{y}) \setminus \mathcal{D}_1(\bm{y})\bigr\} \cup \mathcal{I}_1(\bm{y})$ &  \\ \hline
\end{tabularx}
\end{threeparttable}
\caption{Sets of valid clients for each move.  $\mathcal{D}(\bm{y})$ contains all dispatched clients in solution $\bm{y}$. $\mathcal{D}_1(\bm{y})$ contains all dispatched clients that are the only client in their route. $\mathcal{D}_2(\bm{y})$ contains all dispatched clients that are in a route with $2$ clients or less. $\mathcal{D}^\text{last}(\bm{y})$ contains all dispatched clients that are the last of their route. $\bar{\mathcal{D}}(\bm{y})$ contains all non-dispatched clients. $\mathcal{I}_D(\bm{y})$ contains the depot of the first empty route, if it exists (all routes may be non-empty), or else is the empty set. $\mathcal{I}_1(\bm{y})$ contains the only client in the last non-empty route if it contains exactly one client, or else is the empty set.}
\label{table:valid_clients}
\end{table*}

\paragraph{Neighborhood graph symmetrization.} Then, we ensure that each individual neighborhood graph $\mathcal{N}_s$ is undirected. This is already the case for the moves \texttt{swap}, \texttt{swap pair} and \texttt{2-opt}, as they are actually involutions (applying the same move on the same couple $(i, j)$ from $\bm{y}'$ will result in $\bm{y}$). However, this is obviously not the case for \texttt{serve request} and \texttt{remove request}. Indeed, if solution $\bm{y}'$ is obtained from $\bm{y}$ by removing a dispatched client (respectively serving an non-dispatched one), $\bm{y}$ cannot be obtained by removing another one (respectively, serving another one). To fix this, we merge these two moves into a single one. First, it evaluates which of the two moves are allowed (i.e., if they are such that $V^1_s(\bm{y})\neq \emptyset$). Then, it samples one (the probability of selecting ``remove'' is chosen to be equal to the number of removable clients divided by the number of removable clients plus the number of servable clients) in the case where both are possible, or else simply performs the only move allowed. Thus, the corresponding neighborhood graph is undirected as it is always possible to perform the reverse operation (as when removing a client, it becomes unserved, thus allowing the \texttt{serve request} move from $\bm{y}'$, and vice-versa). We also allow the \texttt{serve request} move to insert a client after the depot of the first empty route, to allow the creation of new routes. In consequence, we allow the \texttt{remove request} move to remove the only client in the last non-empty route if it contains exactly one client (to maintain symmetry of the neighborhood graph).

For the \texttt{relocate} and \texttt{relocate pair} moves, the non-reversibility comes from the fact that they only relocate client $i$ (or clients $i$ and $\mathsf{next}(i)$ in the pair case) after client $j$, so that if client $i$ was the first in its route, relocating it back would be impossible (the depot, which is the start of the route, cannot be selected as $j$). Thus, we allow insertions before clients too, and add a random choice with probability $(\frac{1}{2}, \frac{1}{2})$ to determine if the relocated client(s) will be inserted before or after $j$. We also add this feature to the \texttt{serve request} move.

\paragraph{Correction ratio computation.} Next, we implement the computation of the individual correction ratio $\tilde{\alpha}_s(\bm{y}, \bm{y}')=\frac{q_s(\bm{y}',\bm{y})}{q_s(\bm{y},\bm{y}')}$ for each proposal $q_s$.

\begin{itemize}
    \item In the case of \texttt{swap} and \texttt{2-opt}, we have $\tilde{\alpha}_s(\bm{y}, \bm{y}')=1$. Indeed, let $\bm{y}'$ be the result of applying one of these moves $s$ on $\bm{y}$ when sampling $i\in V^1_s(\bm{y})$ and $j\in V^2_s(\bm{y})[i]\setminus\{i\}$. We then have:
    \begin{align*}
        q_s(\bm{y},\bm{y}') &= \frac{1}{|V^1_s(\bm{y})|}\cdot \frac{\exp\left[-d(i, j)/\beta\right]}{\sum_{k\in V^2_s(\bm{y})[i]\setminus\{i\}}\exp\left[- d(i, k)/\beta\right]} \\
        &+ \frac{1}{|V^1_s(\bm{y})|}\cdot \frac{\exp\left[-d(j, i)/\beta\right]}{\sum_{k\in V^2_s(\bm{y})[j]\setminus\{j\}}\exp\left[- d(j, k)/\beta\right]},
    \end{align*}
    where the first term accounts for the probability of selecting $i$ then $j$ and the second term accounts for that of selecting $j$ then $i$ (one can easily check that these two cases are the only way of sampling $\bm{y}'$ from $\bm{y}$). Then, noticing that we have $|V^1_s(\bm{y}')|=|V^1_s(\bm{y})|$, that these moves are involutions (selecting $(i,j)$ or $(j,i)$ from $\bm{y}'$ is also the only way to sample $\bm{y}$), and that we have the equalities $V^2_s(\bm{y})[i]=V^2_s(\bm{y}')[i]$ and $V^2_s(\bm{y})[j]=V^2_s(\bm{y}')[j]$, we actually have $q_s(\bm{y}',\bm{y})=q_s(\bm{y},\bm{y}')$.

    \item For \texttt{swap pair}, the same arguments hold (leading to the same form for $q_s$), except for the equalities $V^2_s(\bm{y})[i]=V^2_s(\bm{y}')[i]$ and $V^2_s(\bm{y})[j]=V^2_s(\bm{y}')[j]$. Thus, we have the following form for the correction ratio:

    \begin{align*}
        \frac{q_s(\y',\y)}{q_s(\y,\y')} = \frac{\sum_{k\in V^2_s(\bm{y})[i]\setminus\{i\}}\exp\left[- d(i, k)/\beta\right] + \sum_{k\in V^2_s(\bm{y})[j]\setminus\{j\}}\exp\left[- d(j, k)/\beta\right]}{\sum_{k\in V^2_s(\bm{y}')[i]\setminus\{i\}}\exp\left[- d(i, k)/\beta\right] +\sum_{k\in V^2_s(\bm{y}')[j]\setminus\{j\}}\exp\left[- d(j, k)/\beta\right]}.
    \end{align*}

    \item In the case of \texttt{relocate}, let $j'$ denote $\mathsf{next}(j)$ if the selected insertion type was ``after'', and $\mathsf{prev}(j)$ if it was ``before'' – where $\mathsf{next}(j)\in\mathcal{R}^\omega$ denotes the request following $j$ in solution $\bm{y}$, i.e., the only index $k$ such that $\bm{y}_{j,k}=1$, and $\mathsf{prev}(j)$ is the one preceding it, i.e., the only $k$ such that $\bm{y}_{k,j}=1$. We have:
    \begin{align*}
        q_s(\bm{y},\bm{y}')&=\frac{1}{2}\cdot\frac{1}{|V^1_s(\bm{y})|}\cdot \frac{\exp\left[-d(i, j)/\beta\right]}{\sum_{k\in V^2_s(\bm{y})[i]\setminus\{i\}}\exp\left[- d(i, k)/\beta\right]}\\
        &+ \frac{1}{2}\cdot\frac{1}{|V^1_s(\bm{y})|}\cdot \frac{\exp\left[-d(i, j')/\beta\right]}{\sum_{k\in V^2_s(\bm{y})[i]\setminus\{i\}}\exp\left[- d(i, k)/\beta\right]}
    \end{align*}
    Indeed, if $i$ was relocated \emph{after} $j$, the same solution $\bm{y}'$ could have been obtained by relocating $i$ \emph{before} $j' = \mathsf{next}(j)$. Similarly, if $i$ was relocated \emph{before} $j$, the same solution $\bm{y}'$ could have been obtained by relocating $i$ \emph{after} $j' = \mathsf{prev}(j)$.  For the reverse move probability, the way of obtaining $\bm{y}$ from $\bm{y}'$ is either to select $(i, \mathsf{prev}(i))$ in the after-type insertion case, or $(i, \mathsf{next}(i))$ in the before-type insertion case (where $\mathsf{prev}$ and $\mathsf{next}$ are taken w.r.t. $\bm{y}$, i.e., before applying the move). Thus, we have:
    \begin{align*}
        q_s(\bm{y}',\bm{y})&=\frac{1}{2}\cdot\frac{1}{|V^1_s(\bm{y}')|}\cdot \frac{\exp\left[-d(i, \mathsf{prev}(i))/\beta\right]}{\sum_{k\in V^2_s(\bm{y}')[i]\setminus\{i\}}\exp\left[- d(i, k)/\beta\right]}\\
        &+ \frac{1}{2}\cdot\frac{1}{|V^1_s(\bm{y}')|}\cdot \frac{\exp\left[-d(i, \mathsf{next}(i))/\beta\right]}{\sum_{k\in V^2_s(\bm{y}')[i]\setminus\{i\}}\exp\left[- d(i, k)/\beta\right]}.
    \end{align*}

    \item For the \texttt{relocate pair} move, the exact same reasoning and proposal probability form hold for the forward move, but we have for the reverse direction:
    \begin{align*}
        q_s(\bm{y}',\bm{y})&=\frac{1}{2}\cdot\frac{1}{|V^1_s(\bm{y}')|}\cdot \frac{\exp\left[-d(i, \mathsf{prev}(i))/\beta\right]}{\sum_{k\in V^2_s(\bm{y}')[i]\setminus\{i\}}\exp\left[- d(i, k)/\beta\right]}\\
        &+ \frac{1}{2}\cdot\frac{1}{|V^1_s(\bm{y}')|}\cdot \frac{\exp\left[-d(i, \mathsf{next}(\mathsf{next}(i)))/\beta\right]}{\sum_{k\in V^2_s(\bm{y}')[i]\setminus\{i\}}\exp\left[- d(i, k)/\beta\right]},
    \end{align*}
    as client $\mathsf{next}(i)$ is also relocated.

    \item For the \texttt{serve request} / \texttt{remove request} move, we have the forward probability:
    \begin{align*}
    q_s(\bm{y},\bm{y}') &= \frac{|\bigl\{\mathcal{D}(\bm{y}) \setminus \mathcal{D}_1(\bm{y})\bigr\} \cup \mathcal{I}_1(\bm{y})|}{|\bigl\{\mathcal{D}(\bm{y}) \setminus \mathcal{D}_1(\bm{y})\bigr\} \cup \mathcal{I}_1(\bm{y})|+|\bar{\mathcal{D}}(\bm{y})|}\times \frac{1}{|\bigl\{\mathcal{D}(\bm{y}) \setminus \mathcal{D}_1(\bm{y})\bigr\} \cup \mathcal{I}_1(\bm{y})|} \\
    &= \frac{1}{|\bigl\{\mathcal{D}(\bm{y}) \setminus \mathcal{D}_1(\bm{y})\bigr\} \cup \mathcal{I}_1(\bm{y})|+|\bar{\mathcal{D}}(\bm{y})|}
    \end{align*}
    
    if the chosen move is \texttt{remove request}. The expression corresponds to the composition of move choice sampling and uniform sampling over removable clients. 
    
    Still in the same case (\texttt{remove request} is chosen) and if the removed request $i$ was in $\mathcal{I}_1(\bm{y})$ (i.e., was the only client in the last non-empty route if the latter contained exactly one client), we have the reverse move probability:
    \begin{align*}
        q_s(\bm{y}',\bm{y}) &= \frac{1}{|\bigl\{\mathcal{D}(\bm{y}') \setminus \mathcal{D}_1(\bm{y}')\bigr\} \cup \mathcal{I}_1(\bm{y}')|+|\bar{\mathcal{D}}(\bm{y}')|}\\
        &\times \frac{\exp\left[-\bar{d}(i)/\beta\right]}{\exp\left[-\bar{d}(i)/\beta\right]+\sum_{k\in \mathcal{D}(\bm{y}')}\exp\left[- d(i, k)/\beta\right]}\\
        &= \frac{1}{|\bigl\{\mathcal{D}(\bm{y}) \setminus \mathcal{D}_1(\bm{y})\bigr\} \cup \mathcal{I}_1(\bm{y})|+|\bar{\mathcal{D}}(\bm{y})|}\\
        &\times \frac{\exp\left[-\bar{d}(i)/\beta\right]}{\exp\left[-\bar{d}(i)/\beta\right]+\sum_{\substack{k\in \mathcal{D}(\bm{y}) \\ k\neq i}}\exp\left[- d(i, k)/\beta\right]}.
    \end{align*}
    The expression corresponds to the composition of move choice sampling and softmax sampling of the depot of the first empty route (which was the route of the removed client $i$, so that $\mathcal{I}_D(\bm{y}')\neq\emptyset$ in this case). We use the average distance to dispatched clients $\bar{d}(i)\coloneqq \frac{1}{|\mathcal{D}(\bm{y}')|}\sum_{k\in\mathcal{D}(\bm{y}')}d(i,k)$ as distance to the depot.

    In the case where the removed request $i$ was not in $\mathcal{I}_1(\bm{y})$, we have instead:
    \begin{align*}
        q_s(\bm{y}',\bm{y}) &= \frac{1}{|\bigl\{\mathcal{D}(\bm{y}') \setminus \mathcal{D}_1(\bm{y}')\bigr\} \cup \mathcal{I}_1(\bm{y}')|+|\bar{\mathcal{D}}(\bm{y}')|} \\
        &\times \frac{\frac{1}{2}\cdot \exp\left[-d(i,\mathsf{prev}(i))/\beta\right]+\frac{1}{2}\cdot \exp\left[-d(i,\mathsf{next}(i))/\beta\right]}{\mathbf{1}_{\{\mathcal{I}_D(\bm{y}')\neq\emptyset\}}\cdot\exp\left[-\bar{d}(i)/\beta\right]+\sum_{k\in \mathcal{D}(\bm{y}')}\exp\left[- d(i, k)/\beta\right]}\\
        &= \frac{1}{|\bigl\{\mathcal{D}(\bm{y}) \setminus \mathcal{D}_1(\bm{y})\bigr\} \cup \mathcal{I}_1(\bm{y})|+|\bar{\mathcal{D}}(\bm{y})|} \\
        &\times \frac{\frac{1}{2}\cdot \exp\left[-d(i,\mathsf{prev}(i))/\beta\right]+\frac{1}{2}\cdot \exp\left[-d(i,\mathsf{next}(i))/\beta\right]}{\mathbf{1}_{\{\mathcal{I}_D(\bm{y}')\neq\emptyset\}}\cdot\exp\left[-\bar{d}(i)/\beta\right]+\sum_{\substack{k\in \mathcal{D}(\bm{y})\\ k\neq i}}\exp\left[- d(i, k)/\beta\right]}.
    \end{align*}
    
    The right term corresponds to softmax sampling of the previous node with ``after'' insertion type (which has probability $1/2$) and of the next node with ``before'' insertion type. The non-emptiness of $\mathcal{I}_D(\bm{y}')$ is not guaranteed anymore, as all routes might be non-empty (indeed, we did not create an empty one by removing $i$, as $i\in\mathcal{D}(\bm{y})\setminus \mathcal{D}_1(\bm{y})$ in this case).

    Similarly, if the chosen move is \texttt{serve request}, we have the forward probability:
    \begin{align*}
        q_s(\bm{y},\bm{y}') &= \frac{|\bar{\mathcal{D}}(\bm{y})|}{|\bigl\{\mathcal{D}(\bm{y}) \setminus \mathcal{D}_1(\bm{y})\bigr\} \cup \mathcal{I}_1(\bm{y})|+|\bar{\mathcal{D}}(\bm{y})|} \\
        &\times \frac{\frac{1}{2}\cdot \exp\left[-d(i,j)\right]+\frac{1}{2}\cdot \exp\left[-d(i,j')\right]}{\mathbf{1}_{\{\mathcal{I}_D(\bm{y})\neq\emptyset\}}\cdot\exp\left[-\bar{d}(i)/\beta\right]+\sum_{k\in \mathcal{D}(\bm{y})}\exp\left[- d(i, k)/\beta\right]}
    \end{align*}
    if the selected insertion node $j$ is not in $\mathcal{I}_D(\bm{y})$ (i.e., is not the depot of the first empty route in $\bm{y}$), where $j'=\mathsf{prev}(j)$ if the insertion type selected was ``before'' (which has probability $1/2$), and  $j'=\mathsf{next}(j)$ if it was ``after''.

    We have instead the forward probability:
    \begin{align*}
        q_s(\bm{y},\bm{y}') &= \frac{1}{|\bigl\{\mathcal{D}(\bm{y}) \setminus \mathcal{D}_1(\bm{y})\bigr\} \cup \mathcal{I}_1(\bm{y})|+|\bar{\mathcal{D}}(\bm{y})|} \\
        &\times \frac{\exp\left[-\bar{d}(i)/\beta\right]}{\exp\left[-\bar{d}(i)/\beta\right]+\sum_{k\in \mathcal{D}(\bm{y})}\exp\left[- d(i, k)/\beta\right]}
    \end{align*}
    if the selected insertion node $j$ is in $\mathcal{I}_D(\bm{y})$ (i.e., is the depot of the first empty route in $\bm{y}$).

    In every case, we have the reverse move probability:
    \begin{align*}
    q_s(\bm{y}',\bm{y}) = \frac{1}{|\bigl\{\mathcal{D}(\bm{y}) \setminus \mathcal{D}_1(\bm{y})\bigr\} \cup \mathcal{I}_1(\bm{y})|+|\bar{\mathcal{D}}(\bm{y})|}.
    \end{align*}
    
\end{itemize}

In each case, we set $d(i, D) = \infty$ to account for the fact that the depot can never be sampled during the process (except in the \texttt{serve request} / \texttt{remove request} move, where we allow the depot of the first empty route / last non-empty route to be selected, for which we use the average distance to other requests as explained earlier) – in fact, the distance measure from a client to the depot is not even defined in the original HGS implementation.

The second correction factor needed is $\frac{|Q(\bm{y})|}{|Q(\bm{y}')|}$ (see \cref{algo:mixture_SA}). We compute it by checking if each move is allowed, i.e., if there exists at least one $i\in V^1_s(\bm{y})$ such that $V^2_s(\bm{y})[i]\setminus \{i\} \neq \emptyset$. This can be determined in $\mathcal{O}(\mathcal{R}^\omega)$ for each move.

\subsection{Performance metric.} 

As the Fenchel-Young loss $\ell_t$ actually minimized is intractable to compute exactly, we only use the challenge metric. More precisely, we measure the cost relative to that of the anticipative baseline, $\frac{c_\mathcal{W}(f_W)-c_\mathcal{W}(f^\star)}{c_\mathcal{W}(f^\star)}$, which we average over a test dataset of unseen instances.

\subsection{Results.} 

In \cref{fig:dvrptw_init_iter}, we observe that the initialization method plays an important role, and the ground-truth-based ones greatly outperform the random one.

We observe that the number of Markov iterations $K$ is an important performance factor. Interestingly, the ground-truth initialization significantly improves the learning process for small $K$.

In \cref{tab:fixed_compute}, we compare training methods with fixed compute time budget for the layer (perturbed solver or proposed MCMC approach), which is by far the main computational bottleneck. This parameter limits the time allowed for a single forward pass through the combinatorial optimization layer (be it the perturbed inexact oracle or the proposed method). In both cases, the backward pass through the layer is immediate, as a property of the expression of the gradient of Fenchel-Young losses. The models are selected using a validation set and evaluated on the test set. We observe that the proposed approach significantly outperforms the perturbation-based method \citep{berthet_learning_2020} using $\widetilde{\y}$ in low time limit regimes, thus allowing for faster and more efficient training.

Full experimental details and additional results on the impact of temperature are given in \cref{sec:additional_dvrptw}.

\subsection{Additional experimental details and results}
\label{sec:additional_dvrptw}
\paragraph{Model, features, dataset, hyperparameters, compute.} Following \citet{baty_combinatorial_2024}, the differentiable ML model $g_W$ is implemented as a sparse graph neural network. We also use the same feature set, which represents the system state $\x^\omega$ as a vector comprising request-level features, such as coordinates, time windows, demands, travel time to the depot, and quantiles from the distribution of the travel time to all other requests (named \emph{complete} feature set, and described in the Table 4 of their paper). We use the same training, validation, and testing datasets, which are created from $30$, $15$ and $25$ problem instances respectively. The training set uses a sample size of $50$ requests per wave, while the rest use $100$. The solutions in the training dataset, i.e., the examples from the anticipative strategy $f^\star$ imitated by the model, are obtained by solving the corresponding offline VRPTWs using HGS \citep{Vidal_2022} with a time limit of $3600$ seconds. During evaluation, the PC-HGS solver $\tilde{\y}$ is used with a constant time limit of $60$ seconds for all models. We use Adam \citep{kingma_adam_2017} together with the proposed stochastic gradient estimators, with a learning rate of $5\cdot10^{-3}$. Each training is performed using only a single CPU worker. For \cref{fig:dvrptw_init_iter}, we use a temperature $t=10^2$. For \cref{tab:fixed_compute}, we use $1$ Monte-Carlo sample for the perturbation-based method and $1$ Markov chain for the proposed approach (in order to have a fair comparison: an equal number of oracle calls / equal compute).

\paragraph{Statistical significance.} Each training is performed $50$ times with the same parameters and different random seeds. Then, the learning curves are averaged, and plotted with a $95\%$ confidence interval. For the results in \cref{tab:fixed_compute}, we report the performance of the best model iteration (selected with respect to the validation set) on the test set. This procedure is also averaged over $50$ trainings, and reported with $95\%$ confidence intervals.

\begin{figure}[H]
    \centering
    \begin{subfigure}[b]{0.4\textwidth}
        \centering
        \includegraphics[width=\textwidth]{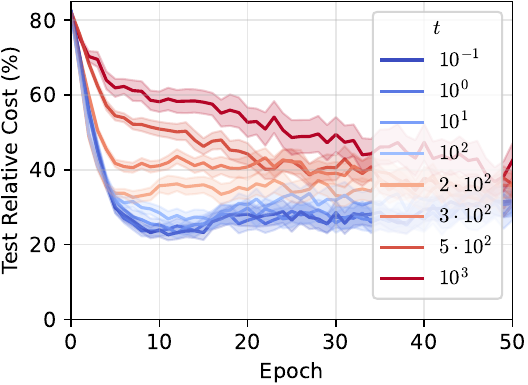}
    \end{subfigure}
    \begin{subfigure}[b]{0.4\textwidth}
        \centering
        \includegraphics[width=\textwidth]{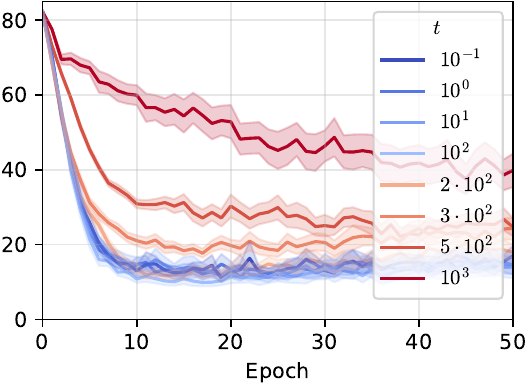}
    \end{subfigure}
    \caption{Test relative cost (\%). \textbf{Left}: varying temperature $t$ with random initialization. \textbf{Right}: varying temperature $t$ with ground-truth initialization.}
    \label{fig:dvrptw_temp}
\end{figure}

\paragraph{Additional results.} In \cref{fig:dvrptw_temp}, we report model performance for varying temperature $t$. Interestingly, lower temperatures perform better when using random initialization. In the ground-truth initialization setting, a sweet spot is found at $t=10^2$, but lower temperatures do not particularly decrease performance. 

\section{Experimental and methodological details for the MKP (\cref{sec:mkp})}
\label{sec:mkp_details}

First, we recall that the combinatorial optimization layer is defined as:
\begin{align*}
    \widehat{\y}(\thetav)\coloneqq \argmax_{\y\in\{0,1\}^d}\;&\sum_{i=1}^d\theta_i y_i \quad = \argmax_{\y\in\cY}\langle\thetav,\y\rangle\,,\\
    \text{s.t.}\;\forall j\in[M],\; &\sum_{i=1}^d w_{i,j}y_i\leq C_j
\end{align*}
where $\thetav=g_W(\x) \in \RR^d$ are the item values, $w_{i,j}\geq 0$ is the weight of item $i$ in dimension $j$, and $C_j$ is the capacity of dimension $j$. The feasible set is $\cY\coloneqq \{\y\in\{0,1\}^d\mid \forall j\in[M],\;\sum_{i=1}^d w_{i,j}y_i\leq C_j\}$.

\subsection{Proposal distribution design}
\label{sec:mkp_proposals}

In this experiment, defined in \cref{sec:mkp}, we use \cref{algo:mixture_SA} with a mixture of three proposal distributions $q_1$, $q_2$ and $q_3$ ($S=3$).

Let $\y\in\cY$ be a current feasible solution, and let $I(\y) = \{i \mid y_i=1\}$ and $\bar{I}(\y) = \{j \mid y_j=0\}$ denote the indices of selected and unselected items. Given a binary vector $\y\in\{0,1\}^d$ and an index $i\in[d]$, we denote by $\y_{y_i\to\bar{y_i}}$ the vector where the $i$-th bit is flipped, i.e.:
\begin{align*}
    (\y_{y_i\to\bar{y_i}})_k = \begin{cases}
    1-y_i &\text{if $k=i$,}\\
    y_i &\text{else.}
\end{cases}
\end{align*}
Given two indices $i,j\in[d]$, we denote by $\y_{i \leftrightarrow j}\in\{0,1\}^d$ the vector where the $i$-th and $j$-th bits are swapped, i.e.:
\begin{align*}
    (\y_{i \leftrightarrow j})_k = \begin{cases}
    y_j &\text{if $k=i$,}\\
    y_i &\text{if $k=j$,}\\
    y_k &\text{else.}
\end{cases}
\end{align*}

We use a sampling temperature $\beta=1.0$ and define the following moves:

\begin{itemize}[leftmargin=*, topsep=2pt, itemsep=2pt]
\item \textbf{Uniform swap ($q_1$).} The neighborhood $\cN_1(\y)$ consists of all feasible solutions obtained by swapping an active item $i \in I(\y)$ with an inactive one $j \in \bar{I}(\y)$, i.e.:
\begin{align*}
   \cN_1(\y)=\{\y'\in\cY\mid \exists i\in I(\y), j\in\bar{I}(\y):  \y'=\y_{i \leftrightarrow j}\}. 
\end{align*}
The proposal is uniform over this neighborhood: $\forall \y_{i \leftrightarrow j}\in\cN_1(\y),\;q_{1}(\y, \y_{i \leftrightarrow j}) = \frac{1}{|\cN_1(\y)|}$.

\item \textbf{Guided swap ($q_2$).} Using the same swap neighborhood $\cN_2(\y) = \cN_1(\y)$, we bias the selection using the predicted item values $\thetav$. We sample item $i\in I(\y)$ to drop with probability $p_{\text{drop}}(i) \propto e^{-\theta_i/\beta}$ and item $j\in\bar{I}(\y)$ to add with $p_{\text{add}}(j) \propto e^{\theta_j/\beta}$.

The proposal distribution is therefore: $\forall \y_{i \leftrightarrow j}\in\cN_1(\y),\;q_2(\y, \y_{i \leftrightarrow j}) \propto \exp\left(\frac{\theta_j - \theta_i}{\beta}\right)$.

\item \textbf{Guided flip ($q_3$).} The neighborhood $\cN_3(\y)$ consists of all feasible solutions obtained by flipping a single bit $i$, i.e.:
\begin{align*}
    \cN_3(\y)=\{\y'\in\cY\mid \exists i\in[d]:  \y'=\y_{y_i\to\bar{y_i}}\}.
\end{align*}
We sample index $i$ with probability proportional to $e^{-\theta_i/\beta}$ if $y_i=1$ (favoring dropping low-value items) and $e^{\theta_i/\beta}$ if $y_i=0$ (favoring adding high-value items).

The proposal distribution is therefore: $\forall \y_{y_i\to\bar{y_i}}\in\cN_3(\y),\; q_3(\y, \y_{y_i\to\bar{y_i}}) \propto \exp\left(\frac{(1-2y_i)\cdot\theta_i}{\beta}\right)$.
\end{itemize}

\subsection{Data generation}
\label{sec:mkp_data}

For the benchmark experiment in \cref{fig:mkp_benchmark}, we generate a synthetic dataset of $5,000$ instances using the \texttt{PyEPO} library \citep{tang_pyepo_2023}. We set the problem size to $d=100$ items and $J=50$ constraints. For each instance, we sample feature vectors $\x \in \RR^{64}$ and generate the item values $\thetav$ (cost vector) with a polynomial dependence on $\x$ of degree $4$ and multiplicative noise $\epsilon=0.5$. The item weights $w_{i,j}$ are sampled uniformly, and the capacities $C_j$ are generated using a capacity ratio of $0.5$.

To obtain the ground-truth labels $\y_i$ for the conditional learning task, we solve each instance using the Gurobi ILP solver with a time limit of $1000$ms. The dataset is partitioned into training ($80\%$), validation ($10\%$), and test ($10\%$) sets. We use the validation set to select best model iterations (in terms of relative regret on the validation set), before evaluating their test relative regret.

\subsection{Implementation details}
\label{sec:mkp_implementation}

The predictive model $g_W$ is a Multi-Layer Perceptron (MLP) with two hidden layers of size $64$ and ReLU activations. We train the model for $20$ epochs using the Adam \citep{kingma_adam_2017} optimizer with a learning rate of $5 \times 10^{-3}$ and a batch size of $32$.

For the benchmark experiment in \cref{fig:mkp_benchmark}, we use a time limit of $1.0$ms for both the LS-MCMC layer and the Gurobi ILP solver at training time. For the evaluation on the validation and test sets, we ensure a uniform setup by using the Gurobi ILP solver with a fixed time limit of 50ms to output the final solutions for all evaluated methods.

\subsection{Additional results}
\label{sec:mkp_additional_results}
\begin{figure*}[t]
    \centering
    \includegraphics[width=0.99\linewidth]{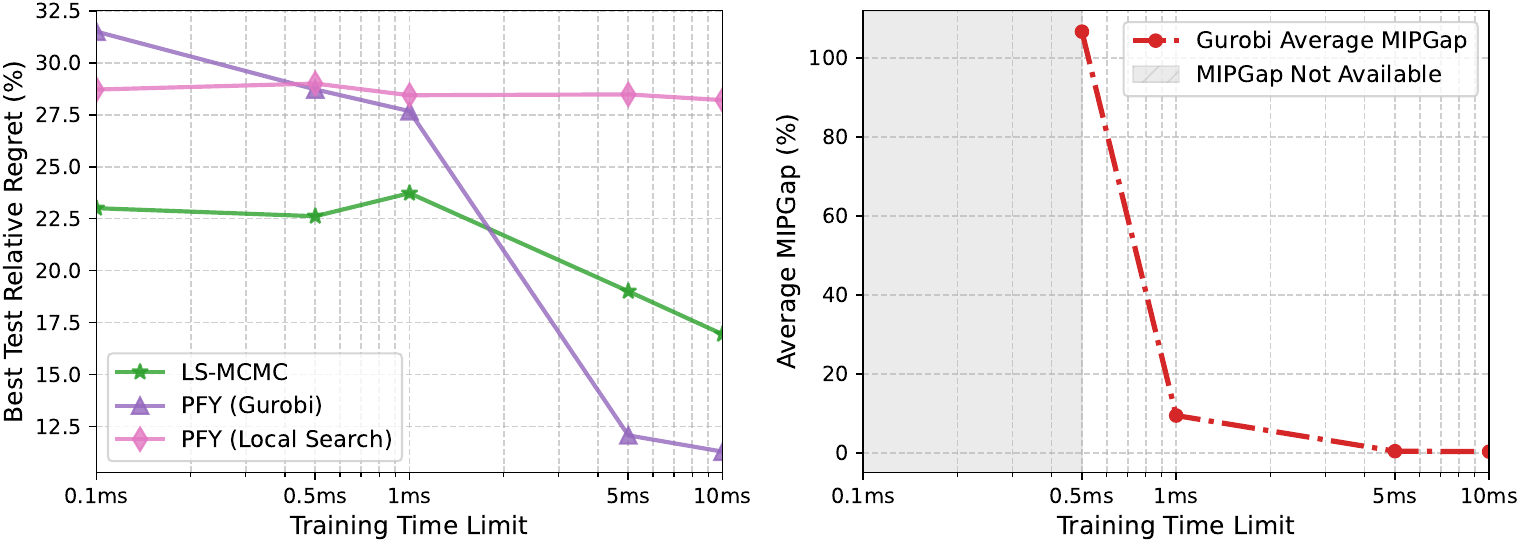}
    \caption{Performance comparison on the MKP across varying training time limits for the combinatorial layer. \textbf{Left:} Best test relative regret for LS-MCMC, PFY using Gurobi, and PFY using the last local search iterate as a black-box oracle. \textbf{Right:} Average MIPGap achieved by the Gurobi solver over the same time limits. At the most restrictive time limit (0.1ms), the MIPGap is unavailable because the solver returns a feasible solution before computing an initial dual bound.}
    \label{fig:mkp_additional}
\end{figure*}

In this section, we perform an extended analysis on the MKP, varying the maximum time limit allocated to the optimization layer during training. We compare our proposed approach against two baselines: PFY \citep{berthet_learning_2020}, using either the Gurobi ILP solver, or the final iterate of our local search as a black-box (inexact) oracle. The results are gathered in \cref{fig:mkp_additional}.
\paragraph{Gradient estimation.} Across all evaluated training time limits, our method consistently achieves a lower test relative regret than the PFY baseline using the final local search iterate. This isolates the solver's impact and shows the superiority of our principled MCMC-based gradient estimator for local search oracles. 
\paragraph{Near-optimal regime.} As the allowed training compute time increases, the Gurobi-based PFY baseline eventually overtakes our approach as the ILP solver is able to perform more thorough exploration, and reaches near-zero MIPGap (therefore returning nearly optimal solutions). Note that for the lowest time limit evaluated (0.1ms), the Gurobi solver's MIPGap is unavailable: the solver terminates before it has sufficient time to compute an initial dual bound, even though it still returns a feasible solution.

\section{Proofs}

\subsection{Proof of convergence to the Gibbs distribution}
\label{proof:stationary_gibbs}
\begin{proof}
    At fixed temperature $t_k = t$, the iterates of \cref{algo:SA} (MH case) follow a time-homogenous Markov chain, defined by the following transition kernel $P_{\bm{\theta}, t}$:

\begin{equation*}
    P_{\bm{\theta}, t}(\bm{y},\bm{y}')= \begin{cases}
        q\left(\bm{y},\bm{y}'\right)\min\left[1, \frac{q(\bm{y}',\bm{y})}{q(\bm{y},\bm{y}')}\exp\left({\frac{\langle \bm{\theta} \,,\, \bm{y}' \rangle + \varphi(\bm{y}') - \langle \bm{\theta} ,\, \bm{y} \rangle - \varphi(\bm{y})}{t}}\right)\right] & \text{if $\bm{y}'\in\mathcal{N}(\bm{y})$,}\\
        1 - \sum_{\bm{y}''\in\mathcal{N}(\bm{y})}P_{\bm{\theta}, t}(\bm{y},\bm{y}'') & \text{if $\bm{y}'=\bm{y}$,}\\
        0 &\text{else.}
    \end{cases}
\end{equation*}

\paragraph{Irreducibility.} As we assumed the neighborhood graph $G_\mathcal{N}$ to be connected and undirected, the Markov Chain is irreducible as we have $\forall \bm{y} \in \mathcal{Y}, \forall \bm{y}' \in \mathcal{N}(\bm{y}), P_{\bm{\theta}, t}(\bm{y}, \bm{y}')>0$.

\paragraph{Aperiodicity.} For simplicity, we directly assumed aperiodicity in the main text. Here, we show that this is a mild condition, which is verified for instance if there is a solution $\bm{y}\in\mathcal{Y}$ such that $q(\bm{y},\bm{y})>0$. Indeed, we then have:

\begin{align*}
    P_{\bm{\theta}, t}(\bm{y}, \bm{y}) &= 1 - \hspace{-1em} \sum_{\bm{y}'\in\mathcal{N}(\bm{y})}P_{\bm{\theta}, t}(\bm{y},\bm{y}')\\
    &= 1 - \sum_{\bm{y}'\in\mathcal{N}(\bm{y})}q\left(\bm{y},\bm{y}'\right)\min\left[1, \frac{q(\bm{y}',\bm{y})}{q(\bm{y},\bm{y}')}\exp\left({\frac{\langle \bm{\theta} \,,\, \bm{y}' \rangle + \varphi(\bm{y}') - \langle \bm{\theta} ,\, \bm{y} \rangle - \varphi(\bm{y})}{t}}\right)\right] \\
    &\geq 1 - \sum_{\bm{y}'\in\mathcal{N}(\bm{y})}q\left(\bm{y},\bm{y}'\right)\\
    &\geq q(\bm{y},\bm{y}')\\
    &>0.
\end{align*}

Thus, we have $P_{\bm{\theta}, t}(\bm{y}, \bm{y}) > 0$, which implies that the chain is aperiodic. As an irreducible and aperiodic Markov Chain on a finite state space, it converges to its stationary distribution and the latter is unique \citep{h_t_convergence_2017}. Finally, one can easily check that the detailed balance equation is satisfied for ${\pi}_{\bm{\theta}, t}$, i.e.:
\begin{align*}
    \forall \bm{y}, \bm{y}' \in \mathcal{Y},\,{\pi}_{\bm{\theta}, t}(\bm{y})P_{\bm{\theta}, t}(\bm{y}, \bm{y}') = {\pi}_{\bm{\theta}, t}(\bm{y}')P_{\bm{\theta}, t}(\bm{y}', \bm{y}),
\end{align*}

giving that ${\pi}_{\bm{\theta}, t}$ is indeed the stationary distribution of the chain, which concludes the proof.
\end{proof}

\subsection{Proof of \cref{prop:layer_properties}}
\label{proof:layer_properties}

\begin{proof}
    Let $\bm{\theta}\in\mathbb{R}^d$ and $t>0$. The fact that $\yt(\bm{\theta})\in\relint(\mathcal{C})=\relint(\conv(\mathcal{Y}))$ follows directly from the fact that $\yt(\bm{\theta})$ is a convex combination of the elements of $\mathcal{Y}$ with positive coefficients, as $\forall \bm{y}\in\mathcal{Y},\, {\pi}_{\bm{\theta},t}(\bm{y})>0$.
    
    \textbf{Low temperature limit.} Let $\bm{y}^\star\coloneqq \argmax_{\bm{y}\in\mathcal{Y}}\langle\bm{\theta},\bm{y}\rangle+\varphi(\bm{y})$. The $\argmax$ is assumed to be single-valued. Let $\bm{y}\in\mathcal{Y}\setminus\{\bm{y}^\star\}$. We have:
    \begin{align*}
       {\pi}_{\bm{\theta}, t}(\bm{y}) &= \frac{\exp\left(\frac{\langle\bm{\theta},\bm{y}\rangle+\varphi(\bm{y})}{t}\right)}{\sum_{\bm{y}'\in\mathcal{Y}}\exp\left(\frac{\langle\bm{\theta},\bm{y}'\rangle+\varphi(\bm{y}')}{t}\right)}\\
       &\leq \frac{\exp\left(\frac{\langle\bm{\theta},\bm{y}\rangle+\varphi(\bm{y})}{t}\right)}{\exp\left(\frac{\langle\bm{\theta},\bm{y}^\star\rangle+\varphi(\bm{y}^\star)}{t}\right)}\\
       &\leq \exp\left(\frac{\left(\langle\bm{\theta},\bm{y}\rangle+\varphi(\bm{y})\right) - \left(\langle\bm{\theta},\bm{y}^\star\rangle+\varphi(\bm{y}^\star)\right)}{t}\right)\\
       &\xrightarrow[t\rightarrow 0^+]{} 0,
    \end{align*}
    as $\langle\bm{\theta},\bm{y}\rangle+\varphi(\bm{y}) <\langle\bm{\theta},\bm{y}^\star\rangle+\varphi(\bm{y}^\star)$ by definition of $\bm{y}^\star$.
    Thus, we have:
    \begin{align*}
        {\pi}_{\bm{\theta}, t}(\bm{y}^\star) = 1-\!\!\!\!\!\!\sum_{\bm{y}\in\mathcal{Y}\setminus\{\bm{y}^\star\}}\!\!\!{\pi}_{\bm{\theta}, t}(\bm{y}) \xrightarrow[t\rightarrow 0^+]{}1.
    \end{align*}
    Thus, the expectation of ${\pi}_{\bm{\theta}, t}$ converges to $\bm{y}^\star$. Naturally, if the $\argmax$ is not unique, the distribution converges to a uniform distribution on the maximizing structures.\newline
    
    \textbf{High temperature limit.}  For all $\bm{y}\in\mathcal{Y}$, we have:
    \begin{align*}
        {\pi}_{\bm{\theta}, t}(\bm{y}) &= \frac{\exp\left(\frac{\langle\bm{\theta},\bm{y}\rangle+\varphi(\bm{y})}{t}\right)}{\sum_{\bm{y}'\in\mathcal{Y}}\exp\left(\frac{\langle\bm{\theta},\bm{y}'\rangle+\varphi(\bm{y}')}{t}\right)}\\
        &\xrightarrow[t\rightarrow \infty]{}\frac{1}{|\mathcal{Y}|},
    \end{align*}
    as $\exp(x/t)\xrightarrow[t\rightarrow \infty]{}1$ for all $x\in\mathbb{R}$. Thus, ${\pi}_{\bm{\theta}, t}$ converges to the uniform distribution on $\mathcal{Y}$, and its expectation converges to the average of all structures.

    \textbf{Expression of the Jacobian.} Let $A_t:\bm{\theta}\mapsto t\cdot\log\sum_{\bm{y}\in\mathcal{Y}}\exp\left(\langle\bm{\theta},\bm{y}\rangle+\varphi(\bm{y})\right)$ be the cumulant function of the exponential family defined by ${\pi}_{\bm{\theta},t}$, scaled by $t$. One can easily check that we have $\nabla_{\bm{\theta}}A_t(\bm{\theta})=\yt(\bm{\theta})$. Thus, we have $J_{\bm{\theta}}\yt(\bm{\theta}) = \nabla^2_{\bm{\theta}}A_t(\bm{\theta})$. However, we also have that the hessian matrix of the cumulant function $\bm{\theta}\mapsto\frac{1}{t}A_t(\bm{\theta})$ is equal to the covariance matrix of the random vector $\frac{Y}{t}$ under ${\pi}_{\bm{\theta},t}$ \citep{wainwright_graphical_2008}. Thus, we have:
    \begin{align*}
    J_{\bm{\theta}}\yt(\bm{\theta}) &= \nabla^2_{\bm{\theta}}A_t(\bm{\theta})\\
    &=t\cdot\nabla^2_{\bm{\theta}}\left(\frac{1}{t}A_t(\bm{\theta})\right) \\
    &= t\cdot \cov_{{\pi}_{\bm{\theta},t}}\left[\frac{Y}{t}\right] \\
    &= \frac{1}{t} \cov_{{\pi}_{\bm{\theta},t}}\left[Y\right].    
    \end{align*}
\end{proof}

\subsection{Proof of \cref{prop:mixture_sa}}
\label{proof:mixture_sa}
\begin{proof}
    Let $K_{\bm{\theta}, t}$ be the Markov transition kernel associated to \cref{algo:mixture_SA}, which can be written as:
    
    \begin{equation*}
    K_{\bm{\theta}, t}(\bm{y},\bm{y}')= \begin{cases}
        \sum_{\substack{s\in Q(\bm{y}) \\ \text{s.t. } q_s(\bm{y}, \bm{y}')>0}} \frac{1}{|Q(\bm{y})|}q_s(\bm{y}, \bm{y}')\min \left(1, \frac{|Q(\bm{y})|}{|Q(\bm{y}')|}\cdot\frac{q_s(\bm{y}', \bm{y}){\pi}_{\thetav,t}(\bm{y}')}{q_s(\bm{y}, \bm{y}'){\pi}_{\thetav,t}(\bm{y})}\right) & \text{if $\bm{y}'\in\bar{\mathcal{N}}(\bm{y})$,}\\
        1 - \sum_{\bm{y}''\in\bar{\mathcal{N}}(\bm{y})}K_{\bm{\theta}, t}(\bm{y},\bm{y}'') & \text{if $\bm{y}'=\bm{y}$,}\\
        0 &\text{else.}
        \end{cases}
    \end{equation*}

    As $\forall \y\in\mathcal{Y},\,\forall\y'\in\bar{\cN}(\y),K_{\bm{\theta}, t}(\bm{y},\bm{y}')>0$, the irreducibility of the chain on $\mathcal{Y}$ is directly implied by the connectedness of $G_{\bar{\cN}}$.
    
    Thus, we only have to check that the detailed balance equation \begin{align*}
        \pi_{\thetav,t}(\bm{y})K_{\bm{\theta}, t}(\bm{y}, \bm{y}') = {\pi}_{\thetav,t}(\bm{y}')K_{\bm{\theta}, t}(\bm{y}', \bm{y})
    \end{align*} is satisfied for all $\bm{y}'\in \bar{\mathcal{N}}(\bm{y})$. We have:
    \begin{align*}
        {\pi}_{\thetav,t}(\bm{y})K_{\bm{\theta}, t}(\bm{y}, \bm{y}') = \sum_{\substack{s\in Q(\bm{y}) \\ \text{s.t. } q_s(\bm{y}, \bm{y}')>0}} \biggl[\frac{q_s(\bm{y}, \bm{y}'){\pi}_{\thetav,t}(\bm{y})}{|Q(\bm{y})|}\min \left(1, \frac{|Q(\bm{y})|}{|Q(\bm{y}')|}\cdot\frac{q_s(\bm{y}', \bm{y}){\pi}_{\thetav,t}(\bm{y}')}{q_s(\bm{y}, \bm{y}'){\pi}_{\thetav,t}(\bm{y})}\right)\biggr].
    \end{align*}

    The main point consists in noticing that the undirectedness assumption for each individual neighborhood graph $G_{\mathcal{N}_s}$ implies:
    \begin{align*}
        \{ s\in Q(\bm{y}): q_s(\bm{y}, \bm{y}')>0 \} = \{ s\in Q(\bm{y}'): q_s(\bm{y}', \bm{y})>0 \}.
    \end{align*}
    Thus, a simple case analysis on how $|Q(\bm{y})|q_s(\bm{y}', \bm{y}){\pi}_{\thetav,t}(\bm{y}')$ and $|Q(\bm{y}')|q_s(\bm{y}, \bm{y}'){\pi}_{\thetav,t}(\bm{y})$ compare allows us to observe that the expression of ${\pi}_{\thetav,t}(\bm{y})K_{\bm{\theta}, t}(\bm{y}, \bm{y}')$ is symmetric in $\bm{y}$ and $\bm{y}'$, which concludes the proof.
\end{proof}

\subsection{Proof of the claims of \cref{sec:associated_fy_loss}}
\label{proof:fyl}

\begin{proof}
    \textbf{Strict convexity of $\Omega_t$ on $\relint(\cC)$.}
    As $A_t$ is a differentiable convex function on $\mathbb{R}^d$ (as the log-sum-exp of such functions), it is an essentially smooth closed proper convex function. Thus, it is such that
    \begin{align*}
        \relint\left(\dom((A_t)^*)\right)\subseteq \nabla A_t(\mathbb{R}^d)\subseteq \dom((A_t)^*),
    \end{align*}
    and we have that the restriction of $(A_t)^*$ to $\nabla A_t(\mathbb{R}^d)$ is strictly convex on every convex subset of $\nabla A_t(\mathbb{R}^d)$ (corollary 26.4.1 in \citet{rockafellar_1970}). As the range of the gradient of the cumulant function $\bm{\theta}\mapsto A_t(\bm{\theta})/t$ is exactly the relative interior of the marginal polytope $\conv\left(\{ \bm{y}/t, \bm{y}\in\mathcal{Y} \}\right)$ (see appendix B.1 in \citet{wainwright_graphical_2008}), and $(A_t)^* \eqqcolon \Omega_t$, we actually have that $$\relint\left(\dom(\Omega_t)\right)\subseteq \relint(\mathcal{C}) \subseteq \dom(\Omega_t),$$ and that $\Omega_t$ is strictly convex on every convex subset of $\relint(\mathcal{C})$, i.e., strictly convex on $\relint(\mathcal{C})$ (as $\relint(\mathcal{C})$ is itself convex).
    
    \textbf{Regularized optimization perspective.} As $A_t$ is closed proper convex, it is equal to its biconjugate by the Fenchel-Moreau theorem. Thus, we have:
    $$A_t(\bm{\theta})=\sup_{\bm{\mu}\in\mathbb{R}^d}\left\{\langle\bm{\theta},\bm{\mu}\rangle - (A_t)^*(\bm{\mu})\right\} = \sup_{\bm{\mu}\in\mathbb{R}^d}\left\{\langle\bm{\theta},\bm{\mu}\rangle -\Omega_t(\bm{\mu})\right\}.$$
    Moreover, as $\nabla A_t(\mathbb{R}^d)=\relint(\mathcal{C})$, we have $||\nabla A_t(\bm{\theta})||\leq R_\mathcal{C}\coloneqq\max_{\bm{\mu}\in\mathcal{C}}||\bm{\mu}||$, which gives $\dom(\Omega_t)\subset B(\bm{0}, R_\mathcal{C})$. Thus we can actually write:
    $$A_t(\bm{\theta}) = \sup_{\bm{\mu}\in B(\bm{0}, R_\mathcal{C})}\left\{\langle\bm{\theta},\bm{\mu}\rangle -\Omega_t(\bm{\mu})\right\},$$
    and now apply Danskin's theorem as $B(\bm{0}, R_\mathcal{C})$ is compact, which further gives:
    $$
    \partial A_t(\bm{\theta}) = \argmax_{\bm{\mu}\in B(\bm{0}, R_\mathcal{C})}\left\{\langle\bm{\theta},\bm{\mu}\rangle -\Omega_t(\bm{\mu})\right\},
    $$
    and the fact that $A_t$ is differentiable gives that both sides are single-valued. Moreover, as $\nabla A_t(\mathbb{R}^d)=\relint(\mathcal{C})$, we know that the right hand side is maximized in $\mathcal{C}$, and we can actually write:
    $$
    \nabla A_t(\bm{\theta}) = \argmax_{\bm{\mu}\in \mathcal{C}}\left\{\langle\bm{\theta},\bm{\mu}\rangle -\Omega_t(\bm{\mu})\right\}.
    $$
    Finally, a simple calculation yields $\nabla A_t(\bm{\theta}) =\mathbb{E}_{{\pi}_{\bm{\theta},t}}\left[Y\right]=\yt(\bm{\theta})$. The expression of $\nabla_{\bm{\theta}}\ell_t(\bm{\theta}\,;\bm{y})$ also follows.
\end{proof}

\begin{remark}
    The proposed Fenchel-Young loss can also be obtained via distribution-space regularization. Let \text{$s_{\bm{\theta}} \coloneqq \left(\langle \bm{\theta} \,,\,\bm{y}\rangle + \varphi(\bm{y})\right)_{\bm{y}\in\mathcal{Y}}\in\mathbb{R}^{|\mathcal{Y}|}$} be a vector containing the score of all structures, and denote by $L_{-tH}:\mathbb{R}^{|\mathcal{Y}|}\times\Delta^{|\mathcal{Y}|}\rightarrow \mathbb{R}$ the Fenchel-Young loss generated by $-tH$, where $H$ is the Shannon entropy. We have \text{$\nabla_{s_{\bm{\theta}}}(-tH)^{*}(s_{\bm{\theta}}) = {\pi}_{\bm{\theta}, t}$}. The chain rule further gives \text{$\nabla_{\bm{\theta}}(-tH)^{*}(s_{\bm{\theta}})=\mathbb{E}_{{\pi}_{\bm{\theta},t}}\left[Y\right]$}. Thus, we have \text{$\nabla_{\bm{\theta}}L_{-tH}(s_{\bm{\theta}}\,;\bm{p}_{\bm{y}}) = \nabla_{\bm{\theta}}\ell_t (\bm{\theta}\,;\bm{y})$}, where $\bm{p}_{\bm{y}}$ is the Dirac distribution on $\bm{y}$. In the case where $\varphi \equiv 0$ and $t=1$, we have \text{$\Omega_t (\bm{\mu})= - \left( \max_{\bm{p}\in\Delta^{|\mathcal{Y}|}} H (\bm{p}) \text{ s.t. } \mathbb{E}_{\bm{p}}\left[Y\right]=\bm{\mu} \right)$} \citep{blondel_learning_2020}, and $\ell_t$ is known as the CRF loss \citep{lafferty_crf_2001}.
\end{remark}

\subsection{Proof of \cref{prop:1step_fyl_sup}}
\label{proof:1step_fyl_sup}
\begin{proof}
    The distribution of the first iterate of the Markov chain with transition kernel defined in \cref{eq:markov_kernel} and initialized at the ground-truth structure $\y$ is given by:
    \begin{align*}
    (\pfirstsup)(\bm{y}') &= P_{\bm{\theta},t}(\y,\y')\\
    &=
    \begin{cases}
    q(\y,\bm{y}')\min\left[1, \frac{q(\bm{y}',\y)}{q(\y,\bm{y}')}\exp\left(\left[\langle\bm{\theta}, \bm{y}'-\y\rangle + \varphi(\bm{y}')-\varphi(\y)\right]/t\right)\right] &\text{if $\bm{y}'\in\mathcal{N}(\y)$,}\\
     1- \sum_{\bm{y}''\in\mathcal{N}(\y)} (\pfirstsup)(\bm{y}'') &\text{if $\bm{y}'=\y$,}\\
    0 &\text{else.}
    \end{cases}
\end{align*}
Let $\alpha_{\y}(\bm{\theta},\bm{y}') \coloneqq \frac{q(\bm{y}',\y)}{q(\y,\bm{y}')}\exp\left(\left[\langle\bm{\theta}, \bm{y}'-\y\rangle + \varphi(\bm{y}')-\varphi(\y)\right]/t\right)$. Define also the following sets:
\begin{align*}
    \mathcal{N}_{\y}^-(\bm{\theta}) = \left\{\bm{y}'\in\mathcal{N}(\y)\mid \alpha_{\y}(\bm{\theta},\bm{y}')\leq 1 \right\},\quad \mathcal{N}_{\y}^+(\bm{\theta}) = \left\{\bm{y}'\in\mathcal{N}(\y)\mid \alpha_{\y}(\bm{\theta},\bm{y}')> 1 \right\}.
\end{align*}
The expectation of the first iterate is then given by:
\begin{align*}
    \mathbb{E}_{\pfirstsup}\left[Y\right] &= \sum_{\bm{y}'\in\mathcal{N}(\y)}  (\pfirstsup)(\bm{y}')\cdot\bm{y}' + \left( 1- \!\!\!\!\sum_{\bm{y}''\in\mathcal{N}(\y)} (\pfirstsup)(\bm{y}'')\right)\cdot\y\\
    &= \y+\!\!\!\!\sum_{\bm{y}'\in\mathcal{N}(\y)} \!\! (\pfirstsup)(\bm{y}')\cdot(\bm{y}'-\y)\\
    &= \y +\!\!\!\!\!\!\sum_{\bm{y}'\in\mathcal{N}_{\y}^-(\bm{\theta})}\!\!q(\bm{y}',\y)\exp\left(\left[\langle\bm{\theta}, \bm{y}'-\y\rangle + \varphi(\bm{y}')-\varphi(\y)\right]/t\right)\cdot\left(\bm{y}'-\y\right) \; +\!\!\!\!\!\sum_{\bm{y}'\in\mathcal{N}_{\y}^+(\bm{\theta})}\!\!q(\y,\bm{y}')\cdot\left(\bm{y}'-\y\right).
\end{align*}
Let now $f_{\y}:\mathbb{R}^d\times\mathcal{N}(\y)\rightarrow\mathbb{R}$ be defined as:
\begin{align*}
    f_{\y}:(\bm{\theta}\,;\bm{y}')\mapsto \begin{cases}
        t\cdot q(\bm{y}',\y)\exp\left(\left[\langle\bm{\theta}, \bm{y}'-\y\rangle + \varphi(\bm{y}')-\varphi(\y)\right]/t\right) &\text{if $\alpha_{\y}(\bm{\theta},\bm{y}')\leq1$,}\\
        t\cdot q(\y,\bm{y}')\left(\left[\langle\bm{\theta}, \bm{y}'-\y\rangle + \varphi(\bm{y}')-\varphi(\y)\right]/t+1 -\log\frac{q(\y,\bm{y}')}{q(\bm{y}',\y)}\right) &\text{if $\alpha_{\y}(\bm{\theta},\bm{y}')>1$.}
    \end{cases}
\end{align*}

Let $F_{\y}:\bm{\theta}\mapsto \langle \bm{\theta},\y\rangle +\!\sum_{\bm{y}'\in\mathcal{N}(\y)}f_{\y}(\bm{\theta}\,;\bm{y}')$. We define the target-dependent regularization function $\Omega_{\y}$ and the corresponding Fenchel-Young loss as: $$\Omega_{\y}:\bm{\mu}\mapsto (F_{\y})^*(\bm{\mu}),\quad\quad
\ell_{\Omega_{\y}}(\bm{\theta}\,;\y) \coloneqq (\Omega_{\y})^*(\bm{\theta}) + \Omega_{\y}(\y) - \langle \bm{\theta},\y\rangle.
$$

\begin{itemize}
    \item \underline{$\Omega_{\y}$ is $t/\mathbb{E}_{q(\y,\,\cdot\,)}||Y-\y||_2^2$-strongly convex:}
\end{itemize}

One can easily check that $f_{\y}(\,\cdot\,;\bm{y}')$ is continuous for all $\bm{y}'\in\mathcal{N}(\y)$, as it is defined piecewise as continuous functions that match on the junction affine hyperplane defined by: $$\left\{\bm{\theta}\in\mathbb{R}^d\mid\alpha_{\y}(\bm{\theta};\bm{y}')=1\right\} = \left\{ \bm{\theta}\in\mathbb{R}^d\mid\langle\bm{\theta},\bm{y}'-\y\rangle = t\log\frac{q(\y,\bm{y}')}{q(\bm{y}',\y)}+\varphi(\y)-\varphi(\bm{y}')\right\}.$$  Moreover, we have that $f_{\y}(\,\cdot\,;\bm{y}')$ is actually differentiable everywhere as its gradient can be continuously extended to the junction affine hyperplane with constant value equal to $q(\y,\bm{y}')(\bm{y}'-\y)$.
We now show that $f_{\y}(\,\cdot\,;\bm{y}')$ is $\frac{1}{t}q(\y,\bm{y}') \cdot ||\bm{y}'-\y||^2$-smooth. Indeed, it is defined as the composition of the linear form $\bm{\theta}\mapsto \langle\bm{\theta},\bm{y}'-\y\rangle$ and the function $g:\mathbb{R}\rightarrow\mathbb{R}$ given by:
    $$
    g:x\mapsto\begin{cases}
            t\cdot q(\bm{y}',\y)\exp\left(\left[x+\varphi(\bm{y}')-\varphi(\y)\right]/t\right) &\text{if $x\leq t\log\frac{q(\y,\bm{y}')}{q(\bm{y}',\y)}+\varphi(\y)-\varphi(\bm{y}')$,}\\
            t\cdot q(\y,\bm{y}')\left(\left[x+\varphi(\bm{y}')-\varphi(\y)\right]/t +1 -\log\frac{q(\y,\bm{y}')}{q(\bm{y}',\y)}\right) &\text{if $x>t\log\frac{q(\y,\bm{y}')}{q(\bm{y}',\y)}+\varphi(\y)-\varphi(\bm{y}')$.}
        \end{cases}
    $$
We begin by showing that $g$ is $\frac{1}{t}q(\y,\bm{y}')$-smooth. We have:
    $$
    g':x\mapsto\begin{cases}
            q(\bm{y}',\y)\exp\left(\left[x+\varphi(\bm{y}')-\varphi(\y)\right]/t\right) &\text{if $x\leq t\log\frac{q(\y,\bm{y}')}{q(\bm{y}',\y)}+\varphi(\y)-\varphi(\bm{y}')$,}\\
            q(\y,\bm{y}')&\text{if $x>t\log\frac{q(\y,\bm{y}')}{q(\bm{y}',\y)}+\varphi(\y)-\varphi(\bm{y}')$.}
        \end{cases}
    $$
    Thus, $g'$ is continuous, and differentiable everywhere except in $x_0\coloneqq t\log\frac{q(\y,\bm{y}')}{q(\bm{y}',\y)}+\varphi(\y)-\varphi(\bm{y}')$. Its derivative is given by:
    $$
    g'':x\mapsto\begin{cases}
            \frac{1}{t}q(\bm{y}',\y)\exp\left(\left[x+\varphi(\bm{y}')-\varphi(\y)\right]/t\right) &\text{if $x\leq t\log\frac{q(\y,\bm{y}')}{q(\bm{y}',\y)}+\varphi(\y)-\varphi(\bm{y}')$,}\\
            0&\text{if $x>t\log\frac{q(\y,\bm{y}')}{q(\bm{y}',\y)}+\varphi(\y)-\varphi(\bm{y}')$.}
        \end{cases}
    $$
    \begin{itemize}
        \item For $x_1, x_2\leq x_0$, we have:
        \begin{align*}
            |g'(x_1)-g'(x_2)| &\leq |x_1-x_2|\sup_{x\in\left]-\infty, x_0\right[}|g''(x)| \\
            &= |x_1-x_2|\lim_{\substack{x\rightarrow x_0 \\ x<x_0}}|g''(x)| \\
            &= \frac{1}{t}q(\y,\bm{y}')\cdot |x_1-x_2|.
        \end{align*}
        \item For $x_1, x_2\geq x_0$, we trivially have $|g'(x_1)-g'(x_2)| = 0$.
        \item For $x_1\leq x_0 \leq x_2$, we have:
        \begin{align*}
            |g'(x_1)-g'(x_2)| &= |(g'(x_1)- g'(x_0)) - (g'(x_2)-g'(x_0))|\\
            &\leq |g'(x_1)-g'(x_0)| + |g'(x_2)-g'(x_0)| \\
            &\leq \frac{1}{t}q(\y,\bm{y}')\cdot |x_1-x_0| \\
            &\leq \frac{1}{t}q(\y,\bm{y}')\cdot |x_1-x_2|.
        \end{align*}
    \end{itemize}
    Thus, we have:
    $$
    \forall x_1,x_2\in\mathbb{R}, |g'(x_1)-g'(x_2)|\leq \frac{1}{t}q(\y,\bm{y}')\cdot|x_1-x_2|,
    $$
    and $g$ is $\frac{1}{t}q(\y,\bm{y}')$-smooth. Nevertheless, we have $f_{\y}(\,\cdot\,,\bm{y}') = g(\langle\,\cdot\,,\bm{y}'-\y\rangle)$. Thus, we have, for $\bm{\theta}_1, \bm{\theta}_2\in\mathbb{R}^d$:
    \begin{align*}
        ||\nabla_{\bm{\theta}} f_{\y}(\bm{\theta_1},\bm{y}') - \nabla_{\bm{\theta}} f_{\y}(\bm{\theta_2},\bm{y}')|| &= || g'(\langle\bm{\theta}_1,\bm{y}'-\y\rangle)(\bm{y}'-\y) - g'(\langle\bm{\theta}_2,\bm{y}'-\y\rangle)(\bm{y}'-\y) || \\
        &= |g'(\langle\bm{\theta}_1,\bm{y}'-\y\rangle) - g'(\langle\bm{\theta}_2,\bm{y}'-\y\rangle)|\cdot ||\bm{y}'-\y|| \\
        &\leq \frac{1}{t}q(\y,\bm{y}') \cdot|\langle\bm{\theta}_1,\bm{y}'-\y\rangle-\langle\bm{\theta}_2,\bm{y}'-\y\rangle|\cdot ||\bm{y}'-\y|| \\
        &\leq \frac{1}{t}q(\y,\bm{y}') \cdot ||\bm{y}'-\y||^2\cdot ||\bm{\theta}_1-\bm{\theta}_2||,
    \end{align*}
    and $f_{\y}(\,\cdot\,,\bm{y}')$ is $\frac{1}{t}q(\y,\bm{y}') \cdot ||\bm{y}'-\y||^2$-smooth. Thus, recalling that $F_{\y}$ is defined as $$F_{\y}:\bm{\theta}\mapsto \langle\bm{\theta},\y\rangle+\sum_{\bm{y}'\in\mathcal{N}(\y)}f_{\y}(\bm{\theta};\bm{y}'),$$ we have that $F_{\y}$ is $\sum_{\bm{y}'\in\mathcal{N}(\y)}\frac{1}{t}q(\y,\bm{y}') \cdot ||\bm{y}'-\y||^2 = \mathbb{E}_{q(\y,\,\cdot\,)}||Y-\y||_2^2/t$-smooth. Finally, as $\Omega_{\y} \coloneqq (F_{\y})^*$, Fenchel duality theory gives that $\Omega_{\y}$ is $t/\mathbb{E}_{q(\y,\,\cdot\,)}||Y-\y||_2^2$-strongly convex.

\begin{itemize}
    \item \underline{$\mathbb{E}_{\pfirstsup}\left[Y\right] = \argmax_{\bm{\mu}\in\conv\left(\mathcal{N}(\y)\cup\{\y\}\right)}\left\{\langle\bm{\theta},\bm{\mu}\rangle-\Omega_{\y}(\bm{\mu})\right\}$:}
\end{itemize}

Noticing that $g$ is continuous on $\mathbb{R}$, convex on $\left]-\infty,t\log\frac{q(\y,\bm{y}')}{q(\bm{y}',\y)}+\varphi(\y)-\varphi(\bm{y}')\right[$ and on $\left]t\log\frac{q(\y,\bm{y}')}{q(\bm{y}',\y)}+\varphi(\y)-\varphi(\bm{y}'),+\infty\right[$, and with matching derivatives on the junction: $$g'(t)\xrightarrow[t< t\log\frac{q(\y,\bm{y}')}{q(\bm{y}',\y)}+\varphi(\y)-\varphi(\bm{y}')]{t \rightarrow t\log\frac{q(\y,\bm{y}')}{q(\bm{y}',\y)}+\varphi(\y)-\varphi(\bm{y}')}q(\y,\bm{y}'),\quad g'(t)\xrightarrow[t> t\log\frac{q(\y,\bm{y}')}{q(\bm{y}',\y)}+\varphi(\y)-\varphi(\bm{y}')]{t \rightarrow t\log\frac{q(\y,\bm{y}')}{q(\bm{y}',\y)}+\varphi(\y)-\varphi(\bm{y}')}q(\y,\bm{y}'),$$
gives that $g$ is convex on $\mathbb{R}$. Thus, $f_{\y}(\,\cdot\,;\bm{y}')$ is convex on $\mathbb{R}^d$ by composition. Thus, $$F_{\y}:\bm{\theta}\mapsto \langle \bm{\theta},\y\rangle +\!\sum_{\bm{y}'\in\mathcal{N}(\y)}f_{\y}(\bm{\theta};\bm{y}')$$ is closed proper convex as the sum of such functions. The Fenchel-Moreau theorem then gives that it is equal to its biconjugate. Thus, we have:
$$F_{\y}(\bm{\theta})=\sup_{\bm{\mu}\in\mathbb{R}^d}\left\{\langle\bm{\theta},\bm{\mu}\rangle - (F_{\y})^*(\bm{\mu})\right\} = \sup_{\bm{\mu}\in\mathbb{R}^d}\left\{\langle\bm{\theta},\bm{\mu}\rangle -\Omega_{\y}(\bm{\mu})\right\}.$$

Nonetheless, the gradient of $F_{\y}$ is given by:
\begin{align*}
    \nabla_{\bm{\theta}}F_{\y}(\bm{\theta}) &= \y +\!\!\!\!\!\sum_{\bm{y}'\in\mathcal{N}_{\y}^-(\bm{\theta})}q(\bm{y}',\y)\exp\left(\left[\langle\bm{\theta}, \bm{y}'-\y\rangle + \varphi(\bm{y}')-\varphi(\y)\right]/t\right)\cdot\left(\bm{y}'-\y\right) \; +\!\!\!\!\!\sum_{\bm{y}'\in\mathcal{N}_{\y}^+(\bm{\theta})}q(\y,\bm{y}')\cdot\left(\bm{y}'-\y\right) \\
    &= \mathbb{E}_{\pfirstsup}\left[Y\right].
\end{align*}

Thus, we have $\nabla F_{\y}(\mathbb{R}^d)\subset\conv\left(\mathcal{N}(\y)\cup\{\y\}\right) $, which gives: $$\forall\bm{\theta}\in\mathbb{R}^d,\, ||\nabla F_{\y}(\bm{\theta})||\leq R_{\mathcal{N}(\y)}\coloneqq\max_{\bm{\mu}\in\conv\left(\mathcal{N}(\y)\cup\{\y\}\right)}||\bm{\mu}||,$$ so that we have $\dom(\Omega_{\y})\subset B(\bm{0}, R_{\mathcal{N}(\y)})$. Thus we can actually write:
$$F_{\y}(\bm{\theta}) = \sup_{\bm{\mu}\in B(\bm{0}, R_{\mathcal{N}(\y)})}\left\{\langle\bm{\theta},\bm{\mu}\rangle -\Omega_{\y}(\bm{\mu})\right\},$$
and now apply Danskin's theorem as $B(\bm{0}, R_{\mathcal{N}(\y)})$ is compact, which further gives:
$$
\partial F_{\y}(\bm{\theta}) = \argmax_{\bm{\mu}\in B(\bm{0}, R_{\mathcal{N}(\y)})}\left\{\langle\bm{\theta},\bm{\mu}\rangle -\Omega_{\y}(\bm{\mu})\right\},
$$
and the fact that $F_{\y}$ is differentiable gives that both sides are single-valued. Moreover, as $\nabla F_{\y}(\mathbb{R}^d)\subset \conv\left(\mathcal{N}(\y)\cup\{\y\}\right)$, we know that the right hand side is maximized in $\conv\left(\mathcal{N}(\y)\cup\{\y\}\right)$, and we can actually write:
$$
\mathbb{E}_{\pfirstsup}\left[Y\right]= \nabla F_{\y}(\bm{\theta}) = \argmax_{\bm{\mu}\in \conv\left(\mathcal{N}(\y)\cup\{\y\}\right)}\left\{\langle\bm{\theta},\bm{\mu}\rangle -\Omega_{\y}(\bm{\mu})\right\}.
$$

\begin{itemize}
    \item \underline{Smoothness of $\ell_{\Omega_{\y}}(\,\cdot\,;\y)$ and expression of its gradient:}
\end{itemize}

Based on the above, we have:
$$
\ell_{\Omega_{\y}}(\bm{\theta}\,;\y) = F_{\y}(\bm{\theta}) + \Omega_{\y}(\y) - \langle \bm{\theta},\y\rangle.
$$
Thus, the  $\mathbb{E}_{q(\y,\,\cdot\,)}||Y-\y||_2^2/t$-smoothness of $\ell_{\Omega_{\y}}(\,\cdot\,;\y)$ follows directly from the previously established $\mathbb{E}_{q(\y,\,\cdot\,)}||Y-\y||_2^2/t$-smoothness of $F_{\y}$. Similarly, the expression of $\nabla_{\bm{\theta}}\ell_{\Omega_{\y}}(\bm{\theta}\,;\y)$ follows from the previously established expression of $\nabla_{\bm{\theta}}F_{\y}(\bm{\theta})$, and we have: $$\nabla_{\bm{\theta}}\ell_{\Omega_{\y}}(\bm{\theta}\,;\y)= \nabla_{\bm{\theta}}F_{\y}(\bm{\theta})-\y=\mathbb{E}_{\pfirstsup}\left[Y\right]-\y.$$

\end{proof}

\subsection{Proof of \cref{prop:asymp_normality}}
\label{proof:asymp}
\begin{proof}
    The proof is exactly the proof of Proposition 4.1 in \citet{berthet_learning_2020}, in which the setting is similar, and all the same arguments hold (we also have that ${\pi}_{\bm{\theta}_0,t}$ has full support on $\mathcal{Y}$, giving $\bar{Y}_N\in\relint(\mathcal{C})$ for $N$ large enough). The only difference is the choice of regularization function, and we have to prove that it is also convex and smooth in our case. While the convexity of $\Omega_t$ is directly implied by its definition as a Fenchel conjugate, the fact that is is smooth is due to Theorem 26.3 in \citet{rockafellar_1970} and the essential strict convexity of $A_t$ (which is itself closed proper convex). The latter relies on the fact that $\mathcal{C}$ is assumed to be of full-dimension (otherwise $A_t$ would be linear when restricted to any affine subspace of direction equal to the subspace orthogonal to the direction of the smallest affine subspace spanned by $\mathcal{C}$), which in turn implies that $A_t$ is strictly convex on $\mathbb{R}^d$.  Thus, Proposition 4.1 in \citet{berthet_learning_2020} gives the asymptotic normality:
$$
\sqrt{N}(\bm{\theta}^\star_N - \bm{\theta}_0) \xrightarrow[N\rightarrow \infty]{\mathcal{D}}
\mathcal{N}\left(\bm{0},\,\left(\nabla^2_{\bm{\theta}}A_t(\bm{\theta}_0)\right)^{-1}\cov_{{\pi}_{\bm{\theta}_0,t}}\left[Y\right]\left(\nabla^2_{\bm{\theta}}A_t(\bm{\theta}_0)\right)^{-1}\right).
$$
\noindent Moreover, we already derived $\nabla^2_{\bm{\theta}}A_t(\bm{\theta}_0)=\frac{1}{t}\cov_{{\pi}_{\bm{\theta}_0,t}}\left[Y\right]$ in \cref{proof:layer_properties}, leading to the simplified asymptotic normality given in the proposition.
    
\end{proof}

\subsection{Proof of \cref{prop:stochastic_gradient_estimate}}
\label{proof:stochastic_gradient_estimate}

\begin{proof}
The proof consists in bounding the convergence rate of the Markov chain $\left(\bm{y}^{(k)}\right)_{k\in\mathbb{N}}$ (which has transition kernel $P_{\bm{\theta}, t}$) for all $\bm{\theta}$, in order to apply Theorem 4.1 in \citet{younes_stochastic_1998}. It is defined as the smallest constant $\lambda_{\bm{\theta}}$ such that:

$$
\exists A>0\,:\;\forall \bm{y}\in\mathcal{Y},\,|\mathbb{P}(\bm{y}^{(k)} = \bm{y}) - {\pi}_{\bm{\theta},t}(\bm{y})|\leq A\lambda_{\bm{\theta}}^k.
$$

More precisely, we must find a constant $D$ such that \text{$\exists B>0\,:\;\lambda_{\bm{\theta}} \leq 1 - Be^{-D||\bm{\theta}||}$}, in order to impose \text{$K_{n+1} > \left\lfloor 1 + a'\exp\left(2D||\hat{\bm{\theta}}_n||\right) \right\rfloor$}.\newline

A known result gives \text{$\lambda_{\bm{\theta}} \leq \rho(\bm{\theta})$} with \text{$\rho(\bm{\theta}) = \max_{\lambda\in S_{\bm{\theta}}\backslash\{1\}}|\lambda|$} \citep{madras_markov_2002}, where $S_{\bm{\theta}}$ is the spectrum of the transition kernel $P_{\bm{\theta}, t}$ (in this context, $1-\rho(\bm{\theta})$ is known as the \emph{spectral gap} of the Markov chain). To bound $\rho(\bm{\theta})$, we use the results of \citet{ingrassia_rate_1994}, which study the Markov chain with transition kernel $P^{\prime}_{\bm{\theta}, t}$, such that $P_{\bm{\theta}, t} = \frac{1}{2}\left(I + P^{\prime}_{\bm{\theta}, t}\right)$. It corresponds to the same algorithm, but with a proposal distribution $q'$ defined as:

\begin{align*}
    q'\left(\bm{y}, \bm{y}'\right)=
    \begin{cases}
        \frac{1}{d^*}  & \text{if $\bm{y}' \in \mathcal{N}(\bm{y})$,}\\
        1-\frac{d(\bm{y})}{d^*}  & \text{if $\bm{y}' = \bm{y}$,}\\
        0 &\text{else.}
    \end{cases}
\end{align*}

As $P^{\prime}_{\bm{\theta}, t}$ is a row-stochastic matrix, Gershgorin's circle theorem gives that its spectrum is included in the complex unit disc. Moreover, one can easily check that the associated Markov chain is also reversible with respect to ${\pi}_{\bm{\theta},t}$, and the corresponding detailed balance equation gives:

$$
\forall \bm{y}, \bm{y}' \in \mathcal{Y}, \; {\pi}_{\bm{\theta},t}(\bm{y})P^{\prime}_{\bm{\theta}, t}(\bm{y}, \bm{y}') = {\pi}_{\bm{\theta},t}(\bm{y}')P^{\prime}_{\bm{\theta}, t}(\bm{y}', \bm{y}),
$$

which is equivalent to:

$$
\forall \bm{y}, \bm{y}' \in \mathcal{Y}, \; \sqrt{\frac{{\pi}_{\bm{\theta},t}(\bm{y})}{{\pi}_{\bm{\theta},t}(\bm{y}')}}P^{\prime}_{\bm{\theta}, t}(\bm{y}, \bm{y}') = \sqrt{\frac{{\pi}_{\bm{\theta},t}(\bm{y}')}{{\pi}_{\bm{\theta},t}(\bm{y})}}P^{\prime}_{\bm{\theta}, t}(\bm{y}', \bm{y})
$$

as ${\pi}_{\bm{\theta},t}$ has full support on $\mathcal{Y}$, which can be further written in matrix form as:

$$
\Pi_{\bm{\theta}}^{1/2} P^{\prime}_{\bm{\theta}, t} \Pi_{\bm{\theta}}^{-1/2} = \Pi_{\bm{\theta}}^{-1/2} P^{\prime \top}_{\bm{\theta}, t} \Pi_{\bm{\theta}}^{1/2},
$$

where $\Pi_{\bm{\theta}} = \diag ({\pi}_{\bm{\theta},t})$. Thus, the matrix $\Pi_{\bm{\theta}}^{1/2} P^{\prime}_{\bm{\theta}, t} \Pi_{\bm{\theta}}^{-1/2}$ is symmetric, and the spectral theorem ensures its eigenvalues are real. As it is similar to the transition kernel $P^{\prime}_{\bm{\theta}, t}$ (with change of basis matrix $\Pi_{\bm{\theta}}^{-1/2}$), they share the same spectrum $S'_{\bm{\theta}}$, and we have $S'_{\bm{\theta}}\subset \left[-1, 1\right]$. Let us order $S'_{\bm{\theta}}$ as $-1\leq \lambda'_{\min} \leq \dots \leq \lambda'_2 \leq \lambda'_1 = 1$. As $P_{\bm{\theta}, t} = \frac{1}{2}\left(I + P^{\prime}_{\bm{\theta}, t}\right)$, we clearly have $\rho(\bm{\theta}) = \frac{1+\lambda'_2}{2}$. Thus, we can use Theorem 4.1 of \citet{ingrassia_rate_1994}, which gives \text{$\lambda'_2 \leq 1- G\cdot Z(\bm{\theta})\exp(-m\left(\bm{\theta})\right)$} (we keep their notations for $Z$ and $m$, and add the dependency in $\bm{\theta}$ for clarity), where $G$ is a constant depending only on the graph $G_\mathcal{N}$, and with:

\begin{align*}
    Z(\bm{\theta}) &= \sum_{\bm{y}\in\mathcal{Y}}\exp\left(\frac{\langle\bm{\theta},\,\bm{y}\rangle + \varphi(\bm{y})}{t} - \max_{\bm{y}'\in\mathcal{Y}}\left[\frac{\langle\bm{\theta},\,\bm{y}'\rangle+ \varphi(\bm{y}')}{t}\right]\right)\\
    &\geq |\mathcal{Y}| \exp\left(\frac{1}{t}\left[\min_{\bm{y}\in\mathcal{Y}}\langle\bm{\theta},\,\bm{y}\rangle + \min_{\bm{y}\in\mathcal{Y}}\varphi(\bm{y})  - \max_{\bm{y}'\in\mathcal{Y}}\langle\bm{\theta},\,\bm{y}'\rangle - \max_{\bm{y}'\in\mathcal{Y}}\varphi(\bm{y}')\right]\right)\\
    &\geq |\mathcal{Y}| \exp\left(-\frac{2R_\mathcal{C}}{t}||\bm{\theta}||-\frac{2R_\varphi}{t}\right),
\end{align*}

and:

\begin{align*}
m(\bm{\theta}) &\leq \max_{\bm{y}\in\mathcal{Y}}\biggl\{\max_{\bm{y}'\in\mathcal{Y}}\left[\frac{\langle\bm{\theta},\,\bm{y}'\rangle+ \varphi(\bm{y}')}{t}\right]  - \frac{\langle \bm{\theta}, \bm{y}\rangle + \varphi(\bm{y})}{t} \biggr\} - 2 \min_{\bm{y}\in\mathcal{Y}}\biggl\{\max_{\bm{y}'\in\mathcal{Y}}\left[\frac{\langle\bm{\theta},\,\bm{y}'\rangle+ \varphi(\bm{y}')}{t}\right]  - \frac{\langle \bm{\theta}, \bm{y}\rangle + \varphi(\bm{y})}{t} \biggr\} \\
&=\max_{\bm{y}'\in\mathcal{Y}}\left[\frac{\langle\bm{\theta},\,\bm{y}'\rangle+ \varphi(\bm{y}')}{t}\right] - \min_{\bm{y}\in\mathcal{Y}}\left[\frac{\langle\bm{\theta},\,\bm{y}\rangle+ \varphi(\bm{y})}{t}\right]\\
&\leq \frac{1}{t}\left(\max_{\bm{y}'\in\mathcal{Y}}\langle\bm{\theta},\,\bm{y}'\rangle + \max_{\bm{y}'\in\mathcal{Y}}\varphi(\bm{y}')  - \min_{\bm{y}\in\mathcal{Y}}\langle\bm{\theta},\,\bm{y}\rangle - \min_{\bm{y}\in\mathcal{Y}}\varphi(\bm{y})\right)\\
&\leq \frac{2R_\mathcal{C}}{t}||\bm{\theta}|| + \frac{2R_\varphi}{t},
\end{align*}

\noindent where $R_\mathcal{C} = \max_{\bm{y}\in\mathcal{Y}}||\bm{y}||$ and $R_\varphi =  \max_{\bm{y}\in\mathcal{Y}}|\varphi(\bm{y})|$. Thus, we have: 
$$\lambda'_2 \leq 1 - G|\mathcal{Y}|\exp\left(-\frac{4R_\varphi}{t}\right)\exp\left(-\frac{4R_\mathcal{C}}{t}||\bm{\theta}||\right),
$$
and finally:

$$
\lambda_{\bm{\theta}} \leq 1 - \frac{G|\mathcal{Y}|\exp\left(-\frac{4R_\varphi}{t}\right)}{2}\exp\left(-\frac{4R_\mathcal{C}}{t}||\bm{\theta}||\right),
$$

so taking $D = 4R_\mathcal{C}/t$ concludes the proof. 
    
\end{proof}

\begin{remark}
    The stationary distribution in \citet{ingrassia_rate_1994} is defined as proportional to $\exp\left(-H(\bm{y})\right)$, with the assumption that the function $H$ is such that $\min_{\bm{y}\in\mathcal{Y}}H(\bm{y}) = 0$. Thus, we apply their results with $$H(\bm{y})\coloneqq \max_{\bm{y}'\in\mathcal{Y}}\left[\frac{\langle\bm{\theta},\,\bm{y}'\rangle+ \varphi(\bm{y}')}{t}\right] - \frac{\langle \bm{\theta}, \bm{y}\rangle + \varphi(\bm{y})}{t}$$ (which gives correct distribution ${\pi}_{\bm{\theta},t}$ and respects this assumption), hence the obtained forms for $Z(\bm{\theta})$ and the upper bound on $m(\bm{\theta})$. 
\end{remark}

\subsection{Proof of \cref{prop:1step_fyl_unsup}}
\label{proof:1step_fyl_unsup}
\begin{proof}
    In the unconditional setting, given a dataset $(\y_i)_{i=1}^N$, the distribution of the first iterate of the Markov chain with transition kernel defined in \cref{eq:markov_kernel} and initialized by $\y^{(0)}=\y_i$, with $i\sim\mathcal{U}(\llbracket 1, N\rrbracket)$, is given by:
    \begin{align*}
    (\pfirstunsup)(\y) &= \sum_{\y'\in\mathcal{Y}}\left(\sum_{i=1}^N\bm{1}_{\{\y_i=\y'\}}\cdot \frac{1}{N}\right)P_{\bm{\theta},t}(\y',\y) \\
    &= \sum_{\y'\in\mathcal{Y}}\left(\sum_{i=1}^N\bm{1}_{\{\y_i=\y'\}}\cdot \frac{1}{N}\right)\bm{p}^{(1)}_{\bm{\theta},\y'}(\y)\\
    &= \frac{1}{N}\sum_{i=1}^N \bm{p}^{(1)}_{\bm{\theta},\y_i}(\y).
\end{align*}

Thus, keeping the same notations as in the previous proof, previous calculations give:
\begin{align*}
    \EE_{\pfirstunsup}[Y] &= \frac{1}{N}\sum_{i=1}^N \EE_{\bm{p}^{(1)}_{\bm{\theta},\y_i}}[Y] \\
    &= \frac{1}{N}\sum_{i=1}^N \nabla_{\thetav} F_{\y_i}(\thetav)\\
    &= \nabla_{\thetav}\left(\frac{1}{N}\sum_{i=1}^N F_{\y_i}\right)(\thetav).
\end{align*}
Let $F_{\bar{Y}_N}\coloneqq \frac{1}{N}\sum_{i=1}^N F_{\y_i}$  Then, the exact same arguments as in the conditional case hold, and the results of \cref{prop:1step_fyl_unsup} are obtained by replacing $F_{\y}$ with $F_{\bar{Y}_N}$ in the proof of \cref{prop:1step_fyl_sup}, and noticing that the previously shown $\EE_{q(\y_i,\,\cdot\,)}||Y-\y_i||^2_2/t$-smoothness of $F_{\y_i}$ gives that $F_{\bar{Y}_N}$ is $\frac{1}{N}\sum_{i=1}^N\EE_{q(\y_i,\,\cdot\,)}||Y-\y_i||^2_2/t$-smooth. Similar arguments also hold for the regularized optimization formulation, by noting that this time we have $\nabla F_{\bar{Y}_N}(\RR^d)\subset \conv\left(\bigcup_{i=1}^N\left\{\mathcal{N}(\y_i)\cup\{\y_i\}\right\}\right)$.
\end{proof}

\subsection{Proof of \cref{prop:one_step_properties}}
\label{proof:one_step_properties}
\begin{proof}
    The first point is directly given by the fact that $\EE_{\pfirstsup}[Y]$ is the expectation of a distribution over $\cN(\y)\cup\{\y\}$. For the second and third points, as derived in \cref{proof:1step_fyl_sup}, we have:
    \begin{align*}
        \EE_{\pfirstsup}[Y] = \y +\!\!\!\!\!\!\sum_{\bm{y}'\in\mathcal{N}_{\y}^-(\bm{\theta})}\!\!q(\bm{y}',\y)\exp\left(\left[\langle\bm{\theta}, \bm{y}'-\y\rangle + \varphi(\bm{y}')-\varphi(\y)\right]/t\right)\cdot\left(\bm{y}'-\y\right) \; +\!\!\!\!\!\sum_{\bm{y}'\in\mathcal{N}_{\y}^+(\bm{\theta})}\!\!q(\y,\bm{y}')\cdot\left(\bm{y}'-\y\right).
    \end{align*}
    Define then:
    \begin{align*}
        \cN_{\text{better}}(\y)&\coloneqq\left\{\y'\in\cN(\y)\mid\langle\thetav,\y'\rangle+\varphi(\y')>\langle\thetav,\y\rangle+\varphi(\y)\right\},\\
        \cN_{\text{worse}}(\y)&\coloneqq\left\{\y'\in\cN(\y)\mid\langle\thetav,\y'\rangle+\varphi(\y')<\langle\thetav,\y\rangle+\varphi(\y)\right\}
    \end{align*}
    as the sets of improving and worsening neighbors of $\y$ respectively (assuming no neighbor of $\y$ has exactly equal objective value for simplicity, which is true almost everywhere w.r.t. $\thetav\in\RR^d$).
    \paragraph{Low temperature limit.} We have:
    \begin{align*}
        \cN_{\y}^+(\thetav)\xrightarrow[t\rightarrow 0^+]{}\cN_{\text{better}}(\y), \quad\text{and}\quad \cN_{\y}^-(\thetav)\xrightarrow[t\rightarrow 0^+]{}\cN_{\text{worse}}(\y).
    \end{align*}
    Then, as $x<0\implies \exp(x/t)\xrightarrow[t\rightarrow 0^+]{}0$, we have effectively
    $$
    \EE_{\pfirstsup}[Y] \xrightarrow[t\rightarrow 0^+]{}\y+ \sum_{\y'\in\cN_{\text{better}}(\y)}q(\y,\y')\cdot(\y'-\y).
    $$
    \paragraph{High temperature limit.} As $\forall x\in\RR,\;\exp(x/t)\xrightarrow[t\rightarrow \infty]{} 1$, we have:
    \begin{align*}
        \cN_{\y}^+(\thetav)\xrightarrow[t\rightarrow \infty]{}\left\{\y'\in\cN(\y)\mid q(\y',\y)>q(\y,\y')\right\}, \quad\text{and}\quad \cN_{\y}^-(\thetav)\xrightarrow[t\rightarrow \infty]{}\left\{\y'\in\cN(\y)\mid q(\y',\y)\leq q(\y,\y')\right\}.
    \end{align*}
    Thus, we have:
    \begin{align*}
        \EE_{\pfirstsup}[Y] \xrightarrow[t\rightarrow \infty]{}\y+ \sum_{\y'\mid q(\y',\y)\leq q(\y,\y')}q(\y',\y)\cdot (\y'-\y) + \sum_{\y'\mid q(\y',\y)> q(\y,\y')}q(\y,\y')\cdot (\y'-\y),
    \end{align*}
    which gives effectively:
    \begin{align*}
        \EE_{\pfirstsup}[Y] \xrightarrow[t\rightarrow \infty]{}\y+ \sum_{\y'\in\cN(\y)}\min\left[q(\y,\y'),q(\y',\y)\right]\cdot (\y'-\y).
    \end{align*}
\end{proof}

\end{document}